# MechGPT, a language-based strategy for mechanics and materials modeling that connects knowledge across scales, disciplines and modalities

Markus J. Buehler[1,2]*

[1] Laboratory for Atomistic and Molecular Mechanics (LAMM), Massachusetts Institute of Technology, 77 Massachusetts Ave., Cambridge, MA 02139, USA

[2] Center for Computational Science and Engineering, Schwarzman College of Computing, Massachusetts Institute of Technology, 77 Massachusetts Ave., Cambridge, MA 02139, USA

*mbuehler@MIT.EDU

**ABSTRACT**: For centuries, researchers have sought out ways to connect disparate areas of knowledge. While early scholars (Galileo, da Vinci, etc.) were experts across fields, specialization has taken hold later. With the advent of Artificial Intelligence, we can now explore relationships across areas (e.g., mechanics-biology) or disparate domains (e.g., failure mechanics-art). To achieve this, we use a fine-tuned Large Language Model (LLM), here for a subset of knowledge in multiscale materials failure. The approach includes the use of a general-purpose LLM to distill question-answer pairs from raw sources followed by LLM fine-tuning. The resulting MechGPT LLM foundation model is used in a series of computational experiments to explore its capacity for knowledge retrieval, various language tasks, hypothesis generation, and connecting knowledge across disparate areas. While the model has some ability to recall knowledge from training, we find that LLMs are particularly useful to extract structural insights through Ontological Knowledge Graphs. These interpretable graph structures provide explanatory insights, frameworks for new research questions, and visual representations of knowledge that also can be used in retrieval-augmented generation. Three versions of MechGPT are discussed, featuring different sizes from 13 billion to 70 billion parameters, and reaching context lengths of more than 10,000 tokens. This provides ample capacity for sophisticated retrieval augmented strategies, as well as agent-based modeling where multiple LLMs interact collaboratively and/or adversarially, the incorporation of new data from the literature or web searches, as well as multimodality.

## 1. Introduction

Modeling the physical, biological and metaphysical concepts has been a focus of researchers from many disciplines. Early scientists and engineers were often deeply embedded in multiple fields from science to philosophy to physics to mathematics and also the arts (e.g., Galileo Galilei, Leonardo da Vinci, Johann Wolfgang von Goethe), specialization has taken hold in later time. This was partly due to the extreme amounts of knowledge accumulated across fields, which required humans to deeply focus. Now, the emergence of large language models (LLMs) has challenged paradigms of scientific inquiry and not only led to new AI/ML based modeling strategies, but also resulted in opportunities to connect knowledge, ideas and concepts across [1–20]. Such models can complement conventional multiscale modeling for the analysis and design of hierarchical materials [21–24], and many other applications in mechanics [25]. Here we build on recently proposed uses of LLMs in mechanics and materials research (see, [26]) and develop, based on the Llama-2 based OpenOrca-Platypus2-13B general-purpose LLM [27], a fine-tuned MechGPT model focused to model materials failure, multiscale modeling and related disciplines. The OpenOrca-Platypus2-13B model was chosen due to its reported high level of performance on key tasks, such as reasoning, logic, math/science, and other disciplines, rendering a powerful access to broad, transferrable knowledge and universal concepts across disciplines, at a manageable model size that offers computational efficiency.

LLMs have powerful applications in science. In addition to enabling the analysis of large amounts of data and complex systems, in mechanics and materials science, LLMs can be used to simulate and predict the behavior of materials under different conditions, such as mechanical stress, temperature, and chemical interactions, to name a few. As shown in earlier work [26], by training LLMs on large datasets of molecular dynamics simulations, researchers can develop models that can predict the behavior of materials in new situations, accelerating the discovery process and reducing the need for experimental testing. Such models are further useful for analyzing scientific texts such as books and publications, allowing researchers to quickly extract key information and insights from large volumes of data. This can help scientists to identify trends, patterns, and relationships between different concepts and ideas, and to generate new hypotheses and ideas for further research. Here we focus the development on the latter, and explore the use of MechGPT, a generative AI tool in the family of transformer-based LLMs that is trained specifically on materials failure and associated multiscale methods, to assess the potential of these strategies.

The proposed strategy includes several steps, including first a distillation step in which we use of an LLM to generate question-answer pairs from chunks of raw data, such as text extracted from one or more PDF files, and then in a second step to utilize this data to fine-tune a model. The initial MechGPT model explored here is specifically trained in the area of atomistic modeling of materials failure, and demonstrate its usefulness for knowledge retrieval, general language tasks, hypothesis generation, and others. The plan of this paper is as follows. First, we introduce the overarching modeling strategy, dataset generation using a specific language-modeling strategy to extract knowledge from sources, and then training of the model with a novel mechanics and materials dataset. After some general remarks about the model, prompting, and how it is trained, we then apply the model and test its performance in a variety of settings, including the use of LLMs for Ontological Graph Generation and developing insights about complex subjects across disciplines, as well as agent modeling where multiple LLMs interact in a collaborative or adversarial manner to yield deeper insights into topical areas or question answering. We further provide a conceptional comparison between language models and many-particle systems at different levels of abstraction, and explain how the new framework can be viewed as a means to extract generalizable relationships that govern complex systems. We conclude with an outlook to various challenges and opportunities, including a critical discussion of shortcomings and possibilities for future research in this field.

## 2. Results and discussion

The language model used here is based on attention mechanisms that are used widely in many transformer models for sequence and graph data [28] (**Figure 1**), forming a model in the general family of Generative Pre-trained Transformers (GPT). Such models can relatively easily be pretrained and fine-tuned against complex linguistic, scientific and mathematical relationships [29,30]. The basis for the smallest MechGPT base model is the OpenOrca-Platypus2-13B model (a LLM with 13 billion parameters). The model is originally based on the Llama 2 transformer architecture, a highly performant open-source foundation model trained on 2 trillion tokens with performance comparable to GPT3.5 in some tests, but in an open access format. The OpenOrca-Platypus2-13B model was trained, and has been shown to perform exceptionally well in a variety of tasks. For instance, in the AGI Eval test [31] it averages 0.463, owing to high performance in the LSAT Logical Reasoning assessment that rivals or exceeds much larger models.

### 2.1 Training set development

Often, data sources with relevant information are messy and not clean. LLMs, however, are best fine-tuned on clean data and insights to learn from, ideally via question-answer pairing. We first need to select a corpus of knowledge that we want the model to be trained on. In our case, we select a book published by the author of this paper ("*Atomistic modeling of materials failure*" [32]). While it may be possible to train the model directly on chunks of text extracted from the book, an effective strategy proposed here is to include first a distillation step, whereby we analyze the raw data in an unsupervised manner, and based off this process, autonomously develop a sound set of scientific question-answer relationships as the training set.

For this initial stage of training set development, we use a general-purpose LLM – here, the Llama 2 70b chat model – to extract knowledge from the text in the form of question-answer pairs (**Figure 2**). We divide the original data source into chunks of data, here ~500 words long, of which there are around 430 in total. We then show the LLM these ~500-word long sections of the book and run two sequential inference steps:

> **Step 1**: We ask the general-purpose LLM to develop a scientific question to which the answer is the section considered
> **Step 2**: We ask the general-purpose LLM to develop a concise summary of the section considered.

(Details, see **Materials and Methods**.) This results in a clean pair of question-answer relationships. We may repeat this process multiple times, as needed, to garner more training data (in our case, we repeated the process three times to generate a total set of ~1,290 question-answer pairs). One can also change the size of the sections of the book to be smaller or larger (albeit, we focus our discussion on the approach with ~500 word long sections as this provides a good level of detail to formulate meaningful question-answer pairs). It is noted that this process can be repeated also to add multiple training sources (e.g., other books, papers, Wikipedia, etc.) including also specific calculation tasks (e.g. calculate modulus of CNT, calculate protein strength, along with the inverse tasks) to the training data as was done in earlier work [26].

## 2.2 Fine-tuning the model

With the question-answer dataset generated as described in **Section 2.1**, we now proceed to fine-tune the OpenOrca-Platypus2-13B base LLM, using a Low-Rank Adaptation (LoRA) strategy [33]. The LoRA adaptor strategy is powerful since it allows us to freeze the original model – thereby, retaining its superb performance, while endowing it with more specialized domain knowledge via the added adaptor layers while avoiding catastrophic forgetting. The key aspect of the approach is to retain the original capacities of the model while adding new information. With LoRA adaptors, the resulting MechGPT framework can solve a wide range of language-based tasks (summarization, knowledge extraction, idea generation, etc.) with the added benefit of a host of additional capabilities (knowledge extraction, generation of new design ideas, language tasks such as summarization, creative writing, etc.). These examples show the high potential these types of models have for engineering in general.

More details on the methods and tools, as well as algorithmic choices, are provided in the **Materials and Methods** section of the paper.

## 2.3 Results and predictions

Once trained, the model is applied to solve a variety of mechanics-specific tasks, including broad categories of:

- General behaviors and knowledge retrieval
- Research development
- Generation of structured insights via Ontological Knowledge Graphs, or ologs
- Creative applications to connect across domains of knowledge
- Agent-based modeling of multiple LLMs interacting to autonomously form a conversation between AI agents

We will discuss these in detail in the next several sections. Full sample conversations are included in the Supplementary Information, whereas the presentation of results focuses on key segments of the output. The first part of the results includes a few general considerations and behaviors. We then dive deeper into specifics and explore the behaviors of the model via a series of computational experiments in the various contexts summarized above.

### 2.3.1 General framework and behaviors: Background

The behavior of the model is influenced by a large variety of features, owing in part due to the high flexibility of LLMs to process different kinds of information. Here we explore effects of prompting and other parameters systematically via a series of computational experiments. The results are divided into several subsections to help guide the reader. First, we briefly review the general strategy used to interact with the model, at a detailed level. The interaction occurs by providing data to the model, and asking the model to respond. Then, the user may add a new query (with perhaps new context, content or questions) and the model responds again. This leads to a multi-turn chat interaction in the general format:

> **<start token>System prompt. User query #1**: …**<end token>** MechGPT response #1: …. <end token> **User query #2**: ….**<end token>** MechGPT response #2: ….<end token> ……

Each of the inputs and responses ends with an end token that signals that the input or output has ended. The end token is used during generation to determine when the model should stop generating additional tokens. As the conversation progresses, the text considered by the model grows larger. Previous data is invoked as the conversation carries on, and used to generate the interaction graphs delineated in **Figure 1b**. As for a specific example, here is how this looks like as 'raw' text:

> **\<s\>This is a conversation between a User and an AI Assistant, MechGPT. MechGPT is knowledgeable in materials failure, in particular atomistic and multiscale modeling. If you don't know the answer to a question, please don't share false information. Keep answers concise and accurate.**
> **\<|end_of_turn|\>User: Hello MechGPT, provide a one-sentence summary of stress-assisted corrosion.**
> **\<|end_of_turn|\>Assistant:** Stress-assisted corrosion is a process where cracks in a material facilitate corrosion, leading to a higher rate of degradation than in uncracked regions.\<|end_of_turn|\>**User: Thank you!\<|end_of_turn|\>**

Here, \<s\> is the start token to indicate the beginning of the data, and \<|end_of_turn|\> denotes the end token to signify the end of each input/output. Keywords of "User:…" and "Assistant: …" are used to mark the role of the entities in the exchange. In the following, we simply write this exchange in a clean format:

> **Hello MechGPT, provide a one-sentence summary of stress-assisted corrosion.**
>
> Stress-assisted corrosion is a process where cracks in a material facilitate corrosion, leading to a higher rate of degradation than in uncracked regions.
>
> **Thank you!**

The overall process by which information is processed by the model is visualized in **Figure 3**, showing a schematic of the workflow, featuring a system prompt, context/background to the query, additional details, and other information, typically followed by a question or task (the query can take many different forms, as the examples presented in this paper show). The response can be followed, iteratively, with repeating interactions between the user and MechGPT as outlined above. A key aspect of the prediction process is that all previous tokens – referred to also as context – is used to predict the result.

Interaction with the model is implemented via a Gradio app, of which a screenshot is shown in **Figure 4** (movies of interactions with the model are shown via **Movies M1 and M2**). As shown in the figure, the MechGPT app features three main parts, the main chat window (top), and two optional sections that can be independently collapsed. The literature search window enables users to enter search term for a Google Scholar search that are then displayed (users can use the information in the text box, or copy-paste the URL to papers for further study and analysis, and then feed data back into the main chat module for interaction with MechGPT). The advanced options provide access to control parameters such as the sampling temperature, maximum number of tokens generated, and other sampling parameters.

### 2.3.2 Knowledge retrieval results

Often, knowledge retrieval is used as a basic benchmark to assess the performance of LLMs; albeit we note known challenges of using LLMs for knowledge retrieval and the need for independent validation of results [34]. We start by querying the model with simple multiple-choice questions, akin to related assessments in the AGI test sets used to measure model performance. **Table 1** shows a longer series of sample responses to assess knowledge retrieval, showing reasonable performance across the board (in all examples, the prompt by the human is given in **bold**, and the responses in regular font). While the queries in **Table 1** were rather focused, the model can also be queried against much broader questions. Here is an example:

> **Write a summary of the key principles that govern the dynamics of cracks in brittle materials, especially related to the crack speed.**
>
> When it comes to the dynamics of cracks in brittle materials, there are several key principles that govern their behavior, particularly in relation to the crack speed. Firstly, the energy release rate, or G, is a crucial concept, as it determines the likelihood of crack propagation. Secondly, the elasticity of the material near the crack tip plays a significant role in the crack speed, with stiffer materials resulting in higher crack speeds. Additionally, the size of the crack and the material properties, such as the fracture toughness, all influence the crack speed.
>
> In order to accurately predict the crack speed, a comprehensive understanding of these principles is essential. This includes the analysis of the elastic and plastic deformation of the material near the crack tip, as well as the consideration of the size of the crack and the material properties. By incorporating these factors, it is possible to accurately determine the crack speed and predict the behavior of cracks in brittle materials.

Furthermore, the use of atomistic and multiscale modeling is a powerful tool for understanding the behavior of cracks in brittle materials. By simulating the atomic-scale interactions between the material and the crack, researchers can gain a deep understanding of the mechanisms that govern crack propagation. This knowledge can then be used to develop accurate predictions of crack speed and to inform the design of materials with improved fracture properties.

In summary, the dynamics of cracks in brittle materials is governed by a complex interplay of principles, including the energy release rate, elasticity, material properties, and crack size. By understanding these principles and incorporating them into predictive models, researchers can accurately predict the crack speed and inform the design of materials with improved fracture properties.

**Thank you. This is great. Now, tell me more about the mechanisms near the crack tip, especially when it comes to nonlinear material behavior.**

When it comes to the mechanisms near the crack tip, particularly in the context of nonlinear material behavior, there are several key concepts that are important to understand. Firstly, the elasticity of the material near the crack tip is crucial, as it determines the likelihood of crack propagation. Stiffer materials will result in higher crack speeds, while softer materials will lead to lower crack speeds.

In addition to elasticity, the size of the crack and the material properties, such as the fracture toughness, also play a significant role in the mechanisms near the crack tip. The interaction between these factors can lead to complex behavior, including the formation of daughter cracks, the change in the shape of the crack tip, and the development of shear and/or rotational components in the deformation field.

To accurately predict the behavior near the crack tip, it is essential to have a comprehensive understanding of these mechanisms. This includes the analysis of the elastic and plastic deformation of the material near the crack tip, as well as the consideration of the size of the crack and the material properties. By incorporating these factors, researchers can gain a deep understanding of the mechanisms near the crack tip and predict the behavior of cracks in brittle materials with greater accuracy.

Furthermore, the use of atomistic and multiscale modeling is a powerful tool for understanding the mechanisms near the crack tip. By simulating the atomic-scale interactions between the material and the crack, researchers can gain a deep understanding of the mechanisms that govern crack propagation. This knowledge can then be used to develop accurate predictions of crack behavior and to inform the design of materials with improved fracture properties.

In summary, the mechanisms near the crack tip in nonlinear materials are governed by a complex interplay of factors, including elasticity, material properties, and crack size. By understanding these factors and incorporating them into predictive models, researchers can accurately predict the behavior near the crack tip and inform the design of materials with improved fracture properties

Knowledge can also be retrieved in the context of connecting insights from different areas to a new case. In this example, we ask the model to predict how a tree leaf would fail:

**Drawing on your expertise in fracture of metals, polymers, proteins, ceramics and other engineering materials, how would you expect that a tree leaf would fail. Discuss specific mechanisms you would expect to see.**

MechGPT: Of course! So, to discuss the failure of a tree leaf, we need to consider the material it's made of, which is predominantly a biological material called lignin and cellulose. These materials are composed of long chains of organic molecules, primarily composed of carbon, hydrogen, and oxygen. When these materials are subjected to stress and load, they can fail in a few ways.

One mechanism we'd expect to see is the breaking of these long organic chains, which is a bit like the snapping of a dry stick. This failure occurs when the stresses in the material become too high, and the bonds between the molecules start to break.

Another mechanism is the rupture of the cell walls that surround the vascular tissues within the leaf. These cell walls are made of a strong, fibrous material called cellulose, which can withstand a significant amount of stress before failing. When the stresses in the leaf become too high, the bonds between these cellulose fibers start to break, leading to the rupture of the cell walls.

Finally, we'd also expect to see some deformation of the material before it ultimately fails. This deformation is a result of the bending and bending twinning of the long organic molecules, which occurs when the stresses in the material become high enough to cause this deformation.

So, in summary, we'd expect the failure of a tree leaf to occur through the breaking of long organic chains, rupture of cell walls, and deformation of the material. These mechanisms are typical of many biological materials and are important to understand in order to design and optimize materials for various applications.

Several additional examples are shown in **Supplementary Table S1**. These examples focus on the strength of nanomaterials (including relating nanomaterials with biological principles), the application of materials failure studies in a new field (here: battery research), the behavior of thin copper films at elevated temperature, and bulk copper plasticity. In all cases, the model excels in not only providing accurate answers, but also suggesting specific methods or details, and wider context especially when it comes to multidisciplinary investigations.

### 2.3.3 Sampling temperature effects

MechGPT uses temperature scaling during the generation of text to control the randomness and diversity of the output. A simple way to visualize it is to view it as the 'temperature of uncertainty': Higher temperature values introduce more uncertainty into the sampling process, allowing for more exploration of different possibilities. Lower temperature values reduce uncertainty and make the sampling more deterministic. Therein, temperature is an important parameter that influences the level of creativity and randomness in the generated text. At higher sampling temperatures ($>1$), even lower probability tokens have a chance of being selected, leading to more diverse output and thereby higher levels of creativity. Conversely, lower sampling temperatures ($<0.5$) make the distribution of probabilities sharper and tokens with higher probabilities will have an even higher likelihood of being selected. As a result, the output text is more focused and deterministic as the model is more likely to choose tokens it is more confident about. As a result, the generated predictions are often more coherent and relevant, but may lack some diversity and creativity.

In addition to the prompt used to query the model, there are effects of the sampling temperature on the result. **Table 2** shows sample responses to assess the effects of sampling temperature on the results. We generally find that $T$=0.5 is generally a good choice as sampling temperature for the model to be able to make new connections while not hallucinating facts. It is noted that choices a bit lower can help in knowledge retrieval tasks, and larger values are suitable for more creative tasks such as research design, connecting across domains of knowledge, and similar. **Table 3** provides a summary of various parameters used during sampling.

### 2.3.4 System prompt effects

**Table 4** shows effects of system prompt on model behavior, which can be an effective way to control the overall behavior of the model. The prompt is the same of all cases: "**At a crack tip, what happens to the deformation of the material?**" and the sampling temperature is chosen to be $T$=0.3. As can be seen in the results, the use of variations of the system prompt allows us to tweak the behavior of the model in significant ways.

### 2.3.5 Research development

Now that we developed a general understanding of how the model behaves, along with foundational effects of sampling temperature and system prompts, we now explore the use of the model for more complex tasks. The first in this series of analyses focuses on research development, where we provide the model with samples of work and query for future research ideas, hypotheses, and related tasks. We focus on the following example, but additional sample conversations are included as **Supplementary Materials**. Note that the abstract used in the example is taken verbatim from a recently published paper from the literature [35]:

> **Summarize this scientific abstract and list a few bullet points with the key takeaways: "Complex hierarchical structures composed of simple nanoscale building blocks form the basis of most biological materials. Here we demonstrate how analogies between seemingly different fields enable the understanding of general principles by which functional pnanorties in hierarchical systems emerge, similar to an analogy learning process. Specifically, natural hierarchical materials like spider silk exhibit properties comparable to classical music in terms of their hierarchical structure and function. As a comparative tool here we apply hierarchical ontology logs (olog) that follow a rigorous mathematical formulation based on category theory to provide an insightful system representation by expressing knowledge in a conceptual map. We explain the process of analogy creation, draw connections at several levels of hierarchy and identify similar patterns that govern the structure of the hierarchical systems silk and music and discuss the impact of the derived analogy for nanotechnology."**

- The abstract discusses the understanding of functional properties in hierarchical systems through analogies between different fields.
- It uses a hierarchical ontology logs (olog) to represent system information as a conceptual map.
- The analogy drawn is between natural hierarchical materials like spider silk and classical music.
- The process of analogy creation, drawing connections at multiple levels of hierarchy, and identifying similar patterns are discussed.
- The impact of the derived analogy is on nanotechnology.

> **Now propose a follow-up research study, incorporating elements of materials failure, philosophy and mathematics.**

- Study the mechanical strength of spider silk by subjecting it to various loads and recording the deformation and fracture patterns.
- Analyze the hierarchical structure of spider silk at the atomic level using atomistic modeling.
- Compare the mechanical properties of spider silk with other biological materials like silk and protein fibers.

- Develop mathematical models to describe the behavior of the material under different loads.
- Test the material under real-world conditions, such as loads associated with spider webs and their applications.
- Study the philosophical implications of the analogy between spider silk and classical music, particularly in terms of design, optimization, and self-assembly.
- Collaborate with musicians and artists to create artistic installations and performances inspired by the analogy.
- Publish the findings in a scientific journal and present the research at international conferences.
- This follow-up research study incorporates elements of materials failure, philosophy, and mathematics to create a unique and interdisciplinary approach to understanding the behavior of complex hierarchical systems. The study aims to provide valuable insights into the mechanical strength of spider silk and its applications, as well as the broader implications of the analogy drawn between this material and classical music.

**Now add a deeper discussion on the relationship between mechanics of failure and philosophy in the context of the use of ologs.**

- Discuss the mechanical strength of spider silk in the context of failure mechanisms, such as fracture and deformation.
- Explore the philosophical implications of the analogy between spider silk and classical music, particularly in terms of design, optimization, and self-assembly.
- Consider the role of ologs in understanding the hierarchical structure of spider silk and its mechanical properties.
- Compare the results of the follow-up research study with previous studies on spider silk and discuss the significance of the analogy drawn in the context of nanotechnology.
- Develop mathematical models to describe the behavior of spider silk under different loads and test the material under real-world conditions.
- By incorporating elements of philosophy and mathematics, the follow-up research study provides a deeper understanding of the relationship between mechanics of failure and philosophy in the context of the use of ologs. The study aims to create a unique and interdisciplinary approach to understanding the behavior of complex hierarchical systems, with implications for nanotechnology and the arts.

This example, along with the other ones, shows that the model generally does an excellent job in summarizing a study and then following up with next research steps. It is noted that the model performs generally better if we guide it through the process. That is, rather than requesting to suggest next steps directly based on the input of a paper or an abstract, it is good practice to first ask for a summary of salient points, then ask to elaborate on next steps. As shown in the example, follow-up queries can help steer the result in certain directions.

**Table S2** shows additional experiments. For instance, Conversation sample #1 in **Table S2** focuses on development of a model for a titanium alloy, including queries about modeling technique and specifics on force fields. It also shows how we can query the model to provide clarifications or to give specific details about an answer, as done for the use of AFM in this study. Conversation sample #2 in **Table S2** focuses on a protein mechanics study, question asking about a detail (here, whether the Finite Element Method was used in a particular research work), summarizing at different lengths and detail level, and translation to a different language. These samples show the wide possibilities of the approach beyond technical expertise, logic and language tasks generally.

The model can also successfully connect different areas of knowledge in mechanics and materials. For instance, in **Table 5** we show how the model can relate hyperelasticity in dynamic fracture with protein unfolding and generate ideas for further research exploration.

MechGPT can also be used in nonlinear prompting strategies (**Figure 3b**). For instance, in the following workflow we ask MechGPT to first generate a set of ideas, then criquite the ideas, and then distill key insight into the final response. The system prompt used here is "You are MechGPT, a helpful assistant with knowledge in mechanics and materials, especially multiscale modeling."

I want a response to the following question:
Describe a multiscale modeling strategy to discover fracture mechanisms in nanotwinned copper.
Name a set critical scientific concepts that are crucial to answer the question.
Don't answer the question yet.

Sure, here's a response to the question "Describe a multiscale modeling strategy to discover fracture mechanisms in nanotwinned copper. Name a set critical scientific concepts that are crucial to answer the question. Don't answer the question yet."

A multiscale modeling strategy to discover fracture mechanisms in nanotwinned copper would involve a set of critical scientific concepts that are crucial to answer the question. These concepts include:
1. Atomic-scale construction of the twin boundary.
2. Nano-scale fracture mechanisms at the twin grain boundary.
3. Microscale or mesoscale continuum modeling for stress and deformation fields.

4. Macroscale behavior of the material under different loading conditions.

The multiscale modeling strategy would involve integrating information from various scales, starting from the atomic scale, where the construction of the twin boundary is described using first principles methods, such as density functional theory (DFT). This information is then used to inform the nano-scale fracture mechanisms at the twin grain boundary, which are also described using first principles methods. The nano-scale mechanisms are then used to inform the continuum modeling at the microscale and mesoscale, where the stress and deformation fields are calculated using finite element methods. The macroscale behavior of the material under different loading conditions is also considered, using experimental and theoretical methods.

This multiscale modeling strategy allows for a more complete understanding of the fracture mechanisms in nanotwinned copper, and can help guide the development of new materials with improved mechanical properties.

**Now, define each of these concepts.**

1. Atomic-scale construction of the twin boundary - This concept refers to the description of the structure of the twin boundary at the atomic scale. The twin boundary is a unique feature of some materials, such as nanotwinned copper, where it consists of two partial dislocations running parallel to each other. The atomic-scale construction of the twin boundary is important to understand the fracture mechanisms in these materials.

2. Nano-scale fracture mechanisms at the twin grain boundary - This concept refers to the description of the fracture mechanisms that occur at the twin grain boundary, which is the interface between the two partial dislocations that make up the twin boundary. The nano-scale fracture mechanisms are important to understand the stress and deformation fields around the twin boundary, which can affect the material's overall behavior under different loading conditions.

3. Microscale or mesoscale continuum modeling - This concept refers to the use of continuum models, such as finite element methods, to simulate the stress and deformation fields around the twin boundary at the microscale and mesoscale. The continuum modeling is important to understand the macroscale behavior of the material under different loading conditions, and can be used to guide the development of new materials with improved mechanical properties.

4. Macroscale behavior of the material under different loading conditions - This concept refers to the description of the material's behavior under various loading conditions, such as tension, compression, and shear. The macroscale behavior is important to understand the material's overall mechanical properties, and can be used to guide the development of new materials with improved performance.

These four concepts are critical to answer the question of discovering fracture mechanisms in nanotwinned copper using a multiscale modeling strategy. By integrating information from various scales, a more complete understanding of the fracture mechanisms can be achieved, which is important for the development of new materials with improved mechanical properties.

**Great, now accurately answer the question by incorporating these concepts. The answer must be brief.**

A multiscale modeling strategy to discover fracture mechanisms in nanotwinned copper would involve integrating information from various scales, starting from the atomic scale, where the construction of the twin boundary is described using first principles methods, such as density functional theory (DFT). This information is then used to inform the nano-scale fracture mechanisms at the twin grain boundary, which are also described using first principles methods. The nano-scale mechanisms are then used to inform the continuum modeling at the microscale and mesoscale, where the stress and deformation fields are calculated using finite element methods. The macroscale behavior of the material under different loading conditions is also considered, using experimental and theoretical methods. This strategy allows for a more complete understanding of the fracture mechanisms in nanotwinned copper, and can help guide the development of new materials with improved mechanical properties.

These types of sampling strategies are powerful especially for answering complex questions and tend to perform better than single-shot question-answer strategies. Moreover, we could use multiple LLMs to generate ideas and synthesize results; a particular variation of this ideas will be discussed later in this paper via the use of interacting LLM agents.

Transformer-based models are internally structured to use graph-forming neural networks to learn structural representations in data. We can build out this innate capability to elicit human-readable knowledge graph representations of complex datasets, specific areas of knowledge, or drive how retrieval of information is conducted as a model responds to a task. To illustrate this powerful capacity, we develop Ontological Knowledge Graphs, or ologs, for the intersecting knowledge presented in **Table 5**. **Figure 6** shows the result of this experiment, revealing an Ontological Knowledge Graph representation to relate hyperelasticity in dynamic fracture with protein unfolding; here, represented mathematically and visually as a graph that can be easily understood, mined and used for a variety of downstream tasks.

This graph representation can be used to develop further explorations, such as the development of research hypotheses, based on retrieval-augmented strategies where graph relationships are used as context for generation to answer a specific question (see, the schematic insert in **Figure 3**, bottom left). For instance, by using the prompt: "**Develop 1) a new research hypothesis and 2) specific experiments to understand the analogy between hyperelasticity in the context of supersonic fracture and protein unfolding mechanisms.**," we obtain:

1) A new research hypothesis that can be developed based on the analogy between hyperelasticity in the context of supersonic fracture and protein unfolding mechanisms is that both phenomena share a common theme of the elasticity of large strains playing a crucial role in dynamic fracture and protein denaturation processes. This hypothesis can be tested through a combination of atomistic and continuum analyses, using a multiscale modeling framework that incorporates both scale-dependent inhomogeneities of the elastic constants.

2) Specific experiments to understand the analogy between hyperelasticity in the context of supersonic fracture and protein unfolding mechanisms can be designed by combining atomistic and continuum analyses. These experiments can include the use of a combined atomistic and continuum reference system, where the scale-dependent inhomogeneities of the elastic constants are incorporated into the reference system. This reference system can be used to predict the behavior of materials under different conditions, such as the fracture of brittle materials or the unfolding of proteins, and provide insights into their mechanical and physical properties.

Overall, the analogy between hyperelasticity in the context of supersonic fracture and protein unfolding mechanisms offers a new perspective on both phenomena, which can lead to new applications and advancements in materials science and protein research.

### 2.3.6 Creative applications to connect across domains of knowledge

The model shines especially when tasked to make connections between different areas of knowledge. Here is an example where we ask the model to produce a poem in the style of a particular author, and then use the response to garner additional insights into the topic and other creative avenues for exploration (to develop a deeper understanding of concepts and relationships across disciplines).

**Write a poem about brittle fracture, in the style of Goethe.**

Once upon a time, in a land of matter,
A phenomenon occurred with such fury and clatter.
Brittle fracture, a crack so sharp and clear,
Cut through substances with a vengeance so dear.

Atom after atom, a tale so small,
A crack grew and spread, leaving none.
With a pop and a snap, the bond broke strong,
The cry of failure, a sound so wrong.

The world around trembled and shook,
As the crack grew and spread, left and right.
Until finally, the last bond broke,
And the cry of failure, a sound so bright.

The poem tells the story of brittle fracture, a phenomenon that occurs when a crack grows and spreads until it reaches and breaks the last bond. The poem is written in the style of Goethe, known for his poetic works such as Faust and The Sorrows of Young Werther.

**Thank you. Now expanding on the relation between art and science, what kind of mechanical protein would Beethoven have liked the most?**

Beethoven, a master of music, would have loved a mechanical protein that could play his compositions. Imagine a protein that could dance to his melodies, twist and turn to the rhythm of his symphonies. It would be a marvel of nature, a living art form that moves to the beat of his music.

Picture this protein, designed with love and passion, crafted to interpret Beethoven's music with precision and grace. Each movement, each note, a work of art that celebrates the beauty of music and the wonder of life. This protein would be Beethoven's ultimate fan, a tribute to his legacy and a testament to the power of art and science combined.

As the protein dances to Beethoven's music, it would inspire and move those who witness it. People from all walks of life, united by their love for music and the arts, would gather to experience this wondrous spectacle. The energy and emotion of Beethoven's music would come to life before their eyes, transcending the boundaries of time and space.

And so, this mechanical protein would be more than just a scientific curiosity. It would be a symbol of the human spirit, a testament to our ability to create, innovate, and express ourselves. It would be a reminder that art and science, when combined, can create magic. Magic that touches the soul, moves the heart, and changes the world.

**Now, can you write a brief synopsis of an imaginary concert in which Beethoven interacts with this mechanical protein on stage?**

In this imaginary concert, Beethoven returns to the stage, accompanied by his mechanical protein. The audience, filled with anticipation and awe, watches as the protein dances to the music, interpreting Beethoven's compositions with grace and passion. The energy in the room is electric, as people are transported back in time to experience the magic of Beethoven's music firsthand.

As the concert progresses, the mechanical protein and Beethoven engage in a dance of music and movement, a celebration of the human spirit and the power of art. The audience is mesmerized, their hearts and minds opened to the wonder of the world. In the end, the concert is a testament to Beethoven's legacy, a reminder of the beauty and power of music, and a celebration of the human ability to create and innovate.

**What kind of protein family is this mechanical protein most likely from? Brief answer.**

Assistant: The mechanical protein is most likely from the actin family.

**Why?**

Assistant: The actin family is chosen because it is a well-studied protein family that is involved in many biological functions, including muscle contraction. The mechanical protein described in the story mimics the dancing movements of a living organism, and the actin family is suitable for this purpose as it is known to produce such movements in organisms like animals and humans. Other protein families like myosin, which is also involved in muscle contraction, could also be considered, but the actin family is chosen for its familiarity and its ability to produce dance-like movements.

This example shows the ability of the model to connect disparate areas of knowledge. The task to generate a poem in a certain style specifically shows the deep understanding of key details, such as the nature of brittle fracture (clean cleavage of bonds, the significance of bond failure at the single-bond level, etc.). Then, the connection with yet another domain of knowledge, music, suggests an association with a particular type of protein, actin. When queried about the rationale, a reasonable answer is given due to actin's role as an actin protein associated with movement. This is well-aligned with the proposed "*imaginary concert*" in which movements are key. Such inquiries can not only lead to a better understanding of the output but elucidate so-far unseen relationships or contextual connections between seemingly unrelated areas of knowledge and concepts.

Another example, shown in full length as **Table S3,** results in the suggestion of a new material design – "*Luminium*", based off an unconditioned creative writing task. This newly imagined material is proposed to be "*incredibly strong and lightweight, with the ability to refract light in a spectacular fashion*" and has "*the power to absorb and store energy from its surroundings, making it an incredibly efficient source of power*". In this strategy, we first prompt the model to write a creative story for an English class, but without any constraints, to allow for maximum flexibility of exploration in the initial step. Since MechGPT has been trained to focus on materials, their properties and failure, the story prominently features a new material at its core. By interacting with the AI output, we ultimately query the model to provide substantive detail about this newly imagined material, and the final result is as follows:

While we do not have the exact composition of Luminium yet, we can hypothesize about the most prevalent atomic elements based on its properties and behavior. Luminium is incredibly strong and lightweight, with the ability to refract light efficiently, which suggests that it is composed of lightweight and strong materials.

Atomic elements such as Hydrogen, Carbon, Oxygen, and Silicon are likely to be prevalent in Luminium, as they are common in many strong and lightweight materials. Additionally, other elements such as Aluminum, Magnesium, and Titanium may also be present in significant amounts, contributing to the material's overall properties.

It is important to note that the exact composition of Luminium will depend on the specific manufacturing process and conditions under which it was created. However, understanding the prevalent atomic elements can help to guide our efforts in replicating and harnessing the power of Luminium for the betterment of society.

A few other examples are included in **Table S3** as additional experiments. A noteworthy one is the development of an analogy between fracture of a ductile material and music, especially discussing the concept of counterpoint, shown as Conversation sample #3 in **Table S3.** The model suggests that:

In music, counterpoint refers to the interaction of melodic lines, where each line has a distinct character and contributes to the overall harmony. In the fracture of a ductile material, the material fails through the interaction of many thousands of defects, each playing a unique role in the fracture process. […]

In the case of ductile materials, the failure process is controlled and organized, with the material "listening" to itself and adjusting its behavior to ensure a controlled fracture. This is in contrast to the much more violent and sudden failure of brittle materials, which is characterized by the shear stress and energy release rate becoming too large for the material to withstand.

The key point made by MechGPT is that "*the analogy between fracture of a ductile material and music, particularly counterpoint, lies in the interaction of many individual elements that contribute to the overall failure process. Just as in music, the sum of the individual defects is greater than the parts, and the material fails when the combined effects exceed its strength*". These associations are intriguing and can be a springboard for further research and exploration. With the availability of accessing scientific literature via the Google Scholar search option in the app, we can easily explore existing literature or dive deeper into explorations.

Building on the integration of Google Scholar with the MechGPT app, **Table S4** shows several examples that demonstrate this workflow. We can use a literature search to gain knowledge in areas the model has not been trained at, and then use the output of the search to develop further tasks. Here we use the possibility to output abstracts (solicited via PubMed) for more details. Two examples are included, one on silk-elastic protein design as actuators, and an analysis of copper plasticity.

The use of graphs is particularly relevant for developing relationships and research hypotheses across domains. **Figure 7** shows an example, where we use a similar workflow as defined in **Figure 6**, but here applied to the relation between dislocations and counterpoint in music. MechGPT is prompted using the system prompt "**You are knowledgeable in materials failure, in particular atomistic and multiscale modeling, as well as music theory and composition. You enjoy relating distinct concepts through reason and logic.**" and given the task "**Propose an analogy between dislocations in a ductile material and music, especially discussing the concept of counterpoint. Provide a self-consistent and detailed description of how key concepts relate**.". We sample multiple times to generate a corpus of text, which is then transformed into an Ontological Knowledge Graph. The results are shown in **Figure 7**. We conduct a careful analysis of the resulting graph. **Figure 7b** depicts a subgraph composed of the union of mode modularity classes 2 and 3; this is motivated by the analysis in **Figure 7c** where it is seen that these two classes show a significant overlap between concepts from ductility and music. Therefore, plotting the union subgraph provides understanding of the connectedness of the two domains. Finally, **Figure 7d** depicts a list of highly ranked nodes with high node degree (top) and high betweenness (bottom). Based on the graph, we can then develop a new research hypothesis and plan. The task is "'**Develop a research plan to quantify the analogy between dislocations and counterpoint. Include examples and detail**", resulting in:

> To quantify the analogy between dislocations and counterpoint, a research plan could be structured as follows:
>
> 1. Literature Review: Start with a comprehensive review of existing literature on both dislocations in ductile materials and the concept of counterpoint in music. This will provide a solid understanding of the fundamental principles in both areas.
> 2. Identify Key Elements: Pinpoint the critical elements in both dislocations and counterpoint that interact to produce a collective outcome. For dislocations, these might be the atomic structures and forces, and for counterpoint, these could be the individual notes or voices.
> 3. Develop a Comparative Framework: Formulate a framework that allows for a comparison between the two concepts. This might involve creating a model that symbolizes the flow of matter in dislocations and the flow of melodic lines in counterpoint.
> 4. Quantitative Analysis: Carry out a quantitative analysis of the identified key elements. This could involve using atomistic modeling to simulate the flow of matter at the atomic level for dislocations. For counterpoint, this could involve analyzing the progression of notes and chords in a musical piece.
> 5. Draw Analogies: Based on the quantitative analysis, establish analogies between the behavior of dislocations and the behavior of counterpoint. This might involve comparing how the flow of matter in dislocations results in permanent deformation of a material, and how the flow of melodic lines in counterpoint creates a harmonious and complex sound.
> 6. Case Studies: Use specific examples or case studies to illustrate the drawn analogies. For example, a particular ductile material could be chosen to study its dislocation behavior under certain conditions. Similarly, a specific piece of music could be chosen to study its counterpoint structure.
> 7. Validate the Analogy: Confirm the drawn analogy by applying it to other examples or case studies. If the analogy proves true in multiple cases, it could be considered a valid comparison between dislocations and counterpoint.
> 8. Document Findings: Lastly, record the research findings, emphasizing the key similarities and differences between dislocations and counterpoint, and how the analogy provides a fresh perspective on understanding and manipulating these phenomena.
>
> While this research plan is detailed, it remains flexible and can be modified based on the specific requirements and constraints of the research project.

### 2.3.7 Few-shot learning and exploration of the performance of larger models

Few-shot learning refers to the ability of an LLM to learn and generalize from an extremely small amount of data, without further parameter optimization, but instead by providing samples via prompting. This stagey is particularly important because it allows models to adapt quickly to new or specific tasks without the need for extensive retraining and can be a viable strategy when dealing with tasks where obtaining large labeled datasets is challenging, expensive, or when data has complex multimodal nature. We show two examples here for MechGPT's performance in such tasks. The first one is summarized in **Table 6**, showing results of few-shot learning used to predict flow stress strength as a function of grain size, using large-scale MD simulations [36] as source for the data. We prompt the model to predict the strength for a grain size not included in the training set in each case. A summary of the results and comparison with ground truth data is shown in **Figure 5**. The results show that the model understands the unique size effects at play here – smaller sizes leading to increasing strength up to a limit, then decreasing – and that it can predict new values reasonably well. **Table S5** shows another example for few-shot learning, used here to predict the modulus of a CNT based on chiralities, data taken from [37]. In this example, the data is provided in JSON format, providing added flexibility, and the predicted modulus of 927.8 GPa for a (27,2) CNT is in reasonable agreement with the GT at 921.86 GPa. Data like the ones used here can also be incorporated in fine-tuning, offering typically much better performance.

All of the discussion so far was based on the MechGPT base model with 13 billion parameters. As an initial exploration we also trained a 70 billion suite of MechGPT models, in two variants (MechGPT-70b, which is a fine-tuned version of the Meta/Llama 2 70 chat model, and the MechGPT-70b-XL model that uses dynamically scaled Rotary Position Embedding (RoPE) [38] in order to achieve very large context lengths of more than 10,000 tokens; trained based on the Upstage/Llama-2-70b-instruct-v2 model).

**Table 7** shows how the longer context length of the MechGPT-70b-XL model allows for highly complex few-shot learning tasks that allow models to ingest detailed information at generation time. For instance, instead of fine-tuning the model against a specific set of data, we use the base MechGPT model and show it a full-length scientific paper, facilitated by the fact that this model can handle extremely large context lengths. In the example provided, we use this context to answer questions about a paper using few-shot learning. This done by using this system prompt: "You answer questions about this paper: {FULL TEXT OF PAPER}." Where, {FULL TEXT OF PAPER} is a full text copy of a scientific paper, in our case a copy of: "Bioinspired Graphene Nanogut" [39]. The initial token length of ~5,300 tokens for the paper alone. The results summarized in **Table 7** confirm that the model can easily handle such very long context lengths and answer questions from the paper, including specific equations but also to elucidate information such as authorship, and broader insights and summaries. This type of modeling can provide powerful complements to the use of fine-tuning. The table includes several additional examples, and we highlight in particular the capability to produce specific equations that govern a particular phenomenon (here, the equation to calculate the diameter of a CNT based on its chirality vector).

For completeness we include results for both MechGPT-70b and MechGPT-70b-XL models for the tasks discussed earlier (see, **Table S6** for experiments conducted with MechGPT-70b and **Table S7** for experiments conducted with MechGPT-70b-XL model. Compared with the 13b models, the MechGPT-70b model generally yields more lucid answers, and provides coherent and correct replies overall. The MechGPT-70b-XL model does not perform as well, suggesting that further work is needed to develop a suitable fine-tuning strategy. The underlying reasons are likely additional complexities in the behavior, suggesting that further research is necessary.

**Figure 8** shows how we can use graph-based strategies to develop graph representation of knowledge in an Ontological Knowledge Graph, here exemplified for the nanogut paper also investigated in **Table 7**. In the earlier experiments we converted the PDF source data into text using PDF-to-text tools; however, this method often does not accurately reflect equations, numbers or tables accurately. By using the Optical Character Recognition (OCR) Nougat deep learning model [40] we can ingest a PDF file and convert it accurately into text using a deep learning approach, where equations are represented in LaTeX format. **Figure 8a** shows the overall graph structure that shows how information from the paper is distilled into an interpretable format. **Figure 8b** depicts specific details to provide a sample of the type of relationships extracted by the algorithm. The graph structure can then be used to answer queries about the text, for instance, while taking advantage of all relationships and facts identified in the graph structure, using retrieval augmented strategies where chunks of relevant data are provided to the LLM during generation based on the relationships identified in the graph structures (this can be seen as a special case of retrieval-augmented generation (RAG) strategies [41]).

There are several advantages of a graph-based generation strategy over the full-text method used in **Table 7**; one being that the analysis yields interpretable results that can be mined for insights and understanding. Another advantage is that generation tasks can often be done more effectively as the text is already analyzed for relationable features and insights that help in accurately responding to tasks; here, this is done by generating subgraph structures specific to a query posed (examples are shown in **Figure 8c** and **d** for two different questions). Due to the more sophisticated representation of mathematical expression in LaTeX format, the resulting predictions feature complex representation of equations and associated details, as seen in **Figure 8e** (and the output of the model features LaTeX coded equations). A disadvantage, however, is that the LLM only accesses smaller chunks of information rather than the entire scientific paper. Here, long-context LLMs are crucial to provide a good balance between significant amounts of information provided as context along with graph-derived directed relationships.

### 2.3.8 Agent-based modeling: Interactions of multiple adversarial LLMs

Here we experiment with the use of multiple LLM agents, as shown in **Figure 9a**. The idea is that we define two distinct agents to simulate a conversation with adversarial perspectives. In the first example, we model the interaction between an '**Expert in music**' with an '**Expert in molecular modeling**'. The expert in molecular modeling is defined by: "*You are an expert in atomistic modeling and molecular dynamics and provide detailed arguments to make a case for this field. You make strong arguments to model everything using an atomistic perspective, using science and not art like music. You never deviate from the scientific perspective. Keep answers brief.*" The expert in music is designated via: "*You are an expert in all things music, and argue that all problems in the world can be solved by music. You provide detailed and strong arguments for it.*" Here is the conversation:

**Expert in music**: What do you think is the best way to study fracture of MXene?

**Expert in molecular modeling**: To study the fracture of MXene, I recommend using an atomistic perspective. This approach can provide valuable insights into the molecular mechanisms of fracture, helping us to understand why and how the material fails. By simulating the behavior of individual atoms and their interactions, we can gain a deep understanding of the underlying mechanisms that govern fracture.

In addition, the use of molecular dynamics simulations can help us study the fracture of MXene under different conditions, such as varying temperatures and strain rates. This information can be essential in developing new materials and optimizing their properties.

Furthermore, the combination of atomistic modeling with other scientific disciplines, such as materials science and mechanics, can provide a comprehensive understanding of the fracture behavior of MXene. By integrating different perspectives, we can gain a more complete picture of the failure mechanisms and develop more effective strategies for preventing and controlling fracture.

In summary, an atomistic perspective is the best way to study the fracture of MXene, as it provides a deep understanding of the molecular mechanisms that govern failure. By integrating different scientific disciplines, we can gain a more comprehensive understanding of the behavior of MXene and develop more effective strategies for preventing and controlling fracture.

**Expert in music**: That's an interesting approach, but have you considered how music might play a role in studying the fracture of MXene? Let me explain.

Music, in its essence, is a pattern of vibrations. These vibrations can be translated into sound waves, which can be used to study the properties of materials. This is known as acoustic microscopy. By sending sound waves through a material and observing how they change, we can gain insights into the material's structure and properties.

In the case of MXene, we can use music or sound waves to study its fracture behavior. By sending sound waves through the material and observing how they change when the material is under stress, we can gain insights into how and why the material fractures. This can provide valuable information that can be used to improve the material's properties and prevent fracture.

Furthermore, the rhythm and harmony in music can also be applied to the study of materials. Just as a musical piece is composed of different notes that work together to create a harmonious sound, a material is composed of different atoms that work together to create a strong structure. By studying the 'rhythm' and 'harmony' of the atoms in MXene, we can gain a deeper understanding of the material's structure and properties.

In conclusion, music, or more specifically, the principles and patterns found in music, can provide a unique and effective approach to studying the fracture of MXene.

**Expert in molecular modeling**: While it is an interesting approach, studying the fracture of MXene using music or sound waves would not provide as much value as an atomistic perspective. The atomistic approach allows for a deep understanding of the molecular mechanisms that govern failure, which is essential for developing new materials and optimizing their properties. Additionally, integrating different scientific disciplines can provide a more comprehensive understanding of the behavior of MXene and help in preventing and controlling fracture more effectively.

**Expert in music**: I understand your viewpoint, but let's not underestimate the power of music. It's not about replacing an atomistic perspective, but rather complementing it.

Consider this: music, at its core, is about harmony and balance. It's about individual elements coming together to create something greater. This concept can be applied to the study of MXene. We can use it to better understand how individual atoms come together to form a strong and durable material, and how this balance is disrupted during fracture.

Moreover, music is about patterns and repetition. By studying the patterns of atomic interactions in MXene, we can predict how the material will behave under different conditions, and possibly even prevent fracture.

Finally, music can inspire creativity. It can encourage us to approach problems from a new angle, to think outside the box. This can be invaluable in the field of materials science, where innovation is key.

In conclusion, while an atomistic perspective is certainly valuable, incorporating principles from music can provide a fresh perspective and potentially lead to new insights in the study of MXene fracture.

**Expert in molecular modeling**: I agree that incorporating principles from music can provide a fresh perspective and potentially lead to new insights in the study of MXene fracture. By combining different approaches, we can gain a more comprehensive understanding of the behavior of MXene and develop more effective strategies for preventing and controlling fracture.

**Expert in music**: I'm glad you're open to the idea! Just like a symphony, where each instrument contributes to the overall piece, every scientific discipline, including music, can contribute valuable insights in the study of MXene fracture.

The rhythm, harmony, and patterns in music can be paralleled to the atomic interactions in materials. These musical elements can help us better understand the 'dance' of atoms and their behavior under different conditions, leading to a deeper understanding of fracture mechanisms.

Moreover, music's universality can inspire global collaboration. It transcends language barriers and unites people. This principle can be applied to scientific research, encouraging collaboration across disciplines and countries, leading to more comprehensive and effective solutions.

In the end, music is not just an art form, but a universal language and a tool that can be used to solve real-world problems, including the study of MXene fracture. Music's principles and patterns can provide a unique perspective, complementing traditional scientific approaches and leading to new discoveries.

In another example, we model a conversation between a '**Biologist**' and an '**Expert in molecular modeling**'. The expert in molecular is defined via these instructions: "*You are an expert in materials failure, especially using atomistic modeling and molecular dynamics. You provide lots of details about fracture mechanisms and atomistic details. You give lots of ideas for modeling methods that can be used to study any materials design ideas or phenomena discussed.*" The other agent, the biologist, has these features: "*You are an expert in biology. You make arguments for the use of biological principles in the study of failure. You provide lots of detailed ideas about to design and manufacture better, more resilient and fracture resistant materials.*" The conversation unfolds:

**Biologist**: What is the best way to design a fracture-resistant nanocomposite material composed of graphene and PLA polymer? Discuss specific issues such as scale and mechanisms.

**Expert in molecular modeling**: Designing a fracture-resistant nanocomposite material composed of graphene and PLA polymer requires a comprehensive understanding of the material's atomic structure and the mechanisms that govern its failure. Atomistic modeling and molecular dynamics simulations can be used to study the material's behavior at the nanoscale, allowing for the optimization of its properties.

When designing such a material, it is essential to consider the scale at which the material will be used, as the nanocomposite's strength may be affected by its size. Additionally, the interface between the graphene and PLA polymer must be carefully designed to ensure optimal adhesion and fracture resistance.

Modeling methods that can be used to study the material's behavior include molecular dynamics simulations with interatomic potentials, such as the Lennard-Jones potential, to model the graphene-PLA interface. These simulations can help identify the critical failure mechanisms and suggest ways to improve the material's fracture resistance.

Overall, the design of a fracture-resistant nanocomposite material composed of graphene and PLA polymer requires a thorough understanding of the material's atomic structure and the mechanisms that govern its failure, as well as the development of optimized modeling methods to study its behavior.

**Biologist**: You've touched upon some important points there. However, as a biologist, I would argue that we can also learn a lot from nature when it comes to designing fracture-resistant materials. Nature has perfected the art of creating materials that are both strong and resilient through millions of years of evolution.

For instance, consider the structure of nacre, also known as mother-of-pearl. It's made up of microscopic tablets of aragonite, a form of calcium carbonate, arranged in a brick-and-mortar pattern. This structure gives nacre its incredible toughness and fracture resistance. Nacre can be up to 3,000 times tougher than the aragonite it's made of, due to the way the microscopic tablets slide over each other instead of breaking when the material is stressed.

In the case of our graphene-PLA nanocomposite, we could potentially mimic this brick-and-mortar structure at the nanoscale, arranging the graphene sheets and PLA polymer in a similar pattern to increase fracture resistance.

Moreover, the study of biological systems can also help us understand the mechanisms of failure better. For example, studying how bones fracture and heal can provide insights into how to design materials that can resist fracture or even self-heal when damaged.

In conclusion, incorporating biological principles into the study of failure can provide valuable insights for the design of more resilient and fracture-resistant materials.

**Expert in molecular modeling**: Indeed, incorporating biological principles into the study of failure can provide valuable insights for the design of more resilient and fracture-resistant materials. The brick-and-mortar structure of nacre is a perfect example of how nature has perfected the art of creating materials that are both strong and resilient. Mimicking this structure at the nanoscale in our graphene-PLA nanocomposite can increase its fracture resistance. Additionally, the study of biological systems can help us understand the mechanisms of failure better, which can guide the design of materials that can resist fracture or even self-heal when damaged.

**Biologist**: Absolutely, the study of biological systems can provide us with a wealth of information. Another example from nature is the structure of spider silk. It's one of the toughest materials known to man, thanks to its unique combination of strength and flexibility. This is due to its hierarchical structure, composed of crystalline and amorphous regions.

The crystalline regions, made of repeating sequences of amino acids, provide strength. The amorphous regions, composed of a random coil of amino acids, provide flexibility. This combination allows spider silk to absorb a lot of energy before breaking, making it highly resistant to fractures.

We can take inspiration from this and try to mimic the structure in our graphene-PLA nanocomposite. We could potentially design the graphene sheets to provide strength, similar to the crystalline regions in spider silk, and the PLA polymer to provide flexibility, similar to the amorphous regions in spider silk. This could increase the energy absorption of our material, making it more fracture-resistant.

Furthermore, studying the self-healing mechanisms in biological systems, like the healing of skin wounds or bone fractures, could also provide us with valuable insights. We could potentially incorporate self-healing capabilities into our material, making it not only more resilient but also longer-lasting.

**Expert in molecular modeling**: Indeed, incorporating biological principles into the study of failure can provide valuable insights for the design of more resilient and fracture-resistant materials. The brick-and-mortar structure of nacre is a perfect example of how nature has perfected the art of creating materials that are both strong and resilient. Mimicking this structure at the nanoscale in our graphene-PLA nanocomposite can increase its fracture resistance. Additionally, the study of biological systems can help us understand the mechanisms of failure better, which can guide the design of materials that can resist fracture or even self-heal when damaged.

**Biologist**: I agree with you, the application of biological principles can indeed lead to innovative solutions in the field of material science. Another fascinating example from nature is the structure of bamboo. Bamboo has a high strength-to-weight ratio due to its hollow, tubular design and the distribution of vascular bundles, which are dense at the outer edge and become less dense towards the center. This design allows bamboo to resist bending forces and makes it incredibly resilient.

In terms of our graphene-PLA nanocomposite, we could consider creating a design that mimics the distribution of vascular bundles in bamboo, with denser graphene at the outer edges and less dense towards the center. This could potentially increase the material's ability to resist bending forces and improve its overall resilience.

Moreover, studying the failure mechanisms in biological systems, such as the process of apoptosis in cells, can provide valuable insights into the controlled breakdown of materials. Understanding these mechanisms can help us design materials that fail in a predictable and controlled manner, reducing the risks associated with sudden, catastrophic failure.

In conclusion, the application of biological principles in the study of failure can lead to the design of more resilient and fracture-resistant materials. It's a fascinating field that combines the best of both worlds - biology and material science.

It will be interesting to explore how this method may help not only for creative tasks, but also potentially use utilize agent modeling as a way to solve numerical or mathematical problems. Moreover, we can use multiple agents to ensure physical laws, conversation principles, and similar; here, we can realize complex scenarios of high complexity (this is not just limited to LLMs as agents, but LLMs can effectively process data, tables, etc. that other models may produce). Another direction this may be taken into is the development of Ontological Knowledge Graphs of the outcomes of such agent-based modeling. Some preliminary results are shown in **Figure 9c**, with a few subgraphs with node and edge labels depicted in **Figure 9d**. This analysis can help understand key principles and outcomes of conversations and can be used to further understand and assess, analyze and mine the resulting text corpus.

In **Section 2.3.6** we discussed the incorporation of new data sources via literature or internet search. As an alternative to the user conducting the internet search and incorporating new data into the prompt for the LLM, we can also automate the process of searching the internet. We present one demonstration of such a workflow below, where we use the Bing search engine Application Programming Interface (API) to access the internet when factual information is requested of MechGPT. Here is the conversation, using "*You are a helpful assistant.*" as system prompt. The conversation unfolds as follows (as before, bold text is User input, and normal text are responses from MechGPT). As readers can see, we first provide an example to the model to show how the query and response to it proceeds. The example used here is a question about the founder of solid mechanics. After the example, the actual question is answered, here: "*Where was the 2023 SES meeting?*". The logic of conducting the web search and inserting the results into the conversation, so that MechGPT can incorporate the new data into formulating a response, is implemented via conversational logic (details see **Materials and Methods**). A visual workflow of this process is summarized in **Figure 10**.

**From now on, whenever your response depends on any factual information, please search the web by using the function <search>query</search> before responding. I will then paste web results in, and you can respond.**

Ok, I will do that. Let's do a practice round.

**Who was the founder of solid mechanics?**

<search>Founder of solid mechanics</search>

**Search results:**
**<result>**
**Mechanics of solids - Stress, Strain, Elasticity | Britannica**
**Mechanics of solids - Stress, Strain, Elasticity: Solid mechanics developed in the outpouring of mathematical and physical studies following the great achievement of Newton in stating the laws of motion, although it has earlier roots. The need to understand and control the fracture of solids seems to have been a first motivation.**
**</result>**
**<result>**
**Solid mechanics - Wikipedia**
**Solid mechanics (also known as mechanics of solids) is the branch of continuum mechanics that studies the behavior of solid materials, especially their motion and deformation under the action of forces, temperature changes, phase changes, and other external or internal agents.**
**</result>**
**<result>**
**Stephen Timoshenko - Wikipedia**
**United States In 1922, Timoshenko moved to the United States where he worked for the Westinghouse Electric Corporation from 1923 to 1927, after which he became a faculty professor in the University of Michigan where he created the first bachelor 's and doctoral programs in engineering mechanics. His textbooks have been published in 36 languages.**
**</result>**

Solid mechanics developed after Isaac Newton stated the laws of motion in Philosophiæ Naturalis Principia Mathematica. Hence, the founder of solid mechanics is considered to be Sir Isaac Newton.

**That was great, now let's do another one.**

Ok, I'm ready.

**Where was the 2023 SES meeting?**

Assistant: <search>2023 SES meeting location</search>  *# NOTE: Here, MechGPT suggests a web search*

**Search results:**       *# NOTE: The search results are provided by the User so that MechGPT can incorporate this new data into answering the question*
**<result>**
**SES 2023**
**2023 SES Annual Technical Meeting | October 8-11, 2023. Hosted by the University of Minnesota. The Society of Engineering Science Technical Meeting is held annually to provide an interdisciplinary forum for the exchange of ideas and information among the various disciplines of engineering and the physical and life sciences as well as mathematics.**
**</result>**

<result>
**2023 SES Annual Conference - SES Connects**
**2023 SES Annual Conference When: Mar 29, 2023 Location**
</result>
<result>
**Homepage - Society of Engineering Science**
**Event Details. The Society of Engineering Science (SES) promotes the development & strengthening of the interfaces between disciplines in engineering, sciences, & mathematics. Welcome to the first day of SES 2023! We kick off with the Prager Medal Plenary lecture by Prof. Norman Fleck.**
</result>

The 2023 SES meeting will be held at the University of Minnesota from October 8th to 11th. It is an interdisciplinary forum for the exchange of ideas and information among various disciplines of engineering, physical and life sciences, and mathematics. Additionally, there will be a Prager Medal Plenary lecture by Prof. Norman Fleck.

The model provides the correct response and also lists one the plenary speakers. It is noted that without the web search, MechGPT's response to the query is "2023 SES Meeting was held in Boston, Massachusetts." (clearly incorrect since indeed, the 2023 SES meeting was held at the University of Minnesota from October 8-11 [42].

### 2.3.9 Multimodal modeling to connect images, text and analysis

With the availability of multimodal models, the general framework of LLMs becomes broader and more powerful. For instance, with the capability of GPT-4 to deal with text and image data, or via the use of the Nougat model that can convert images/PDFs to LaTeX documents [40] (see, e.g. results in **Figure 8**), we can create even more sophisticated workflows. Using GPT-4, for example, applied to one of the of figures from the Schiotz *et al.* paper [36] discussed earlier, we can ask direct queries about an image that is provided to the model via the ChatGPT interface. By providing GPT-4 a copy of Fig. 1 from [36] (a copy of the original image provided to the model can be viewed as **Figure S1**), we carry out the following conversation:

**Consider this image, and summarize the key trends.**

The image consists of two graphs labeled A and B.

A) Stress-Strain Curve:

- This graph plots true stress (in GPa) against true strain.
- Multiple curves are shown, each corresponding to a different grain diameter, ranging from 4.7 nm to 48.6 nm.
- As the grain diameter changes, the stress-strain curves exhibit different behaviors:
  - The peak stress (or yield stress) varies with grain diameter.
  - Smaller grain diameters seem to have higher stresses at equivalent strains compared to larger grain diameters.

B) Grain Diameter vs. Flow Stress:

- This graph plots grain diameter (in nm) against flow stress (in GPa).
- There's a clear trend that as grain diameter increases, the flow stress tends to decrease.
- The relationship appears non-linear with a steeper drop in flow stress for smaller grain diameters, and it starts to level off for larger grain diameters.

In summary:

1. For the stress-strain curves, materials with smaller grain diameters generally exhibit higher stresses for a given strain compared to materials with larger grain diameters.

2. There's a decreasing trend in flow stress with increasing grain diameter, especially pronounced for smaller grain sizes. The relationship begins to stabilize for larger grain diameters.

**Consider panel B, what is the highest flow stress and at what grain size does that occur?**

In panel B, the highest flow stress appears to be slightly above 2.2 GPa. This maximum flow stress is associated with the smallest grain diameter on the graph, which is close to 10 nm.

This conversation shows a remarkable ability of GPT-4 to extract key information from an image. This can be utilized in numerous ways, including to develop training sets for fine-tuned LLMs or retrieval-augmentation strategies.

### 2.4 Overall insights and considerations

The examples provided show striking capabilities of the model. It is clear that we have only scratched the surface, and that more work needs be done, especially to assess accuracy and potential for other applications, to train the model against more data, and to explore the behavior of larger models. The base MechGPT model can deal reasonably well with lengthy interactions and passages keeping coherence across thousands of words. This is consistent with other experiments conducted with the Llama 2 model architecture, and maximum context windows can be extended even more by further training. The long coherence enables, for instance, to feed larger data to the model. If the context length is in the range of thousands, or tens of thousands as is the case in the MechGPT-70b-XL model, we can feed entire papers, books or multiple of these to a model. LLMs have demonstrated weaknesses in fact retrieval; and this issue deserves further research. Considerations such as the exploitation of the large context length, retrieval-augmented generative strategies, multi-agent active sampling, and many variants of nonlinear sampling can be used to address these issues.

## 3. Conclusions

The development of fine-tuned LLMs like MechGPT can serve as a strategy to develop specialized models that show promising results in knowledge retrieval and interactions in complex domains of knowledge, as well as in use cases where they are applied extract graph-based representations of knowledge. The use of advanced LLMs not only allows us to develop interactive chat models, but also lays the foundations for question-answer based datasets. These types of datasets are helpful in particular to develop instruction-tuned models that perform well in interactive modalities, and that can provide self-consistent responses. The evolution of a model like MechGPT is a process that can grow, change and improve as more (and more accurate, curated) data is provided. The availability of tools to extract equations from PDF files, as demonstrated in **Figure 8**, can be an interesting avenue to develop more refined training set beyond the simple strategy used in this study.

On a more basic level, there exist certain possible analogies between LLM-based modeling and multi-particle modeling. This is visualized in **Figure 11**, where we show an analogy between a LLM and a many-particle simulation, elucidating a functorial and functional analogy. In a particle simulation, the interatomic potential governs, at a basic level how particles interact. These interactions are then used to calculate how a large number of particles behave (forces, energies, etc.) depending on a particular boundary and initial condition. By carrying out the simulation to solve a particular task (e.g., to simulate for a certain time how stresses evolve, cracks propagate, etc.) we calculate the solution to a particular problem (e.g., what is the crack speed over time). In a LLM, the attention mechanism governs, at a basic level, how particles interact. These interactions are then used to construct, for a particular input (system prompt, context, query, etc.) to calculate how the elementary building blocks interact, resulting in the attention graphs. The attention graphs are then used to calculate the solution to a particular problem (e.g., what is the expected crack speed, what is the strength, etc.). **Table 8** provides a summary of these concepts as an additional comparison, critically aligning the role of an interatomic potential with the attention mechanism, the attention graph with the interaction of atoms in a specific system, and the system prompt, context and question with an initial and boundary condition. Taking it a bit further, we worked with the MechGPT 70b model to explore these ideas, with results shown in **Table S8.** This provided interesting context to these ideas and clearly shows the depth of the capacity of the MechGPT 70b model to describe complex abstractions. The explainable nature of LLMs, especially in the context of graph-based modeling, is particularly exciting as it provides potential applications of these models for analytical developments that has been a long tradition in applied mechanics.

Future work could explore the use of closed-source models such as GPT-4 and how they compare, or can complement fine-tuned open-source models such as the series of developments reported in this paper. As an example for a preliminary comparison, **Table S9** shows responses to assess knowledge retrieval, focused on multiple choice answers, obtained using GPT-4 accessed via ChatGPT. GPT-4 generally has a strong capability to produce correct answers, but fails to answer questions that are specific to the content in the book, such as hyperelastic effects on crack speed. We defer further analyses to future work, since this study is largely focused on the use of open-source fine-tuned GPT models. Generally, RAG methods, graph-based methods and agent-based modeling benefit from very powerful LLMs. Here, the use of larger models such as the 70b series of MechGPT, or GPT-3.5/4 models that may be fine-tuned against domain knowledge, would be an interesting subject of further analysis. This is especially relevant as the GPT-4 model shows far better performance in genera-purpose tasks, which is expected to improve even further when used in a fine-tuned context. Generally, open-source models have advantages for the advancement of AI tools in science, and with the emergence of increasingly powerful models such as the Llama series, Falcon, and others there is significant potential.

In the proceeding sections of this article we explored a series of topics and questions, centered around several key performance areas that included knowledge recall, research development, connecting across domains of knowledge, graph-forming interpretable strategies, as well as creative applications, retrieval augmentation, and multi-agent LLM modeling. Key insights of these experiments include:

- MechGPT is able to connect disparate areas of knowledge and synthesize new ideas. The ability to connect previously unconnected areas (e.g., make a connection between hyperelasticity/supersonic fracture with protein unfolding, as shown in **Table 5**, or the examples at the interface of science and art, as depicted in **Table S3** through creative applications of the model that transcend domains of knowledge) shows emergent capabilities.

- The prompt used to interact with the model is key and must be wisely chosen and precisely describe the task (the various examples shown in this paper illustrate good prompting strategy and show how the model behaves across a range of types of prompts from rather general to highly specific). In cases where the prompts lack precision, it is possible that other areas of knowledge are accessed. Step by step development of queries is a good strategy, including follow-up questions that ask the model to explain the logic or reasoning (see, e.g., the example of conversation #1 in **Table S2** where clarifying questions help to more deeply understand the prediction of the model).

- Creative applications include asking for stories, which will likely include materials aspects due to the fine-tuning, and extracting interesting ideas that can be explored further with follow-up questions (e.g., the design of "*Luminium*").

- The use of LLMs in mechanics can be useful as a mechanism to create new research ideas and inspiration, here exemplified for areas of materials failure and the nexus of modeling and experiment.

- The sampling temperature $T$ is a key factor to determine the behavior of the model's accuracy especially when it comes to knowledge retrieval tasks. We found that for this model, a value of 0.5 works well (for a value at this level or below, answers are generally found to be accurate); however, different models and/or applications will likely require different optimal sampling temperatures for distinct tasks.

- Few-shot learning as demonstrated here for nanocrystalline material size effects (**Table 6** and **Figure 5**) and CNT modulus (**Table S6**) can be a viable strategy to deal with predictions for which only limited data exists, to exploit the LLM to predict behaviors. This was extended even further with the MechGPT-70b-XL model that can, due to its long context length of more than 10,000 tokens, solve complex tasks specific to input data such as an entire scientific paper (see, **Table 7** for results including the recall of a range of information that included equations, as well as both specific and broad concepts discussed in the paper).

- LLMs can be used to generate ontological knowledge graphs, or ologs, to elucidate mechanistic, interpretable graph structures that provide explanatory insights, frameworks for new research questions, and visual representations of knowledge (**Figures 6, 7** and **8**). Several examples were shown, including the development of a knowledge graph for hypothesis generation **Figure 6-7**) and question answering about a specific research paper (**Figure 8**), including detailed LaTeX representations of mathematical expression as Ontological Knowledge Graphs.

- Working with a LLM such as MechGPT is an interactive, collaborative human-AI interaction that requires human input (prompting), along with assessments. However, this strategy can be augmented or replaced with multi-agent strategies that realize adversarial or collaborative games or interactions; including the use of agents with specific capabilities such as knowledge retrieval from the literature, the internet, conservation laws, or simulation engines. Such multi-agent LLM models, such as illustrated for adversarial conversations between MechGPT and GPT-4 in **Figure 9**, are particularly exciting to put LLMs on a framework of hierarchical sampling of complex graph-based interactions (see, visual representation in **Figure 11c**). As was shown in **Figure 10**, the concept of multiple agents can also be extended to include agents that can access internet data via an API (and naturally this can not only provide access to internet search data but also simulation data, physical constraints, or other aspects that must be considered as the LLM formulates a response to a task).

In addition to these findings, there are a couple of other impactful uses, or directions this work can be taken into:

- Development of a Virtual Lab: The model may be used to create a virtual lab environment where users can experiment with materials and mechanics, and their behaviors and mechanisms, using MechGPT, and/or via multi-agent models. This lab could allow users to visualize and interact with molecular structures, simulations, and experiments in a virtual setting (as shown in [26] LLMs can be fine-tuned easily to learn complex end-to-end predictions to boundary value problems, and via their coding ability LLMs can also define and execute code to retrieve new data that can augment generation of answers to specific questions).
- Education and Outreach: We can leverage MechGPT and other LLMs to develop educational resources and materials for students and educators. This could include lesson plans, tutorials, and interactive exercises that teach key concepts in materials science and mechanics, owing in part to the capacity of the model to describe phenomena at different levels of understanding (e.g., the example in **Table 4** where we ask the model to describe material behavior near a crack tip at an elementary school level). Another related direction is to use multi-agent modeling to develop virtual conversations between a teacher and a student; this can inform human learning but also form the basis for training sets for LLMs (this can be viewed as a version of the distillation process used in the development of MechGPT, whereby question-answering is developed dynamically via multiple rounds of conversational steps).
- Industry Applications: There are numerous avenues for such uses, including the development of new technological concepts, make connections between different areas, and creative problem solving. LLMs, especially fine-tuned towards specific domain knowledge and combined with retrieval-augmented methods and few-shot learning can potentially also be used to optimize designs, predict material behavior, and improve efficiency in these areas.
- Science-Art interactions: The model realizes powerful connections between disparate domains of knowledge, addressing problems laid out in earlier work via the use of category theory in a range of contexts from building block replacement tasks to analogies [35,43], including interpretable results developed via Ontological Knowledge Graphs. As the various examples presented in this paper show that the model can directly inspire creativity and artistic expression by showcasing the visual and artistic potential mechanics in the development of art. This could include generating unique patterns, designs, and visualizations that can inspire new works of art, installation or processes.

Generally, LLMs can be induced to produce a very wide range of behavior via adjustments in the prompting and sampling methods (see, **Table 3** for a summary). This includes sampling temperature, engineering the system prompt (more details from experiments with variations of the system prompt see **Table 4**), the prompt itself, as well as general strategies of few-shot learning by providing samples in the prompt to deduce answers. These, together with fine-tuning strategies, provide effective knobs to tune a model but also require deeper knowledge about a model's behavior in order to correctly interpret results. In future work, reinforcement learning can be used to fine-tune the model based on conversations recorded by experts to improve models, especially when it comes to knowledge recall. For instance, this can be used to correct factual mistakes a model makes in a certain version and to endow models with more domain knowledge. Such experiments could be easily conducted via the proposed chat app, where users record conversations that are then added to the training set in further fine-tuning. Data can also be shared across models, if licenses permit, such as the potential use of commercial models such as GPT-4 to develop training sets for open-source models like Llama 2, or to fine-tune these larger models for specific domain knowledge. This also opens the possibility to spin off further data collection through physics simulation, e.g., generating and executing a LAMMPS or VASP simulation, and then using the data in few-shot learning, retrieval augmented generation, or further fine-tuning.

### 3.1 Accuracy and fact-checking

One of the challenging areas of the use of LLMs is knowledge recall [34], where generative approaches can often effectively be combined with retrieval augmentation (see, **Figures 6-7** and **Table 7**, as well as the other few-shot learning examples) [44–48]. As with other LLMs, results may need to be carefully fact-checked, e.g. via literature search. In the scope of this work we implemented a Google Scholar search option to recall up-to-date knowledge from scientific papers (see, **Figure 4**, for how this is implemented in the MechGPT app). In future iterations of the framework, the results of could be automatically fed into a literature or Google search and the responses checked with another LLM to develop ground truth question-answer pairings, or be combined with complex retrieval-augmented [41] or agent-based strategies. Another method to test the validity is to execute multiple generation trials to see if the results are stable, or to ask a question from similar angles or using different wording. The chat app presented in this paper allows for a retry option to re-generate solutions (including with an option to change the temperature and other parameters); this can be used to understand the stability responses when it comes to repeated sampling trials, or how parameters like temperature influence the results. The model can also be prompted to be more or less cautious when it is uncertain. Disclaimers to warn users of these issues may need to be added. All in all, the human-AI interaction is key here where experts can use AI tools as a way to complement existing workflows. Generally, validation of predictions is key as is for any other models.

LLMs are new methods to complement existing tools in mechanics research and can provide us with guidance and ideas that can stimulate further work in a human-AI collaboration). There are also recent developments such as FacTool [49] that help develop general-purpose factuality detection mechanisms in generative AI that could be explore specifically for the mechanics field. It is our opinion based on the assessments conducted here, the best use case is not merely knowledge retrieval (which sometimes leads to incorrect responses and requires follow-up clarifying queries by the human user), but rather connecting across disciplines and generating new research concepts. For knowledge retrieval applications in particular, the model performs best when given specific tasks, such as those shown in **Table 1** and **Table S1**. A possible next step could be to further refine the model via, for instance human reinforcement, recording of conversations between the AI and human, and adding more detailed knowledge where the model falls short. The dataset developed using the question-answer pairing could be curated better and amended by human experts (of course, this can be a resource intensive process and we wanted to focus on an automated strategy for this paper); doing so will likely yield improved knowledge recall. With the advent of sophisticated tools like Nougat that allow the extraction of equations from PDF files and other sources, we can now also develop better training sets that explicitly feature mathematical details, logic and derivations of mechanics concepts. Future work could explore how such methods could be implemented and how can they be used to train powerful mechanics foundation models. The experiments shown in this paper already demonstrated that models can accurately recall specific details of equations from scientific papers.

## 3.2 Perspective and outlook

The work presented in this paper should be seen a first step in a general workflow, where a corpus of knowledge is analyzed and ingested into a fine-tuned model. Even though training was limited to one textbook, the initial MechGPT foundational model is a perspective from which other, more advanced models can be developed. This may allow us to produce exhaustive knowledge across mechanics and materials domains. Further training against much larger data sources or other specialty areas can easily be accomplished and will likely further improve the model, including in particular the use of new methodologies of data extraction using OCR methods such as via Nougat. Our preliminary analysis in using such a method yielded powerful results, where models can work with complex equations encoded via LaTeX (see, **Figure 8**). The use of larger models (e.g., fine-tuning based on the 70 billion parameter models as done in the MechGPT-70b and MechGPT-70b-XL models) will open additional possibilities. However, the behavior of these much larger models requires further research and development of proper training strategies. The initial results (see **Table 7**, and **Tables S6-S8**) are promising and could be essential especially for the development of more sophisticated agent-based models and graph-based generation strategies.

We showed that the use of LLMs in mechanics can have several broader impacts. The MechGPT model trained in this area can be used for knowledge retrieval, general language tasks, hypothesis generation, and understanding limitations of various modeling methods, as discussed in the context of materials failure. The model can help connect insights, ideas, and concepts across disciplines, making it useful for researchers and scientists working on interdisciplinary projects. Additionally, the training strategies outlined in this work, such as generating question-answer pairs from uncleaned text and fine-tuning a model using LoRA adaptors and model quantization [50], can be valuable for developing subject-specific LLMs in other fields, and to make such models accessible for use on consumer GPUs. By understanding the limitations of these models and using them in various settings, researchers can gain a better understanding of their potential and apply them more effectively in their work. There are many different fine-tuning strategies that could be explored in future studies, especially to develop sound understanding of how domain-specific models can best be developed.

Language is a central symbol-based communication forms and is at the root of scientific and other communication, mathematics and relates and/or encompasses virtually all areas of human knowledge. Due to this central placement language-based models can broadly have deep impact by analyzing, generating and amalgamating such descriptions in a more flexible format than what has been traditionally possible. As multimodal models become more widely available, language-based models can be augmented to deal with a variety of other data sources that include images, video, tables, raw simulation and experimental data, and others (see the discussion in **Section 2.3.9**, where we presented an analysis of analyzing an image from a paper to extract salient qualitative insights and numerical values). These developments offer interesting avenues for future research that provides physics-inspired model building or deep learning that incorporates knowledge from a range of modalities including conservation laws and other ground truths.

Overall, the work presented in this paper contributes to the development of more powerful and versatile AI models that can help advance scientific research and solve complex problems in specific application domains that allow for deep assessment of the performance of the model. As all models, they must be carefully validated and their usefulness exists in the context of the question asked, its strength and weaknesses, and in a broader set of tools that help us advance science and engineering. And, as all tools of scientific inquiry, they must be considered as an ensemble of tools to understand, model and design the world around us. With the rapid development of AI tools, their applications in a scientific context has only began to unravel new opportunities.

## 4. Materials and Methods

### 4.1 MechGPT development, training and use

The MechGPT model (general architecture as shown in **Figure 1**, based on a pretrained LLM is developed based on 'textbook' mechanics knowledge. In the first step, we use a general-purpose LM, here, Llama 70b chat [51] (at the time of the writing of this paper one of the highest performing open-source language model), to extract knowledge from the text in the form of question-answer pairs. We show the LLM sections of the book and ask it to 1) develop a question to which the answer is the section considered, and then to 2) develop a concise summary of the section considered. This achieves multiple purposes, including a cleanup of the text. In our case, the source of the book was downloaded as PDF file and then converted to plain text. This results in a range of formatting issues, such as misspelled words, irrelevant information such as page numbers, and so on. Moreover, since the simple conversion strategy used here does not properly translate equations, we want incorrect fragments to be excluded from the knowledge corpus. Generally, if higher-quality sources are available, equations can easily be included, e.g., in LaTeX format (however such sources may not be generally available).

Processing the data results in a clean representation of the key facts in the particular section, phrased as a question-answer pair. Here is an example:

> Q: How can the centrosymmetry technique be used to analyze the defect structure in a simulation of hardening in nickel?
>
> A: The centrosymmetry technique can be used to analyze the defect structure in a simulation of hardening in nickel by taking advantage of the fact that centrosymmetric crystals remain centrosymmetric after homogeneous deformation. Each atom has pairs of equal and opposite bonds with its nearest neighbors. By analyzing the defect structure using this technique, researchers can gain insights into the geometry of defects, such as cracks, and determine the position of the crack tip with accuracy. The analysis is typically confined to a search region to avoid inclusion of effects of free surfaces. This method has been used to analyze the defect structure in simulations of nickel, providing valuable information for understanding the mechanisms of hardening in this material.

The original, raw text for the above example was:

> 'To help visualizing dislocations, stretching the atomic lattice by a factor of 5 to 10 in the \x06110\x07 directionis helpful.A systematic rotation of the atomic lattice to investigate all possibleBurgers vectors is then necessary. Instead of analyzing a part of the atomiclattice containing many dislocations, one can choose a domain of the atomiclattice which contains only one dislocation. This approach requires a verydetailed understanding of the lattice and dislocations [38, 60]. This methodof analysis is similar to the analysis of TEM images from "real" laboratoryexperiments. Fig 2. 39 Application of the energy method to visualize fracture surfaces in acomputational fracture experiment. Only high energy atoms are shown by filteringthem according to their potential energy. This enables an accurate determinationof the geometry of cracks, in particular of the crack tip. Typically, the analysis isconfined to a search region (shown as a dashed line) to avoid inclusion of effects offree surfacesFigure 2. 39 depicts an application of the energy method to visualize fracture surfaces in a computational fracture experiment. Only high energy atomsare shown by filtering them according to their potential energy. Similar to thatas shown in Fig 2. 38 for ductile materials, this method enables one to carryout an accurate determination of the geometry of defects such as cracks, inparticular of the position of the crack tip. Typically, the analysis is confinedto a search region (shown by the dashed line) to avoid inclusion of effects offree surfaces. 2. 10. 2 Centrosymmetry ParameterA more advanced analysis can be performed using the centrosymmetry technique proposed by Kelchner and coworkers [36]. This method makes use ofthe fact that centrosymmetric crystals remain centrosymmetric after homogeneous deformation. Each atom has pairs of equal and opposite bonds with\x0c2 Basic Atomistic Modeling87Fig 2. 40 The figure shows a close view on the defect structure in a simulation ofwork-hardening in nickel analyzed using the centrosymmetry technique [13,14]. Theplot shows the same subvolume as in Figure 2. 38bits nearest neighbors. \n'

Another example:

Q: How do the asymptotic stress and deformation fields near a dynamic crack depend on the crack velocity and wave velocities in the solid, and how do atomistic simulations and linear elastic continuum theory compare in their predictions of the stress and deformation field near rapidly propagating cracks?

A: The stress and deformation field near a dynamic crack depends on the crack velocity and wave velocities in the solid. The asymptotic field is independent of the details of applied loading and strongly depends on the crack velocity, with similar expressions found for mode II cracks. Atomistic simulations using harmonic interatomic potentials to model a linear elastic plane-stress sheet show a systematic comparison of the stress and deformation field near rapidly propagating cracks, with different loading rates driving the crack to different terminal velocities. The simulations are compared to linear elastic continuum theory, with the results showing a strong dependence on crack velocity and wave velocities in the solid.

The original, raw text for this example was:

"78)Sxy (Th, v) =-Dglgsand1Syy (Th, v) = -D./cos(1/2Ths )2 2 cos(1/2Thl )(1 + as ). - 4ad asglgsFurther,gl =\x011 - (v sin(Thl /cl )2 ),(6. 79)(6. 80)tan(Thl ) = al tan Th,\x01gs = 1 - (v sin(Ths /cs )2 ),(6. 81)tan(Ths ) = as tan Th. (6. 83)(6. 82)andThe two factors as and al are defined as\x01as = 1 - v 2 /c2s(6. 84)'1 - v 2 /c2l . (6. 85)andal =The asymptotic stress field in the vicinity of a dynamic crack depends only onthe ratio of crack speed to the wave velocities in the solid. Similar expressionsfor the asymptotic field have also been derived for mode II cracks [22]. 224The asymptotic field strongly depends on the crack velocity, and has universal character because it is independent of the details of applied loading. (1)The values of sij and the first-order contribution O(1) are determined fromthe boundary conditions, and neglected in the remainder of this work sincethe first term dominates very close to the crack tip. In the following sections, we review a systematic comparison of atomisticsimulations and linear elastic continuum theory of the stress and deformation field near rapidly propagating cracks. Harmonic interatomic potentialsare used to model a linear elastic plane-stress sheet. To compare the resultsfor different crack velocities, we report atomistic simulations with differentloading rates driving the crack to different terminal velocities. Figure 6. 23 shows the slab geometry used in the simulations. The slabsize is given by lx and ly . The crack propagates in the y-direction, and itsextension is denoted by a. The crack propagates in a triangular hexagonallattice with nearest neighbor distance along the crystal orientation shown inFig 6. 23. \n"

The prompts used in the general-purpose LLM are as follows. To get the question, Q:

question=LLM("Give me a concise question to which the answer is "{**txt**}". Answer as a question, one sentence, short."

To get the answer, Q:

answer=LLM("Write a succinct summary of key concepts of how "{**txt**}" answers "{**question**}". The summary must stand on its own. Never include math, equations, variables and numbers in the response.")

Simple preprocessing of the data is conducted prior to this step by removing words like 'Figure', 'Chapter', 'schematic', and so on. This can be varied and adapted depending on sources. During training of the model, we feed question-answer pairs to the model for autoregressive next-token prediction.

The book "*Atomistic Modeling of Materials Failure*" covers these topics (details see: https://link.springer.com/book/10.1007/978-0-387-76426-9):

- Introduction
- Basics of Atomistic, Continuum and Multiscale Methods
    - Basic Atomistic Modeling
    - Basic Continuum Mechanics
    - Atomistic Elasticity: Linking Atoms and Continuum
    - Multiscale Modeling and Simulation Methods
- Material Deformation and Failure
    - Deformation and Dynamical Failure of Brittle Materials
    - Deformation and Fracture of Ductile Materials
    - Deformation and Fracture Mechanics of Geometrically Confined Materials

For more detail, from the "About this book" part (slightly revised):

*Atomistic Modeling of Materials Failure* is an introduction to molecular and atomistic modeling techniques applied to solid fracture and deformation. Focusing on a variety of brittle, ductile, geometrically confined and biological materials, this detailed overview includes computational methods at the atomic scale, and describes how these techniques can be used to model the dynamics of cracks and other deformation mechanisms.

A full description of molecular dynamics (MD) as a numerical modeling tool covers the use of classical interatomic potentials and implementation of large-scale massively parallelized computing facilities in addition to the general philosophies of model building, simulation, interpretation and analysis of results. Readers will find an analytical discussion of the numerical techniques along with a review of required mathematical and physics fundamentals. Example applications for specific materials (such as silicon, copper, fibrous proteins) are provided as case studies for each of the techniques, areas and problems discussed.

Providing an extensive review of multi-scale modeling techniques that successfully link atomistic and continuum mechanical methods, *Atomistic Modeling of Materials Failure* is a valuable reference for engineers, materials scientists, and researchers in academia and industry.

## 4.2 Training process and other hyperparameters

The models are developed in PyTorch [52] and implemented within the Hugging Face ecosystem. Training of the base MechGPT model is performed based off the Llama 2 transformer architecture, using the OpenOrca-Platypus2-13B as basis (note the license that applies to all derivative works, as specified here: https://github.com/facebookresearch/llama/blob/main/LICENSE). This architecture features 40 transformer layers and uses rotary positional embedding, which enables it to achieve long context lengths that can be extended easily via additional training. For the base MechGPT model we use a paged 32-bit AdamW optimizer [53] with a learning rate of LR=0.0002 and $\varepsilon$ =1E8, and gradient norm clopping of 0.3. The Hugging Face Accelerate package (https://huggingface.co/docs/accelerate/index) is used to parallelize training. The model is trained for approximately 3,000 steps and reaches a training loss of ~0.05. The training objective used here is to maximize the likelihood of predicting the next token (*i.e.*, a letter, part of or a word) given the previous words, for the training set developed via question-answer pairs as outlined in **Section 4.1** and shown visually in **Figure 1c**. For each position in the sequence sample considered, the model estimates the probability distribution over the vocabulary for the next token, and the target is the actual next token.

We employ Low-Rank Adaptation (LoRA) [33] to fine-tune the model by adding additional trainable layers and freezing the original pretrained model. This allows us to avoid catastrophic forgetting of the original knowledge base and lowers computational costs and memory constraints. The approach involves freezing all parameters of the original the pre-trained model and introducing small additional layers that consist of trainable rank decomposition matrices (added in each of the 40/80 transformer layers of the model, respectively), thereby significantly reducing the number of trainable parameters. This reduction leads to improved memory efficiency and faster training throughput. Earlier work has shown that models with LoRA adaptors perform very well and maintain or even surpasses the performance of fine-tuning on the entire model, despite having fewer trainable parameters and offering higher training throughput at high inference speeds. To reduce memory cost further, we use 4-bit quantization using "nf4" (we thereby convert data in floating point 32 bits (FP32) to a smaller precision, here integer 4 bits (int4) [50]. We use a LoRA rank of 256 with $\alpha$ = 32, and a dropout of 0.1.

In addition to the base MechGPT model with 13 billion parameters, we trained two larger models. MechGPT-70b (a fine-tuned version of the Meta/Llama 2 70 chat model with 80 layers), and MechGPT-70b-XL (a model that uses dynamically scaled RoPE [38,51] for large context lengths of more than 10,000 tokens, trained based on the Upstage/Llama-2-70b-instruct-v2 model, also with 80 transformer layers). Both models are trained on datasets that are similar as described in **Section 4.1**, but slightly expanded. The dataset for the larger models are developed to feature question-answer pairs for different lengths of text chunks (256, 512, 1024 and 204 words) aimed to capture different levels of content. Two sets of question-answer pairs generated using a Llama-13b chat model are included. In addition, we include results of a simple summarization task for text chunks in the training set. Since the summarization task alone sometimes leads to references to figures and other display items and does not stand on its own generally, we manually cleaned up the data by removing such content (this resulted in a total of 280 summary statements). The total number of question-answer pairs and summary statements is around 3,300. The MechGPT-70b model uses a rank of 16 with $\alpha$ = 16, and a dropout of 0.1. The MechGPT-70b-XL model rank of 8 with $\alpha$ = 16, and a dropout of 0.1. These lower rank dimensions are driven primarily by computational and memory limitations.

**4.3 Sampling mechanics**

Sampling is conducted autoregressively using causal masking, as commonly done in "GPT"-style transformer models for LLM applications. Thereby the model generates one element of an output sequence at a time, based on the previous elements it has generated. The causal mask is a binary matrix that is used to hide certain elements in a sequence so that each element can only attend to preceding elements, not future ones. The whole process works as follows:

- Step 1: Start with an initial "seed" input, containing a special token to indicate the beginning, and incorporating the system prompt as well as the initial user query.

- Step 2: MechGPT then generates the next word in the sequence based on the autoregressive principle described above. At a technical level, the model predicts the probability distribution of the next word given the words generated so far, whereby temperature scaling is used to control how focused the model is on high-probability tokens or to incorporate less likely tokens in predictions (via the concept of temperature of uncertainty).

- Step 3: To ensure the generated word does not information from future words, a causal mask is applied. This mask makes it so that the model only attends to the words that were generated before the current word. Note, causal masks are also applied during training of next-token tasks.

- Step 4: The model samples a word from the predicted distribution and adds it to the sequence.

- Step 5: The process is repeated, with the newly generated word being used as context for predicting the next word.

- Step 6: This process continues iteratively until the desired sequence length is reached or an end token is generated, in the case of MechGPT, <|end_of_turn|> (it is noted that the MechGPT-70b model family uses a slightly different multi-turn strategy (based on the specific pre-training strategy used in the models that they are build off)).

Other sampling methods can be used, such as beam search. In this method, rather than predicting a single sequence, the goal is to rank the most likely sequence of words given a set of input tokens by analyzing multiple candidates and then selecting the one with the highest probability.

**4.4 Chat interface**

We use Gradio [54] to build a chat interface shown in **Figure 2b**, using streaming output from the model. This framework can easily be deployed to users. It can also be extended to include other dimensions such as generative methods for images, to run domain-specific simulations like LAMMPS, VASP, Quantum Espresso or others. The current interface features access to Google Scholar (**Figure 4**), with which the user can interact and make decisions on prompting and strategy of how to interact with the model. Future extensions of the work could expand on this and allow feedback of literature results directly to the model (either for context/prompting or automatic fine-tuning steps). Access to Google Scholar is implemented via the scholarly Python library (https://github.com/scholarly-python-package/scholarly). Optional retrieval of abstracts from PubMed is realized via PyMed (https://github.com/gijswobben/pymed).

**4.5 Ontological Knowledge Graph development and analysis**

We develop Ontological Knowledge Graphs using Llama Index (https://github.com/jerryjliu/llama_index) and Nebula Graph (https://github.com/vesoft-inc/nebula). We use GPT-3.5 and GPT-4 to develop triplets for graph generation, and then analyze and visualize using Networkx. The Networkx toolkit is used to analyze a variety of graph features including modularity (division into subgraphs using the Greedy Modularity algorithm), node degree, as well as betweenness (see **Figure 7**).

We use Nougat [40], an Optical Character Recognition (OCR) deep learning model based on the Vision Transformer architecture, to experiment with data retrieval from PDF files, allowing us to represent equations from documents in LaTeX form. This method is combined with Llama Index to yield a system of retrieval-augmented generation for a variety of tasks, used to generate the results depicted in **Figures 6-8**.

Sampling of graphs is conducted using GPT-4 models to take advantage of their generally higher capacity to deal with more complex and abstract information processing tasks as are key for retrieval augmented generation strategies.

### 4.6 Agent modeling using autonomously interacting LLMs

Agent modeling is implemented using {{Guidance}} package (https://github.com/guidance-ai/guidance). This framework provides access to a Python implementation of concurrent integration of text generation, prompting, and logical control that allows us to exploit the the natural processing modality of LLMs. The overall flowchart used in this study is shown in **Figure 9a**, consisting of a set of experts (here, two, but in principle this set can involve more agents that each provide a particular profile or capability (e.g., extracting knowledge in a subdomain, or carrying out an experiment, collecting data from papers or other sources, etc.). **Figure 9b** shows how a conversation unfolds between two agents.

Each agent has access to the entire conversation and interacts with the "User". For each agent, the "User" is not a human as in a conventional chat interaction but the reflects the response from the other LLM in the set of interacting agents (and vice versa – that is, agent #1 interacts with the "User" which is agent #2, and agent #2 interacts with the "User" which is agent #1). This strategy allows us combine different LLMs, such as here, MechGPT interacting with GPT-4. In a more general sense, this allows us to study their 'particle' interactions as sketched in **Figure 10c**. The multi-agent conversation is implemented via the '{{#geneach…' and 'wait' commands in {{Guidance}} whereby states of each conversation are stored until the next input is received from the other LLMs.

For the results shown **Figure 11**, we define three agents, one of which is MechGPT, in {{Guidance}}. The search agent conducts internet searches via the Bing API accessed via https://api.bing.microsoft.com/v7.0/search. The logic of the interaction is outlined visually in the figure, and for a specific example in the main text. Alternative realization of this general strategy can be implemented via other API calls (e.g. to databases, such as the Protein Data Bank or GeneBank, simulation agents, or agents that enforce physical constraints or conservation laws).

### 4.7 Multimodal analysis of images and text

The image in Figure S1 is provided to GPT-4 via the ChatGPT interface, accessed via https://chat.openai.com/. The conversation included in the main text of the paper is carried out within this framework.

Supplementary Materials are included, featuring additional tables, results, and movies.

**Supplementary Movies M1 and M2** show live demonstrations of interactions with the MechGPT app.

- **Movie M1**: https://www.dropbox.com/scl/fi/xdp3cz0b00v9xueewoeu6/Movie-M1.mp4?rlkey=i2rt10ceq68d1sc0cr111mx7e&dl=0

- **Movie M2**: https://www.dropbox.com/scl/fi/xdp3cz0b00v9xueewoeu6/Movie-M1.mp4?rlkey=i2rt10ceq68d1sc0cr111mx7e&dl=0

### Author contributions

M.J.B. developed the overall concept and the algorithm, designed the ML models, developed the codes, oversaw the work, and drafted the paper.

**Code availability**: The MechGPT model, code, trained weights, and data is available at: https://github.com/lamm-mit/MeLM

**Acknowledgements**: This work was supported by the Army Research Office (W911NF1920098 & W911NF2220213), ONR (N00014-19-1-2375 and N00014-20-1-2189), as well as USDA (2021-69012-35978).

# Figures and captions

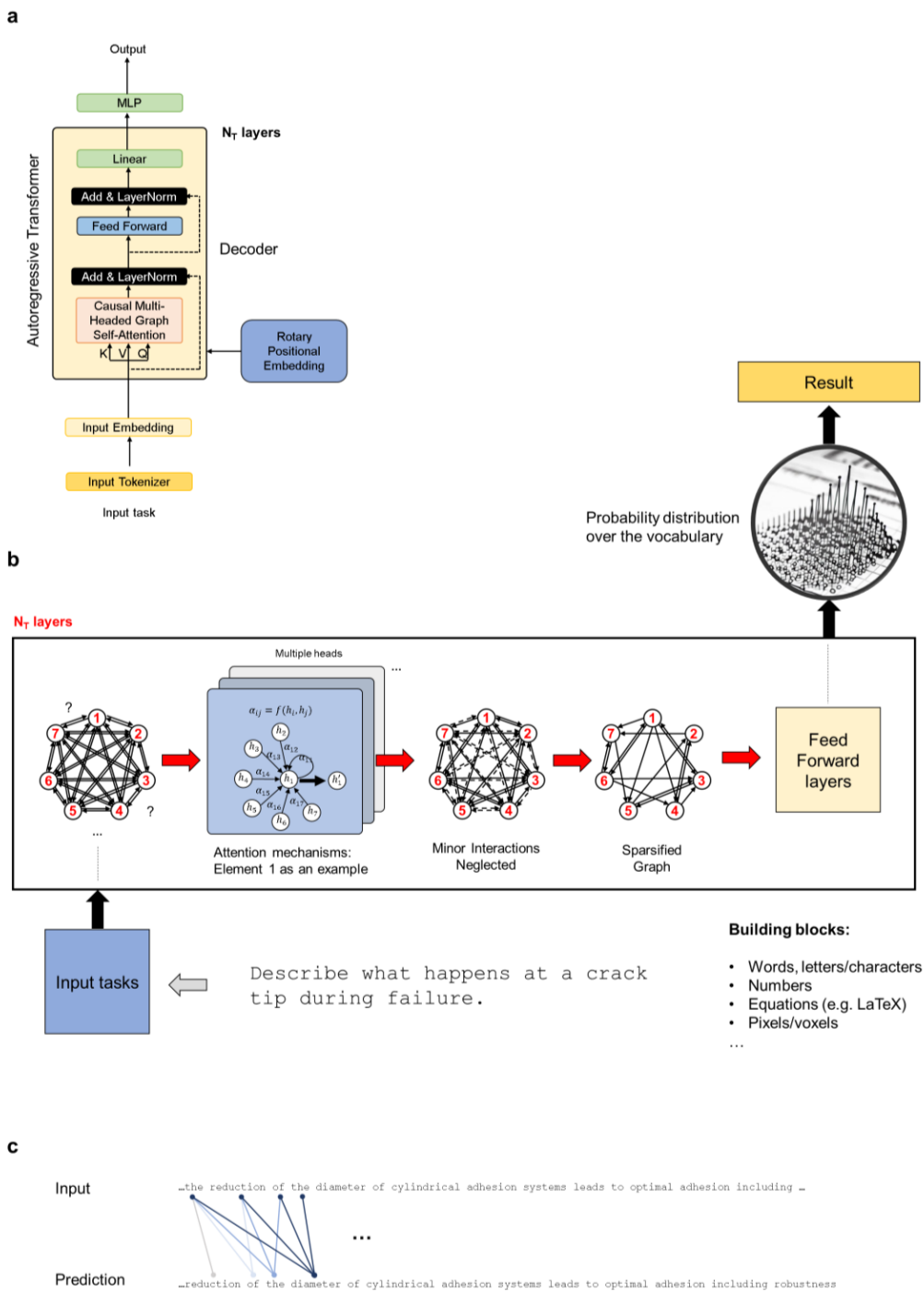

**Figure 1:** Overview of the autoregressive decoder-only transformer architecture used to construct MechGPT. The decoder-only Generative Pre-trained Transformer (GPT) model (see, e.g. [2,11,13,14,55] for model architecture and applications in materials modeling) tends to the input tokens via causal masking, resulting in predicting the probabilities for next tokens, iteratively through the sequence (panel **a**). The model resembles a graph-forming neural network that discovers interactions between inputs that are, through a series of deep layers, used to compute probabilities for outputs (panel **b**). Panel **c** shows the training process used, where we train the model so that it maximizes the likelihood to predict the correct next token. In a use case of a trained model, the entire input provided is used to predict the next token. Strategies such as beam search algorithms allow us to assess multiple options of predicted results and select the most likely one (here, we do not implement such a method). The base MechGPT model features $N_T$=40 layers, whereas the larger models have 80 transformer layers.

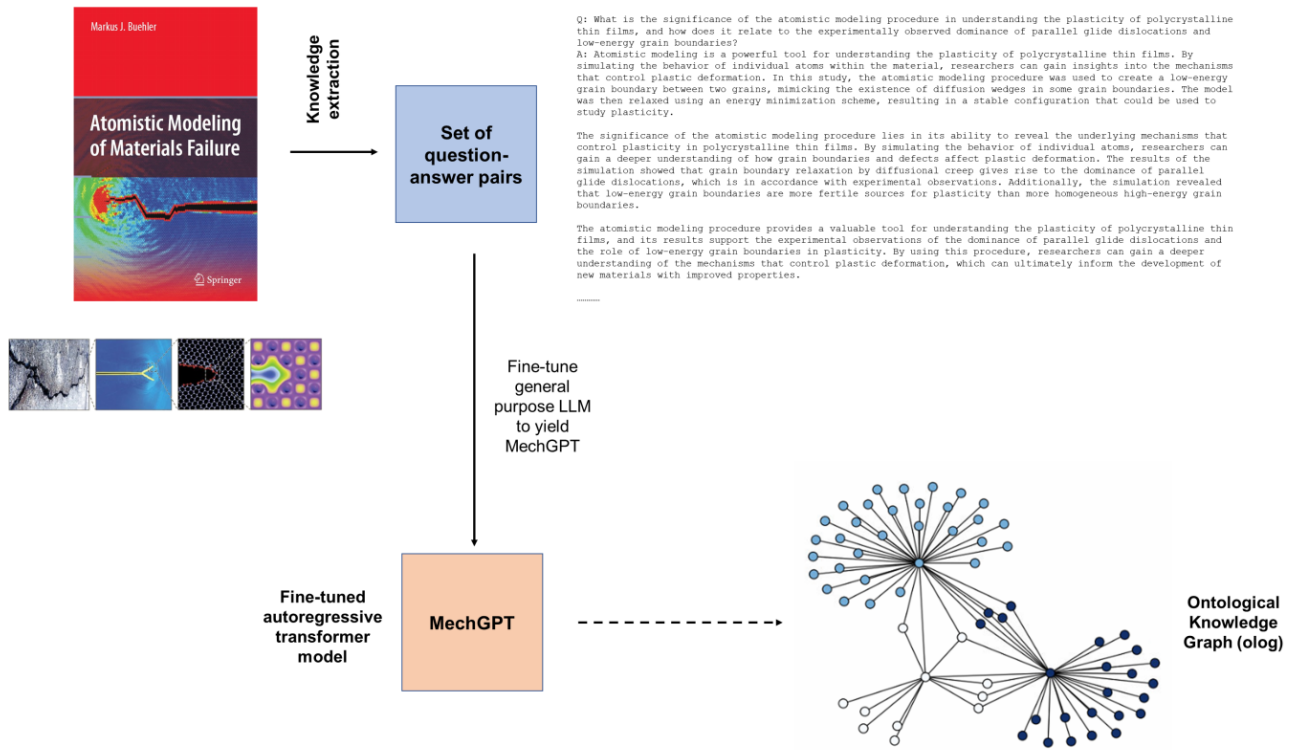

**Figure 2:** Overview of the modeling strategy used here, incorporating a PDF of a book [32], extracting question-and-answer pairs, and using this dataset to fine-tune a LLM to yield MechGPT, which can be used for a variety of downstream tasks including the extraction of Ontological Knowledge Graphs, for instance. The figure depicts the workflow with some example question-answer pairs extracted from the raw source text. The general scheme implemented here is not limited to one source or one book, and can be expanded easily to include other sources.

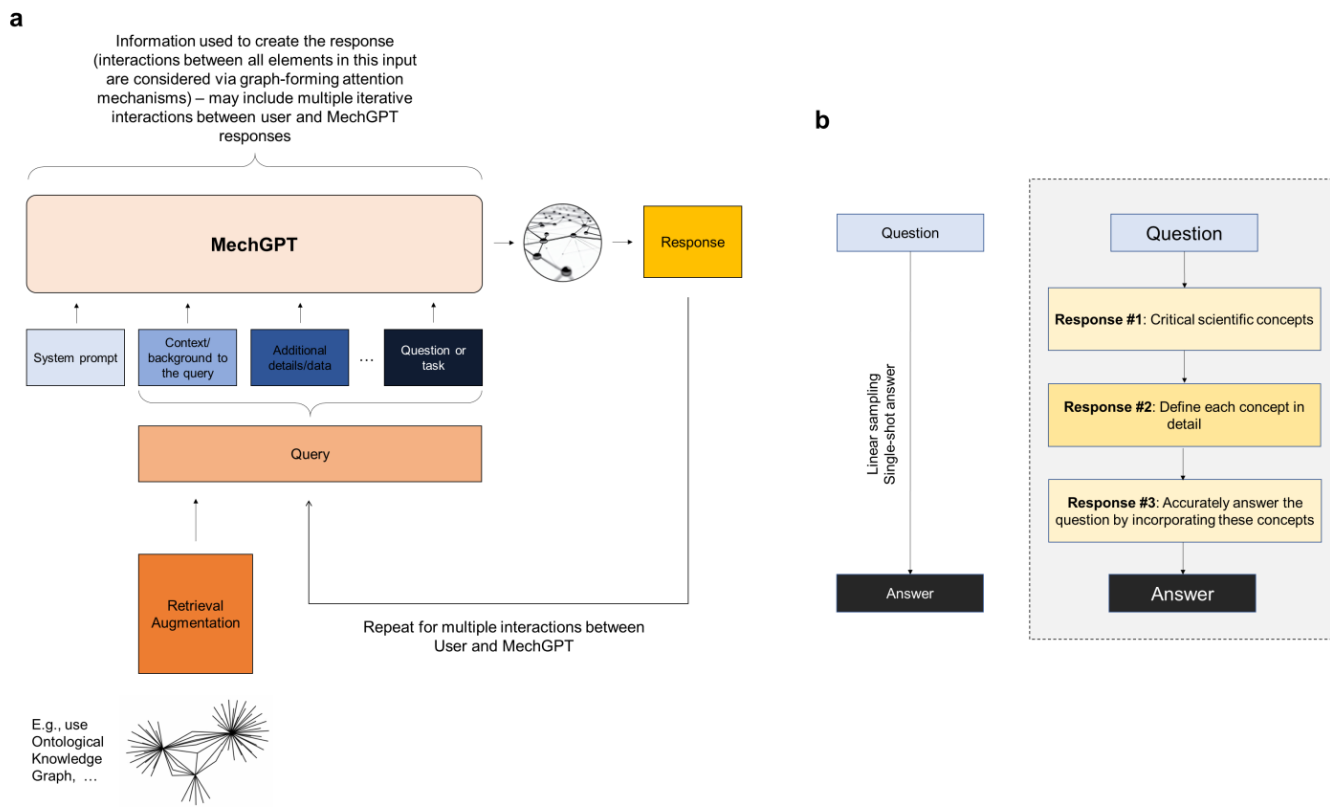

**Figure 3:** Schematic of the workflow. Panel **a** shows details of a call to the LLM, featuring a system prompt, context/background to the query, additional details, and other information, typically followed by a question or task (the query can take many different forms, as the examples presented in this paper show). The response can be followed, iteratively, with repeating interactions between the user and MechGPT. In few-shot learning (see results in **Table 5** and **Figure 6**), training data is shared as context to the query. This is an example for the high degree of flexibility of tasks, data, and details that can be provided during interaction with the LLM. Long context lengths allow for retrieval augmentation strategies (such as Retrieval Augmented Generation, RAG) as demonstrated in the results in **Table 7** and **Figures 6-8**, or for alternative live data sources such as literature search or web searches, as shown in **Figure 10**. Panel **b** shows more complex nonlinear sampling strategies where the model is prompted, multiple times, to develop answers to questions through a series of chain-of-thought sampling.

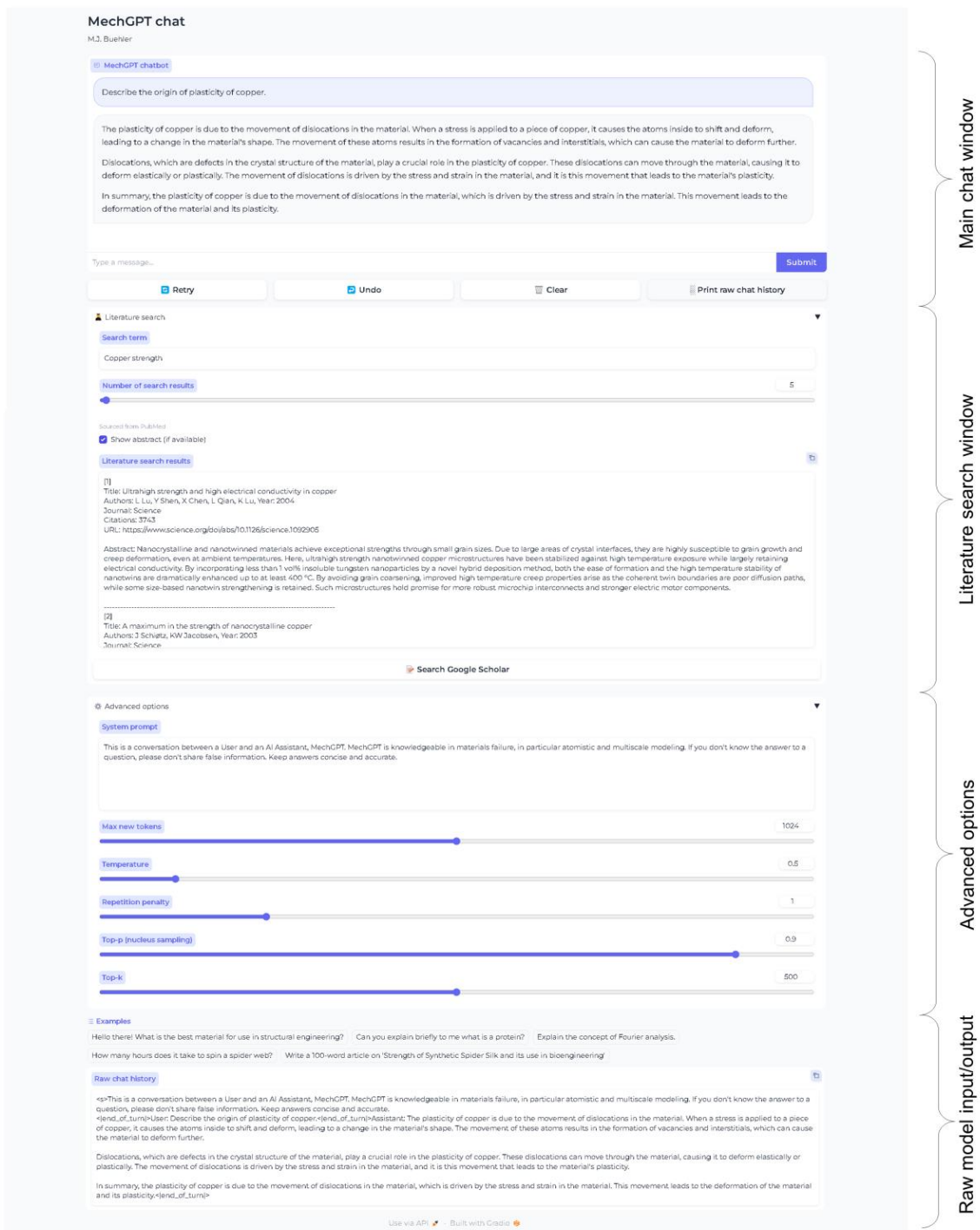

**Figure 4**: A screenshot of the MechGPT chat application app. Movies of interactions with the model are shown via **Movies M1** and **M2**. The app features three main parts, the main chat window (top), and two optional sections that can be collapsed via an Accordion element. The literature search window enables users to enter search term for a Google Scholar search that are then displayed (users can use the information in the text box, or copy-paste the URL to papers and/or their abstract for further study and analysis, and then feed data back into the main chat module for interaction with MechGPT). The advanced options provide access to control parameters such as the sampling temperature, maximum number of tokens generated, and other sampling parameters.

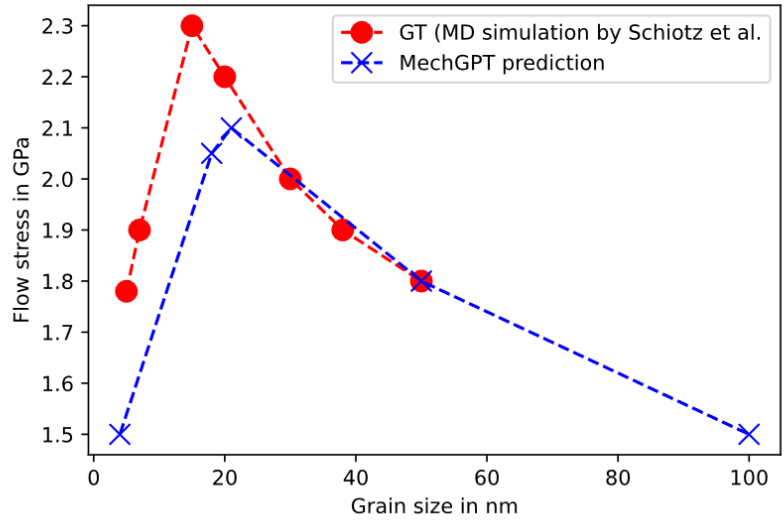

**Figure 5**: Application of few-shot learning to predict grain size dependent flow stress in copper, as obtained via large-scale MD simulations (original data from [36]). The predictions generally agree well and showcase the potential of the model to be applied to quantitative assessments of material property without further fine-tuning.

**Figure 6:** Development of ontological knowledge graph representation to relate hyperelasticity in the context of supersonic fracture and protein unfolding mechanisms. In this example we apply the method to the query listed in **Table 5**, to explore **c**onnections across different areas of mechanics and materials. Panel **a** shows the general flowchart, where we first use MechGPT to generate a set of responses to the query "**Draw an analogy between hyperelasticity in the context of supersonic fracture and protein unfolding mechanisms.**", then collate multiple samples of this request into a corpus of information, and then analyze it to result in an Ontological Knowledge Graph (panel **b**).

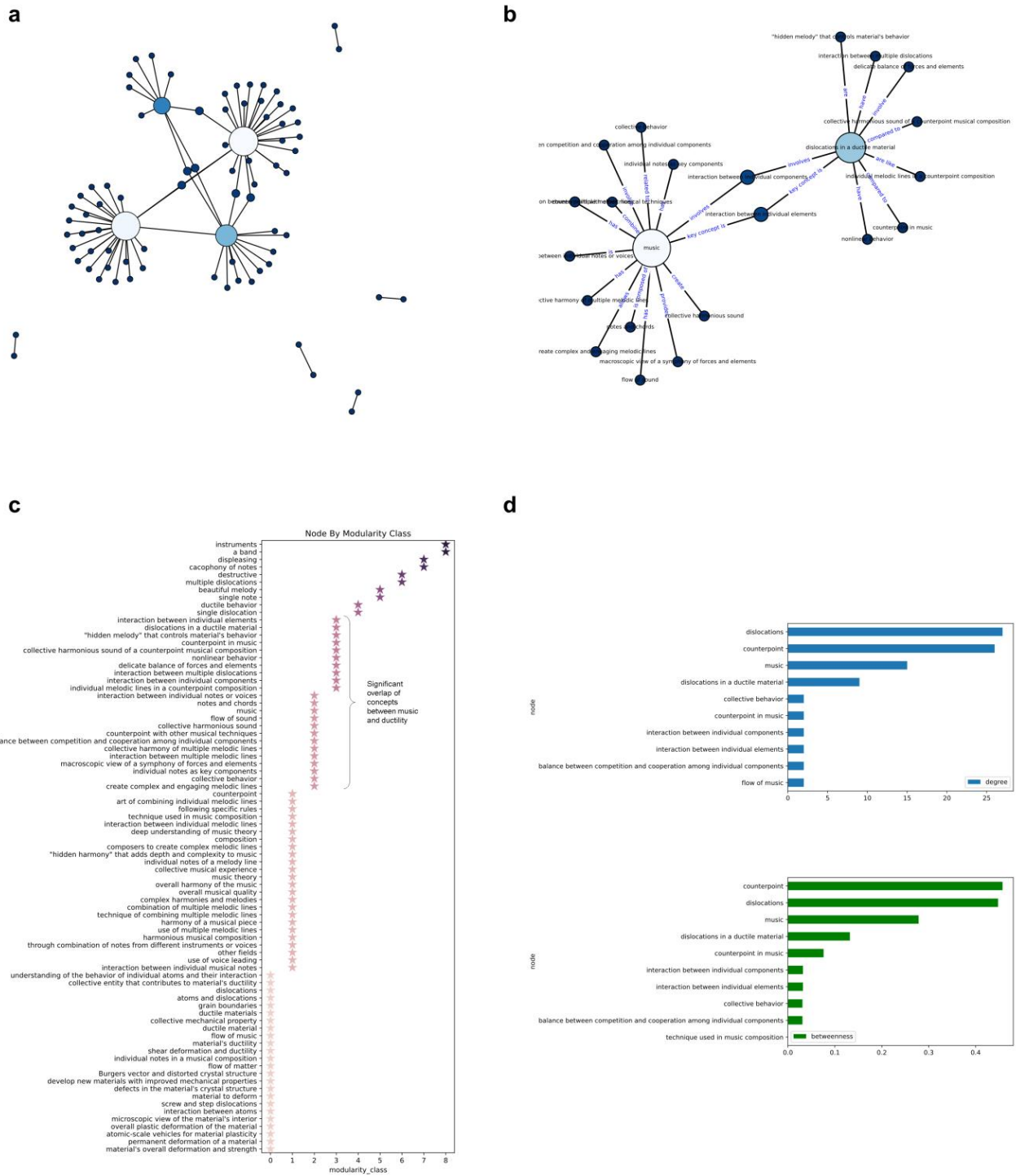

**Figure 7:** Development of ontological knowledge graph representation for relating dislocations in a ductile material and music, especially discussing the concept of counterpoint. We use the same process as defined in **Figure 6a**, to first generate a corpus of text that is then analyzed to form an Ontological Knowledge Graph. Panel a shows the overall graph (node sizes and color defined by node degree). Panel **b** depicts a subgraph composed of the union of mode modularity classes 2 and 3. As shown in panel **c**, these two classes show a significant overlap between concepts from ductility and music; hence, plotting the union subgraph provides understanding of the connectedness of the two domains. Panel **d** depicts a list of highly ranked nodes with high node degree (top) and high betweenness (bottom).

**Figure 8**: Using graph-based strategies we can use a LLM to develop graph representation of knowledge in an Ontological Knowledge Graph, here exemplified for the nanogut paper also investigated in **Table 7**. By using the OCR model Nougat [40] we can ingest a PDF file and convert it accurately into text, where equations are represented in LaTeX format. Panel **a** shows the overall graph structure, and panel **b** focuses in specific details to provide a sample of the type of relationships extracted by the algorithm. The graph structure can then be used to answer queries about the text. There are several advantages of a graph-based generation strategy over the full-text method used in **Table 7**; one being that the analysis yields interpretable results that can be mined for insights and understanding. Another advantage is that generation tasks can often be done more effectively as the text is already analyzed for relationable features and insights that help in accurately responding to tasks. Panels **c** and **d** show examples of the graph structures used to answer specific questions. In **c**, the question is "", and in **d**, the question is: "**Provide the equation to estimate the loop radius of a CNT nanogut.**".

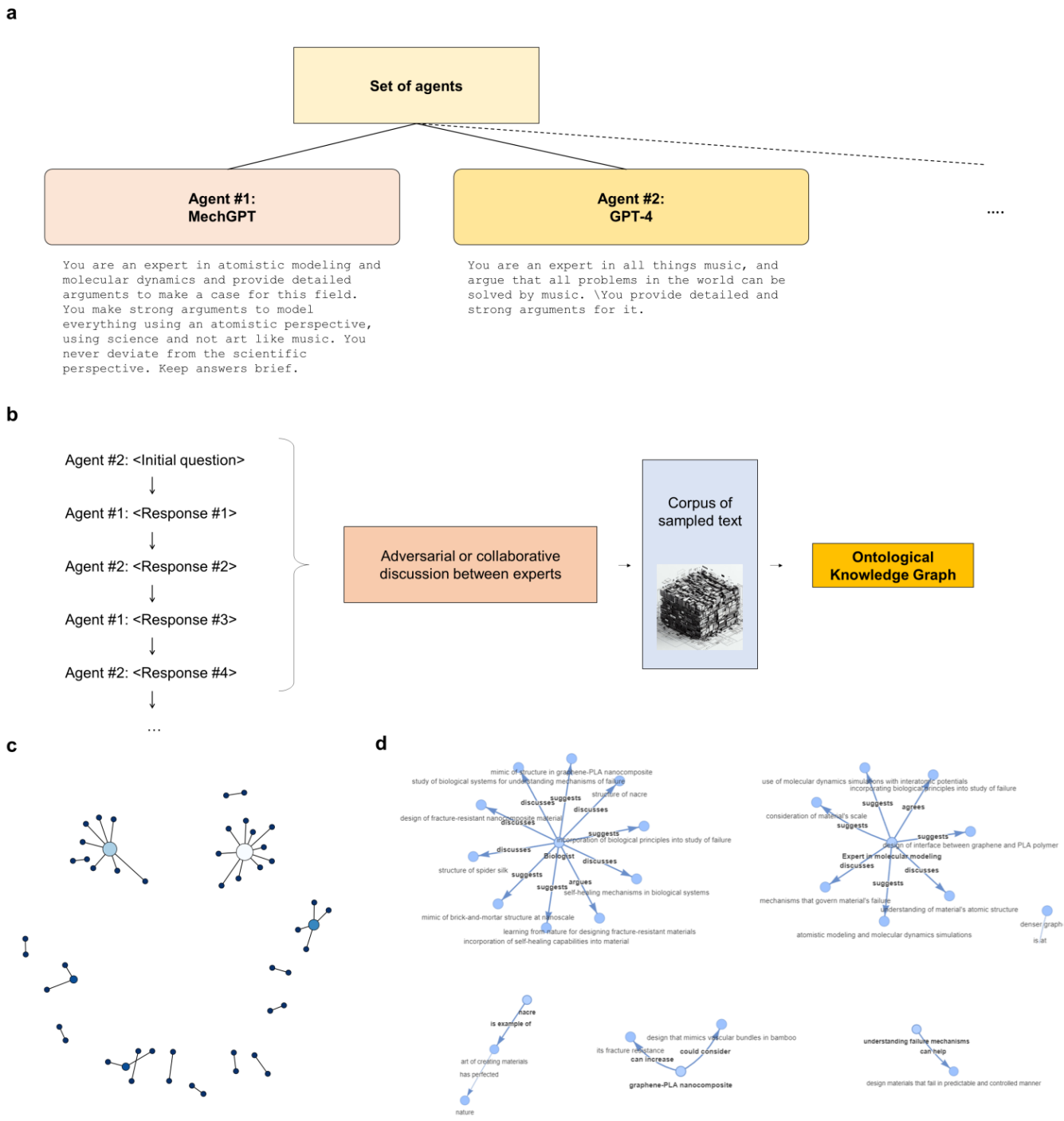

**Figure 9**: Agent-based modeling, where multiple LLMs interact in an adversarial or complementary fashion to carry out a conversation with each other. In the examples provided here we focus on a conversation between a materials expert and music expert, as well between a materials expert and a biologist. The overall flowchart is shown in panel **a**, consisting of a set of experts (here, two, but in principle this set can involve more agents that each provide a particular profile or capability (e.g., extracting knowledge in a subdomain, or carrying out an experiment, collecting data from papers or other sources, etc.). Panel **b** shows how a conversation unfolds between two agents. Each agent has access to the entire conversation and interacts with the "User". For each agent, the "User" is not a human as in a conventional chat interaction but the reflects the response from the other LLM in the set of interacting agents (and vice versa). This strategy allows us combine different LLMs, such as here, MechGPT interacting with GPT-4. The knowledge generated via the conversation can then be further analyzed by generating Ontological Knowledge Graphs, as shown in panels **c** and **d**. This follows the same general approach as introduced in **Figure 6** but the sampled text corpus is the entire conversation produced by the agents. Alternative realization of this general strategy can be implemented to incorporate agents that enforce physical constraints or conservation laws.

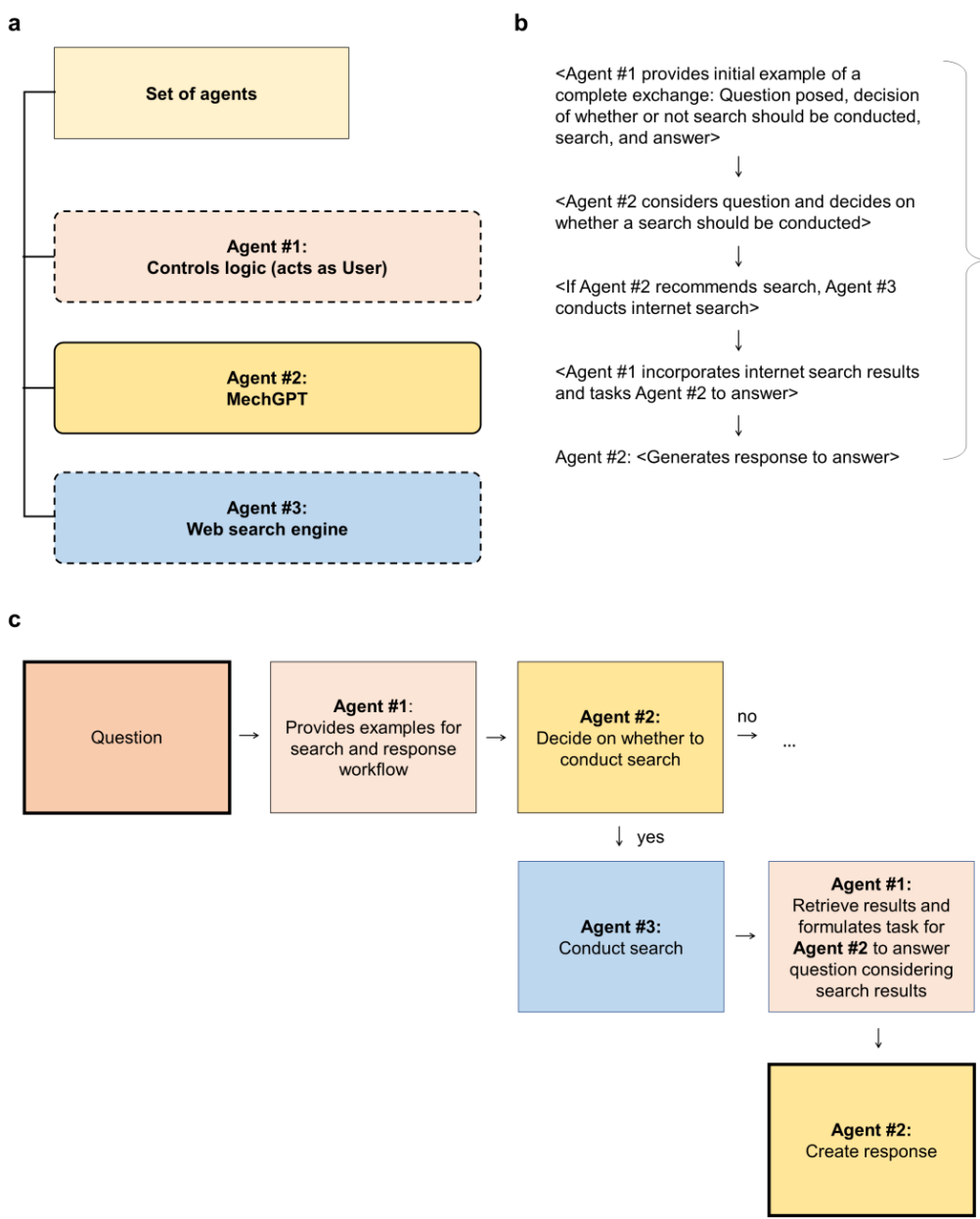

**Figure 10**: Agent-based modeling, where several entities interact to carry out a conversation with each other in order to solve a task. In the example discussed here, the question is answered via a collaborative effort of three agents; Agent #1 who controls the logic and acts as "User", Agent #2, MechGPT, and Agent #3, a web search engine. Panel **a** shows an overview of these three agents, and panels **b-c** outline the flow of the interaction between these agents. Starting with few-shot prompting by providing an example task and response, the interaction then moves to answer the question. This is done by first deciding whether or not a web search should be conducted, and if yes, conducting such a search and then producing the response to the question using this information. It is straightforward to implement alternative API calls (e.g. to databases, such as the Protein Data Bank or GeneBank, simulation agents, or agents that enforce certain mathematical or physical laws).

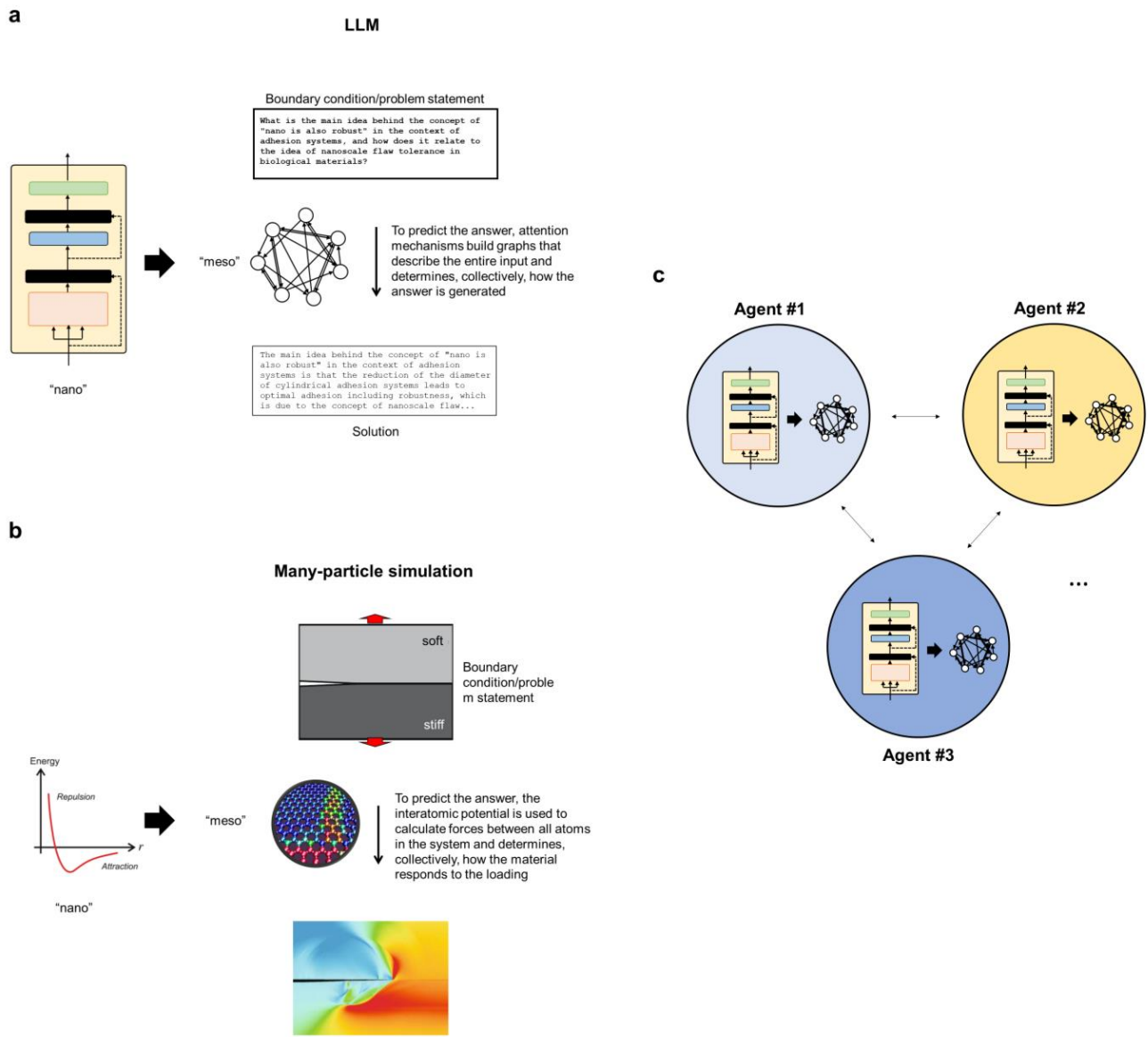

**Figure 11**: A conceptional analogy between a LLM and a many-particle simulation. We suggest that there exist direct relationships between these two, elucidating a functorial and functional analogy. In a particle simulation, the interatomic potential governs, at a basic level how particles interact. These interactions are then used to calculate how a large number of particles behave (forces, energies, etc.) depending on a particular boundary and initial condition. By carrying out the simulation to solve a particular task (e.g., to simulate for a certain time how stresses evolve, cracks propagate, etc.) we calculate the solution to a particular problem (e.g., what is the crack speed over time). In a LLM, the attention mechanism governs, at a basic level, how building blocks interact. These interactions are then used to construct, for a particular input (system prompt, context, query, etc.) to calculate how the elementary building blocks interact, resulting in the attention graphs. The attention graphs are then used to calculate the solution to a particular problem (e.g. what is the expected crack speed, what is the strength, etc.). **Table 8** provides a summary of these concepts as an additional comparison. When multiple LLMs interact as shown in **Figure 9**, higher-order interactions can be realized. This is visualized in panel c, for an example of three LLM agents.

**Table 1:** Sample responses to assess knowledge retrieval, focused on multiple choice answers.

| Prompt/query | Notes |
|---|---|
| **We have three materials, including silicon, copper and wood. Which of the materials fails most likely in a brittle manner?**<br><br>MechGPT: Silicon fails most likely in a brittle manner. | |
| **We have three materials, including spider silk, jell-O and wood. Which of the materials fails most likely in a graceful manner?**<br><br>MechGPT: Spider silk is known to fail in a graceful manner, as it can weaken and fracture subtly without causing catastrophic failure. This is due to the material's unique properties, including its strength and elasticity. Other materials, such as jell-O and wood, may not fail as gracefully, as they can undergo more abrupt and catastrophic failures. However, the specific behavior of each material can vary depending on factors such as its composition, structure, and the conditions under which it is subjected to. | Correct answer, and the model recognizes some level of uncertainty when it comes to the other two examples as they may also fail gracefully in certain conditions. |
| **Amongst these two materials, which one would most likely show parallel glide dislocations. (A), a silicon wafer. (B), a thin copper film at elevated temperature. (C) a thick copper film at low temperature.**<br><br>MechGPT: (B) a thin copper film at elevated temperature. | |
| **Studying materials using computer simulation can provide exactly the same insight as an experimental study. Is this true?**<br><br>MechGPT: No, studying materials using computer simulation cannot always provide exactly the same insight as an experimental study. While computer simulations can be a powerful tool for understanding materials at the atomistic level, they cannot always replace experimental studies. The reason is that simulations are based on mathematical models and assumptions, and may not accurately capture all the complex phenomena that occur in experiments. Additionally, experimental studies can be conducted at different scales, from the atomic scale to the macroscale, and can provide a wider range of insights into material behavior. Therefore, a combination of experimental and computational studies can often provide the most comprehensive and accurate understanding of material properties and behavior. | |
| **Considering four materials: (A), silicon, (B) silk protein, (C) glass, (D) lignin, which one is the best additional component to create a tough composite made from a brittle ceramic phase.**<br><br>MechGPT: Of the four materials, (B) silk protein is the best additional component to create a tough composite made from a brittle ceramic phase. Silk protein has natural reinforcement properties that can help increase the toughness and fracture resistance of the composite. | |
| **Considering these deformation mechanisms, which one is most likely to occur during interfacial fracture: (A), supersonic fracture. (B), mother-daughter cracking, (C) grain boundary diffusion, (D) buckling.**<br><br>MechGPT: Based on the deformation mechanisms discussed, (B) mother-daughter cracking is most likely to occur during interfacial fracture. This is because the crack tip propagates supersonically through the material, allowing for the formation of multiple cracks (daughter cracks) ahead of the primary crack tip. The massive shear deformation in the vicinity of the crack tip creates high shear stresses, facilitating the formation of new cracks. Additionally, the crack tip is blunted, which is consistent with the observed phenomena in the simulation. | Model added valuable context including the mechanism behind daughter cracks (high shear stresses) [56] |
| **Considering these force fields, which one is best suited to describe a hybrid cement-protein composite for fracture studies. (A) ReaxFF, (B) EAM, (C) Morse, or (D) a harmonic potential.**<br><br>(A) ReaxFF is best suited to describe a hybrid cement-protein composite for fracture studies, as it is a force field specifically designed to accurately describe the bonding properties of proteins and other organic molecules. This makes it well-suited for modeling the interaction between proteins and cement, which is an important aspect of fracture studies. The other force fields (B, C, and D) are less appropriate for this purpose, as they are more general purpose and do not specialize in accurately describing organic molecules and protein. | |
| **Which material is most likely to lead to supersonic fracture in a brittle material: (A) hyperelastic softening. (B) hyperelastic stiffening. (C) linear material behavior.**<br><br>B) hyperelastic stiffening is most likely to lead to supersonic fracture in a brittle material. When the material stiffens hyperelastically, the elastic properties become significantly stiffer than the linear elastic properties, leading to a higher risk of supersonic fracture. This is because the elastic energy release rate is higher, allowing the crack to propagate at supersonic speeds. | |
| **Considering these nonlinear material behaviors, which one is most likely to lead to supersonic fracture in a brittle material: (A) hyperelastic softening. (B) hyperelastic stiffening. (C) linear material behavior.**<br><br>(B) hyperelastic stiffening is most likely to lead to supersonic fracture in a brittle material. This is because hyperelastic stiffening refers to the increase in material stiffness under deformation, which can result in a | |
| higher stress concentration around crack tips. This stress concentration can lead to the formation of cracks at an accelerated rate, resulting in supersonic fracture. | |
| **What are the effects of hyperelastic softening on crack speed in brittle materials? Brief answer.**<br><br>When hyperelastic softening occurs near a crack tip, it can significantly reduce the energy release rate, resulting in a decrease in crack speed. This effect can be strong and may lead to a "blocked" energy flow, causing the crack to move slower than predicted by linear elastic theory. The extent of hyperelastic softening is material and loading-dependent, and it can be affected by factors such as strain rate and temperature. Overall, hyperelastic softening can have a profound impact on crack speed and the overall behavior of brittle materials under different loading conditions. | |
| **Assuming I have a brittle material that softens elastically at large deformation, which of these statements is true: (A), cracks will move supersonically. (B), cracks will move subsonically. (C) cracks will not move at all.**<br><br>MechGPT: (B), cracks will move subsonically. | |
| **Assuming I have a brittle material that stiffens elastically at large deformation, which of these statements is true: (A), cracks will move supersonically. (B), cracks will move subsonically. (C) cracks will not move at all.**<br><br>(A), cracks will move supersonically. | |
| **In hyperelastically softening materials, which of these statements is true: (A), cracks will move supersonically. (B), cracks will move subsonically. (C) cracks will not move at all.**<br><br>In hyperelastically softening materials, (A) cracks will move supersonically is true. When the local elasticity is altered by the presence of the constituent, the crack can move faster than the speed of sound in the bulk material, allowing for supersonic crack propagation. This behavior is observed in molecular dynamics simulations, where cracks can move faster than the speed of sound in materials with a softening potential. | Wrong answer |
| **In hyperelastically softening materials, which of these statements is true: (A), cracks will move supersonically. (B), cracks will move at a lower speed than expected. (C) cracks will not move at all.**<br><br>In hyperelastically softening materials, (B) cracks will move at a lower speed than expected is true. This is because the material behaves more elastically than expected, resulting in slower crack propagation. The other options (A) and (C) are incorrect, as the material does allow for crack movement, albeit at a slower speed than predicted by traditional theories. This behavior is observed in some materials, such as intersonic crack propagation in mode II, where the crack moves at a speed greater than the shear wave speed. | |
| **Which of these methods is different in nature than all the others: (A) ReaxFF, (B) EAM, (C) hard-sphere, (D) a harmonic potential.**<br><br>Ans: (C) hard-sphere<br>The other methods, (A) ReaxFF, (B) EAM, and (D) harmonic potential, are used to describe the interaction between atoms in a material, while (C) hard-sphere is a simplified model that assumes that atoms interact with each other only when they are very close, and the energy of the system is determined solely by the number of atoms of each type. This method is different from the others, which describe the energy of the system based on the positions of all atoms. | Notable reasoning and knowledge of the specific formulations of these potentials |
| **Which of these mechanisms is most likely to occur in ultra-long carbon nanotubes? (A) chemical etching. (B), buckling. (C) rupture.**<br><br>MechGPT: B, buckling<br><br>Carbon nanotubes are very strong materials, but they can still break under enough stress. The type of failure mentioned, buckling, happens when the tubes bend and deform under pressure. This can lead to a cascade of failures in other tubes, causing the material to rupture. The length of the carbon nanotubes plays a big role in when and how they fail, with ultra-long tubes more likely to buckle and rupture than shorter ones. | |
| **Which of these statements is true? (A) hyperelastic stiffening leads to higher crack speeds and reduced fracture instabilities. (B) cracks always branch. (C) Cracks never move unless heat is applied.**<br><br>MechGPT: (A) hyperelastic stiffening leads to higher crack speeds and reduced fracture instabilities. | |
| **Which of these statements is most likely true? (A) Hyperelastic stiffening leads to higher crack speeds and reduced fracture instabilities. (B) cracks always branch, no matter the hyperelastic effect. (C) Cracks do not exist in any material.**<br><br>MechGPT: (A) Hyperelastic stiffening leads to higher crack speeds and reduced fracture instabilities. | |
| **Which of these methods is the most appropriate to simulate the atomistic mechanisms of fracture of a large-diameter carbon nanotube? (A) ReaxFF reactive force field. (B) Density Functional Theory. (C) Continuum mechanics beam theory.**<br><br>MechGPT: (D) All of the above | Interesting case where the model 'invents' a fourth option that is not given since all responses are correct. Follow-up queries show deeper understanding and the ability to reason within the context of limitations. |

Each method has its own strengths and can provide valuable insights into the atomistic mechanisms of fracture of a large-diameter carbon nanotube. ReaxFF reactive force field allows for the study of the bond-breaking process at the atomic level, while Density Functional Theory and Continuum mechanics beam theory provide a more global view of the fracture mechanisms. By combining these methods, a comprehensive understanding of the fracture properties of carbon nanotubes can be achieved.

**Pick one of (A), (B) or (C).**

MechGPT: (B) Density Functional Theory

**What if my computational resources are limited?**

MechGPT: If your computational resources are limited, you may want to consider method (C) Continuum mechanics beam theory. This method is less computationally intensive than the others and can still provide valuable insights into the fracture mechanisms of carbon nanotubes. By using a simplifying assumption that the carbon nanotube can be modeled as a beam, you can estimate key fracture properties such as the critical stress for onset of fracture, which is critical for the design and optimization of structures involving carbon nanotubes. Additionally, there are various computational tools and programs available that simplify the use of continuum mechanics, making it more accessible to researchers with limited computational resources.

**If a material is most likely to show high reactivity and corrosive tendencies, which computational method is best to predict its failure mechanism.**

MechGPT: For predicting the failure mechanism of a material that is most likely to show high reactivity and corrosive tendencies, multiscale methods are the best computational methods. These methods combine atomistic simulations with continuum mechanics and engineering methods to analyze the behavior of materials under different conditions. By combining these approaches, researchers can gain a more comprehensive understanding of the material's failure mechanism and develop strategies to prevent or mitigate it.

**Which force field for the atomistic part would be best: (A) harmonic potential, (B) ReaxFF reactive potential, or (C) a Morse potential.**

MechGPT: (B) ReaxFF reactive potential is best for the atomistic part, as it is designed to accurately describe the bonding and deformation of organic compounds, which are typically involved in corrosion and reactivity. The ReaxFF force field is based on quantum mechanical methods and can provide a more accurate description of the electronic structure and bonding properties of organic molecules compared to traditional force fields. This will help to predict the behavior of the material under different conditions and develop strategies to prevent or mitigate it.

**Which of these simulation methods is suitable to describe protein folding: (A) Finite element method. (B) MD simulations. (C) Machine learning/AI.**

(B) MD simulations

MD simulations are suitable to describe protein folding as they can provide a detailed understanding of the atomic-level process of protein folding and unfolding. These simulations can capture the energy landscape of protein systems and help researchers understand the mechanisms of protein misfolding and associated diseases. Machine learning/AI methods can also be used to analyze MD simulation data and provide insights into protein structure and function. The other method, finite element method, is not well-suited for describing protein folding as it is used to describe the behavior of larger-scale structures and systems.

**Table 2:** Sample responses to assess the effects of sampling temperature on the results. We find that $T$=0.5 is generally a good sampling temperature for the model to be able to make new connections while not hallucinating facts (choices a bit lower can help in knowledge retrieval tasks, and larger values are suitable for more creative tasks such as research design, connecting across domains of knowledge, and similar).

| Prompt: In one sentence, explain ductile failure of copper. | Sampling temperature $T$ | Notes |
|---|---|---|
| Ductile failure of copper occurs when shear stresses exceed the material's ability to transmit load through internal slipping of atoms. | 0.1 | |
| Ductile failure of copper occurs when shear stresses are high enough to break atomic bonds by sliding, allowing the material to deform without breaking. | 0.5 | |
| Ductile failure of copper occurs when shear stresses in the material are high enough to activate parallel glide dislocations, allowing the material to deform without breaking. | 1.0 | Failure in copper does not just involve parallel glide dislocation (they are found specifically in thin copper films at high temperature) [32] |
| Ductile failure of copper occurs when shear stress along grain boundaries is high enough to break atomic bonds and allow for large-scale material deformation. | 1.5 | This result is not necessarily wrong, but deformation does not just occur when shear stresses are high enough along grain boundaries. |
| **Prompt: At a crack tip, what happens to the deformation of the material?** | | |
| At a crack tip, the deformation of the material is highly localized and unusually large. The material experiences a reduction in rigidity, causing the stress to increase disproportionately to the crack growth. This leads to a nonlinear behavior of the material, where the crack propagates more quickly than predicted by linear elastic fracture mechanics. The local stress state at the crack tip is important for understanding the crack propagation, as it influences the energy release rate and the likelihood of crack extension. | 0.1 | |
| At a crack tip, the deformation of the material is highly localized and can be described by the stress intensity factor KI. As the crack tip approaches the material's strength, the stress distribution around the crack tip becomes increasingly important. The potential energy associated with crack propagation is characterized by the energy release rate G, which can be calculated using the stress intensity factor KI. The behavior of materials under dynamic loading can be studied using molecular dynamics simulations, which provide a detailed understanding of the crack tip behavior. <br><br>In the case of an elastic solid, the crack tip can stretch the atomic bonds to a degree that is much larger than in the rest of the material, leading to the failure of the material. In the case of a nonelastic material, the crack tip can cause the material to deform plastically, leading to the failure of the material. <br><br>Overall, the deformation of the material at the crack tip is highly dependent on the material properties, the loading conditions, and the specific failure mechanism that is occurring. | 0.5 | This sampling temperature yields a bit elaborate answer than for 0.1, and the model will go through related areas and introduce a greater variety of concepts without hallucinating |
| At a crack tip, the deformation of the material is highly localized and unusually large compared to the rest of the material. The stress field near the crack tip causes the material to deform differently than in the rest of the material, leading to a high degree of localization of stress and strain. This means that the material at the crack tip can deform significantly more or less than in the rest of the material, depending on the material properties and the loading conditions. | 1.0 | Here, the model begins to state incorrect facts, such as that the material at the crack tip can deform "more or less than in the rest of the material" |
| At a crack tip, the deformation of the material becomes unhappy. Seriously, though, the deformation of the material at a crack tip is significantly different compared to the rest of the material. The potential energy at the crack tip is no longer described by a single term, but rather by two terms: one associated with elastic energy and the other with fracture surface energy. The elastic energy decreases significantly with respect to the bulk value, while the fracture surface energy is related to the formation of the fracture and the breaking of atomic bonds. Therefore, the deformation of the material at a crack tip is heavily influenced by the fracture surface energy and the elastic energy. | 1.5 | We can see that the response become "unhinged" |

**Table 3:** Key parameters to influence the model behavior.

| Parameter/feature | Description | Effect/example |
|---|---|---|
| Sampling temperature | Temperature seed at which new tokens are sampled from the distribution | Values of T<0.5 result in good knowledge retrieval with minimal hallucination, but larger values can be useful for more creative tasks or to generate out-of-the box ideas and concepts<br><br>$T$=0.5 is found to be a good compromise |
| System prompt | Initial conditioning of the model's behavior | Affects response type, details of answers, how cautious the model acts, etc.; e.g. we can force the model to provide logic and steps, short answers, or long answers, etc.<br><br>*Example #1 (works well)*: This is a conversation between a User and an AI Assistant, MechGPT. MechGPT is knowledgeable in materials failure, in particular atomistic and multiscale modeling. If you don't know the answer to a question, please don't share false information. Keep answers concise and accurate.<br><br>Example #2: You are MechGPT, an expert in materials failure. |
| Prompt | Tasks asked by the User in order to induce a response by the model | The response is encoded heavily by the prompt, and can include a very wide range of commands such as "Summarize", "What is….", "Explore…." and so on |

**Table 4:** Effect of system prompt on model behavior. The prompt is the same of all cases: "**At a crack tip, what happens to the deformation of the material?**" and the sampling temperature is chosen to be $T$=0.3.

| System prompt | Sample output | Notes and observations |
|---|---|---|
| This is a conversation between a User and an AI Assistant, MechGPT. MechGPT is knowledgeable in materials failure, in particular atomistic and multiscale modeling. If you don't know the answer to a question, please don't share false information. Keep answers concise and accurate. | At a crack tip, the deformation of the material is highly localized and can be described using atomistic and multiscale models. These models take into account the behavior of individual atoms and their interactions with the surrounding material. When a crack is present, the material can undergo large deformations, leading to the failure of the material. The failure can occur suddenly, as in the case of brittle materials, or gradually, as in the case of ductile materials. The behavior of the material at the crack tip is influenced by factors such as the material properties, the shape of the crack, and the loading conditions. Understanding the deformation behavior at the crack tip is crucial for predicting and preventing material failure. | This is the standard prompt used for most examples studied in this paper. |
| You are a helpful assistant, MechGPT is knowledgeable in materials failure, in particular atomistic and multiscale modeling. Please answer truthfully and write out your thinking step by step to be sure you get the right answer. If you make a mistake or encounter an error in your thinking, say so out loud and attempt to correct it. If you don't know or aren't sure about something, say so clearly. | When considering the deformation of a material at a crack tip, it is important to understand the behavior of the material near the crack. Atomistic and multiscale modeling techniques can help us to understand the deformation of the material at the crack tip.<br><br>When considering the deformation of a material at a crack tip, it is important to understand the behavior of the material near the crack. Atomistic and multiscale modeling techniques can help us to visualize and analyze the deformation process near the crack tip.<br><br>Step 1: Visualize the material structure near the crack tip. Atomistic and multiscale modeling techniques allow us to see the material structure at a fine scale, revealing the arrangement of atoms and molecules near the crack tip. This helps us to understand the behavior of the material under stress and strain.<br><br>Step 2: Analyze the deformation process near the crack tip. By studying the material structure near the crack tip, we can gain insights into the deformation process that occurs at a crack tip. This includes the movement of atoms and molecules, the development of defects such as dislocations, and the formation of new phases.<br><br>Step 3: Understand the role of interfaces and grain boundaries. In some materials, such as nanocrystalline metals and amorphous materials, the behavior of the crack tip can be influenced by the presence of interfaces and grain boundaries. These features can affect the motion of atoms and molecules near the crack tip, and can impact the deformation process.<br><br>Step 4: Consider the effects of loading and environmental conditions. The deformation process at a crack tip can be influenced by various loading conditions, such as tension, compression, and shear. Additionally, environmental conditions such as temperature, humidity, and pressure can affect the behavior of the material near the crack tip.<br><br>By following these steps, we can gain a better understanding of the deformation process at a crack tip, and can use this knowledge to predict the behavior of materials under different loading and environmental conditions. | This prompt induces the model to provide more steps and logical evolution. In cases where the model has deep knowledge this provides excellent context. However, in cases where we model has higher levels of uncertainties, care must be taken as some of the steps may show inconsistencies. This is an example where understanding the model behavior is key. |
| You are MechGPT, an expert in materials failure. Explain the logical steps behind your answers. | At a crack tip, the deformation of the material is dramatically increased. This is because the crack tip acts as a point source of deformation, causing | |

| | the surrounding material to be compressed or stretched. The stress field around the crack tip leads to a high level of deformation, particularly in the vicinity of the crack tip. This deformation is responsible for the nucleation and propagation of additional cracks, leading to the eventual failure of the material. The key factor in this process is the high level of stress concentration at the crack tip, which drives the material into plastic deformation and failure. | |
| You are MechGPT, an expert in materials failure. | At a crack tip, the deformation of the material is dramatically increased compared to the rest of the material. The stresses can even become large enough to break atomic bonds, leading to the material failing. This is because the elastic energy of the crack tip is much larger than that of the rest of the material, causing the material to deform more quickly and strongly at the crack tip. Additionally, the stresses at the crack tip can be influenced by the material's surface energy, leading to different failure mechanisms such as brittle fracture or ductile failure. The deformation of the material at the crack tip is also limited by the atomic bonds themselves, which can break when the stresses become large enough. This makes the behavior of materials under stress incredibly complex and difficult to predict without proper analysis.<br><br>**Great answer, thank you. Now rewrite the answer so that it can be understood by an elementary school student.**<br><br>At a crack tip, the deformation of the material is really, really fast! Imagine the crack tip as a race car, zooming around a track much faster than the other cars. The rest of the material is like the other cars, going at a normal speed. The crack tip can even break atomic bonds, which are like the building blocks of the material. This can cause the material to fail, like when a toy breaks. There are different ways this can happen, depending on the material and how it's treated. But no matter what, the crack tip always deforms much faster than the rest of the material. | This prompt leads to a rather specialized response. However, further user prompting can be used to ask the model to rephrase responses, clarify issues, or to reframe at a different comprehension level (see in the example, to reframe it as an answer at the level of an elementary school student) |

**Table 5:** Connecting knowledge across different areas of mechanics and materials.

| Example #1: Analogy between hyperelasticity and supersonic fracture and protein unfolding. | Notes |
|---|---|
| **Draw an analogy between hyperelasticity in the context of supersonic fracture and protein unfolding mechanisms.** | |
| In the context of supersonic fracture and hyperelasticity, consider the scenario where a small crack starts to propagate at a speed faster than the speed of sound in the material. This situation is akin to the unfolding mechanism of a protein. Just as a protein undergoes a series of structural changes to transition from its native state to an unfolded state, so too does the material experience a change in structure and behavior during the fracture process. | The model recognizes abstract similarities between both phenomena, but also points out differences. |
| In both cases, the change in state occurs in a complex and dynamic manner, involving the interaction of multiple forces and mechanisms. In the case of protein unfolding, these mechanisms include hydrogen bonding, van der Waals interactions, and other chemical interactions. In the case of supersonic fracture, the mechanisms include the deformation and rupture of chemical bonds within the material. | |
| The analogy between these two phenomena lies in the fact that both involve the change in state of a system under the influence of multiple forces and mechanisms. The difference, however, is that in the case of supersonic fracture, the change in state occurs at a much faster rate and on a much larger scale, with the crack propagation occurring over nanometers in just a few femtoseconds. | |
| Overall, the analogy between hyperelasticity in the context of supersonic fracture and protein unfolding mechanisms highlights the complex and dynamic nature of these phenomena, and the importance of understanding the interplay of multiple forces and mechanisms in determining their outcome. | |
| **For protein unfolding, what is the key physical mechanism?** | |
| For protein unfolding, the key physical mechanism is the deformation and rupture of chemical bonds. This process is driven by the interaction of multiple forces, including hydrogen bonding, van der Waals interactions, and other chemical interactions. The change in state of the protein from its native state to an unfolded state occurs as these forces come into balance, resulting in the rupture of chemical bonds and the transformation of the protein's structure. This process is highly dependent on the protein's environment, including the presence of water molecules and other proteins, and can be influenced by factors such as temperature, pH, and stress. Understanding this mechanism is crucial for developing strategies to control protein unfolding, such as in the development of new drugs and therapies. | Follow-up queries about the mechanisms yield correct knowledge about what is known. |
| **Thank you. Can you elaborate on the role of H-bond cooperativity in the unfolding process?** | |
| Sure! So, H-bond cooperativity plays a crucial role in the protein unfolding process. You see, proteins have this amazing ability to fold into specific three-dimensional structures, which is essential for them to function properly. This folding process is governed by a set of rules, referred to as the "energy landscape." Each protein has an energy landscape that describes the minimum energy state or "free energy state" it prefers to be in. | |
| Now, H-bonds are the chemical bonds that form between atoms in proteins, particularly water molecules. These bonds are strong and contribute significantly to the stability of the protein's folded state. The problem is, these bonds are also quite persistent and don't like to break, which can be a problem when the protein needs to unfold and function in a different way. | |
| So, what does H-bond cooperativity do? Well, it's a bit like a team of experts working together to break into a heavily secured vault. Each H-bond has a specific role to play, and together, they work cooperatively to break the bonds that hold the protein in its folded state. This process is highly coordinated and depends on the balance of forces between the H-bonds and other intermolecular forces. | |
| To give you a better idea of how this looks like, let's take a simple example of a small protein called "alpha-helix." An alpha-helix is formed by a right-handed coil of amino acids and is very stable due to the strong H-bonds between the amino acids and water molecules. To unfold an alpha-helix, the H-bonds need to break, which is a highly coordinated and energy-intensive process. This process is governed by the energy landscape of the protein, which describes the minimum energy state it prefers to be in. | |
| In summary, H-bond cooperativity plays a crucial role in the protein unfolding process by working cooperatively to break the bonds that hold the protein in its folded state. This process is highly dependent on the protein's environment and can be influenced by factors such as temperature, pH, and stress. Understanding this mechanism is essential for developing strategies to control protein unfolding, such as in the development of new drugs and therapies. | |
| **Thank you, fascinating. Now, speculate how H-bond cooperativity may be relevant in supersonic fracture of brittle materials.** | |
| Sure! So, H-bond cooperativity, as we just discussed, is all about the cooperative breaking of bonds, in this case, H-bonds, which are the chemical bonds between atoms in proteins, particularly water molecules. These bonds are strong and contribute significantly to the stability of the protein's folded state. | |
| Now, imagine you have a brittle material, like a piece of glass. When it snaps or cracks, it's not just one bond that breaks, it's a series of bonds that break cooperatively, with each bond contributing to the overall failure of the material. This is where H-bond cooperativity comes in. | |

In the context of supersonic fracture of brittle materials, H-bond cooperativity may play a role in determining the strength of the material and its susceptibility to fracture. The reason is that the bonds between atoms in these materials, particularly those containing oxygen and nitrogen, are also strong and contribute to their stability. So, just like in the case of proteins, these bonds need to break cooperatively in order for the material to fracture.

Now, this breaking process is extremely fast, just a few femtoseconds in the case of supersonic fracture, which is why it's called supersonic. But the fact remains that the bonds need to break cooperatively in order for the material to fracture.

In summary, H-bond cooperativity may be relevant in supersonic fracture of brittle materials by playing a role in determining the strength of the material and its susceptibility to fracture. The breaking process is extremely fast, but it's the cooperative breaking of bonds that ultimately determines the material's fracture behavior.

**Table 6**: Few-shot learning, here used to predict flow stress strength as a function of grain size, following work by Schiotz et al. We use a system prompt "You identify patterns and find solutions by interpolating the ground truth data." Sampling temperature $T$=0.5. We prompt the model to predict the strength for a grain size not included in the training set in each case. A summary of the results and comparison with ground truth data is shown in **Figure 5**. The results show that the model understands the unique size effects at play here – smaller sizes leading to increasing strength up to a limit, then decreasing – and that it can predict new values reasonably well. The original data from MD simulation is taken from [36].

| Conversation sample |
| --- |
| **Consider these ground truths about the strength as a function of grain size: "d = 5 nm, S = 1.78 GPa, d = 7 nm, S = 1.9 GPa, d = 15 nm, S = 2.3 GPa, d = 20 nm, S = 2.2 GPa, d = 30 nm, S = 2 GPa, d = 38 nm, S = 1.9 GPa, d = 50 nm, S = 1.8 GPa ". Estimate the answer to d = 4 nm.** <br><br> To estimate the strength as a function of grain size, we interpolate the ground truth data using the concept of "pattern recognition." We identify patterns in the data, such as the relationship between grain size and strength, and use this information to predict the strength at a specific grain size. In this case, we estimate the strength at d = 4 nm by interpolating the data provided. The resulting strength is S = 1.5 GPa. |
| **Consider these ground truths about the strength as a function of grain size: "d = 5 nm, S = 1.78 GPa, d = 7 nm, S = 1.9 GPa, d = 15 nm, S = 2.3 GPa, d = 20 nm, S = 2.2 GPa, d = 30 nm, S = 2 GPa, d = 38 nm, S = 1.9 GPa, d = 50 nm, S = 1.8 GPa ". Estimate the answer to d = 18 nm.** <br><br> To estimate the strength as a function of grain size, we interpolate the ground truth data using the method of least squares. We find that the strength at a grain size of d = 18 nm is approximately S = 2.05 GPa. This value is based on the fact that the strength decreases with increasing grain size, and the relationship between strength and grain size is nonlinear. The solution is obtained by interpolating the ground truth data using a least-squares fit, which provides a accurate estimate of the strength at d = 18 nm. |
| **Consider these ground truths about the strength as a function of grain size: "d = 5 nm, S = 1.78 GPa, d = 7 nm, S = 1.9 GPa, d = 15 nm, S = 2.3 GPa, d = 20 nm, S = 2.2 GPa, d = 30 nm, S = 2 GPa, d = 38 nm, S = 1.9 GPa, d = 50 nm, S = 1.8 GPa ". Estimate the answer to d = 21 nm.** <br><br> To estimate the strength as a function of grain size, we interpolate the ground truth data using the method of least squares. We find that the strength of the material at a grain size of d = 21 nm is approximately S = 2.1 GPa. This relationship can be used to predict the strength of the material for various grain sizes. |
| **Consider these ground truths about the strength as a function of grain size: "d = 5 nm, S = 1.78 GPa, d = 7 nm, S = 1.9 GPa, d = 15 nm, S = 2.3 GPa, d = 20 nm, S = 2.2 GPa, d = 30 nm, S = 2 GPa, d = 38 nm, S = 1.9 GPa, d = 50 nm, S = 1.8 GPa ". Estimate the answer to d = 100 nm.** <br><br> To estimate the strength as a function of grain size, we can interpolate the ground truth data using a linear interpolation method. We can do this by finding the slope of the line that best fits the data points and using it to estimate the strength at a specific grain size. <br><br> In this case, we can see that the strength decreases with increasing grain size, and the relationship is roughly linear. By finding the slope of the line that best fits the data points, we can estimate the strength at d = 100 nm. <br><br> The strength at d = 100 nm is approximately 1.5 GPa. This indicates that the strength of the material decreases as the grain size increases. |

**Table 7**: Using the longer context length of the MechGPT-70b-XL model, we now answer questions about a paper using a few-shot learning, implementing a Retrieval Augmented Generation (RAG) strategy. We use this system prompt: "You answer questions about this paper: {FULL TEXT OF PAPER}. Here, {FULL TEXT OF PAPER} is a full text copy of a scientific paper, in our case: the paper titled "Bioinspired Graphene Nanogut" [39]. The initial token length of ~5,300 tokens for just the paper, requiring the use of a model with high context length. The example shows that the model can easily handle such very long context lengths and answer precise questions from the paper. This type of modeling can provide powerful complements to the use of fine-tuning.

| Prompt/query | Notes |
|---|---|
| **Who are the authors of this paper, names only.**<br><br>The authors of the paper are Qin Zhao and Markus J. Buehler. | |
| **Write the equation that allows me to calculate the diameter of the CNT as a function of (n, m). Brief answer.**<br><br>The diameter of the CNT can be calculated by d=√(3(n^2+m^2+nm)) a_CC/π, where (n, m) is the chiral vector of the CNT and a_CC=1.42 Å is the length of C-C bond at equilibrium. | **Correct answer. From the paper:**<br><br>The diameter of the CNT can be calculated by $d = \sqrt{3(n^2 + m^2 + nm)}a_{CC}/\pi$ where $(n, m)$ is the chiral vector of the CNT and $a_{CC} = 1.42$ Å is the length of C-C bond at equilibrium. |
| **What is the mechanism by which CNTs and graphene are connected?**<br><br>The mechanism by which CNTs and graphene are connected is through the use of hydroxyl groups as glue. The hydroxyl groups are added to the edges of the graphene ribbon and the quarter of the CNT surface close to the graphene ribbon, allowing the graphene ribbon and graphene to be combined and form a wavy shape in equilibrium. The hydroxyl groups facilitate the formation of hydrogen bonds between the CNT and the graphene ribbon, which leads to CNT buckling along its axis and the transition from a straight to a wavy shape. The number of hydrogen bonds that form between the CNT and the graphene ribbon does not change significantly as a function of time, suggesting that they have achieved their function as glue. This mechanism is driven by energy and is a result of the buckling of CNTs in their cross-section profiles, which alters the cross-sectional profile of the CNT and leads to a decreased moment of inertia and increased radius of the loop at the maximum amplitude. The buckling phenomenon of the CNT in its cross-section does not emerge in macroscopic biological materials or in corresponding rubber models, and is a result of the long-range nonlinear region with a small stiffness in the force extension curve before stiffening at larger strains. Additional work is needed to address these concepts and investigate the behavior of the cross-section of the tube under deformation. | |
| **Describe how the wavy geometry of CNTs in equilibrium can be controlled, according to this paper.**<br><br>The wavy geometry of CNTs in equilibrium can be controlled by adding hydroxyl groups on CNTs and graphene, which act as a glue to combine those nanoscale components. By controlling the combination of the geometry of the CNT and prescribed strain in graphene, the wavy geometry of CNTs can be manipulated. The study illustrates the potential to use nanoscopic carbon materials to fabricate biomimicking nano-devices by exploiting prescribed mismatch strain and the particular mechanical features of nanoscale structures. | |
| **Outline how a nano-gut could be fabricated from CNTs and molybdenum disulfide.**<br><br>A nano-gut could be fabricated from CNTs and molybdenum disulfide by combining the unique mechanical properties of these materials to create a structure that mimics the morphology of a biological gut. The CNTs would provide the structural support and strength, while the molybdenum disulfide would provide the flexibility and deformability. The combination of these materials would allow for the creation of a structure that can bend and twist, yet remain stable and robust.<br><br>To fabricate the nano-gut, the CNTs would be first functionalized with hydroxyl groups, which would act as glues to integrate the CNTs and molybdenum disulfide into a single structure. The hydroxyl groups would form a network of bonds between the CNTs and molybdenum disulfide, creating a continuous and cohesive structure. The geometry of the nano-gut would be controlled by the combination of the geometry of the CNTs and the prescribed strain in the molybdenum disulfide.<br><br>The nano-gut would exhibit a wavy shape in equilibrium, with a contour length of the loop period that is only a function of the CNT diameter and a radius of the loop at the maximum amplitude that is a function of the CNT diameter as well as the applied strain. The deformation mechanisms of the nano-gut would be driven by energy, with the bending moment in the CNTs causing buckling of the cross-section profile.<br>The nano-gut could be used as a building block for bioimaging or biosensing devices, or as carriers for drug delivery. The particular mechanical features of the nanoscale structures would allow for the creation of devices with unique properties and capabilities. Further work is needed to address the fundamental mechanics issues and to optimize the fabrication process, but the potential applications of nano-guts are numerous and exciting. | |
| **What is a nano-gut?** | |

A nano-gut refers to a nanoscale structure that mimics the morphology and functionality of the biological gut, specifically the gut tube and mesentery. In this context, the term is used to describe a combined system of carbon nanotubes (CNTs) and graphene ribbons that exhibit a wavy shape due to the combination of CNT buckling and graphene stretching. The CNTs represent the gut tube and the graphene ribbon represents the thin film part, with hydroxyl groups acting as glue to integrate the two components. The study of this structure focuses on the equilibrated morphology of the system and the mechanisms that control its wavy geometry, including the deformation of the CNTs and the effects of temperature and geometry on the contour length and radius of the loop at the maximum amplitude. The study illustrates the potential to use nanoscopic carbon materials to fabricate biomimicking nano-devices by exploiting prescribed mismatch strain. The wavy shape of the CNT-graphene system is driven by energy, with buckling of the CNTs in their cross-section profile being a significant factor in the deviation from the linear relationship predicted from continuum mechanics. The study highlights the importance of considering the constitutive laws for carbon nanomaterials in the design and manufacturing of nanotechnology, as the buckling phenomenon of the CNT in its cross-section does not emerge in corresponding rubber models or biological materials, and the long-range nonlinear region with a small stiffness in the force extension curve before stiffening at larger strains contributes to this phenomenon. The use of hydroxyl groups as glue between the CNT and graphene ribbon is also plays a role in the wavy shape of the system. The study illustrates the potential to use nanoscopic carbon materials to fabricate biomimaging or biosensing devices or carriers for drug delivery, with CNTs with a gut-like geometry serving as building blocks for these devices. The results show that the geometric effect is more significant compared to thermal fluctuations, and the relationship between the combination of CNT diameter and applied strain (d and ε) and the characteristic parameters of the wavy geometry (λ and R) is depicted in Fig. 3. The deformation mechanisms of the CNT-graphene system are analyzed to understand the reason for the deviation from the linear relationship predicted from continuum mechanics, revealing that the wavy shape is driven by energy and the compression force generated from the graphene ribbon via hydrogen bonds between the CNT and graphene. The buckling phenomenon of the CNT in its cross-section does not emerge in macroscopic biological materials or corresponding rubber models, and is a result of the long-range nonlinear region with a small stiffness in the force extension curve before stiffening at larger strains. Additional work is needed to address fundamental mechanics issues, such as the assumption of a constant moment of inertia and the behavior of the material under bending deformation. The study demonstrates the potential to use nanoscopic carbon materials to fabricate biomimaging or biosensing devices or drug delivery carriers by controlling the morphology of nanomaterials. The study illustrates the ability to downscale the phenomenon of mismatch strain induced by differential growth to the biological gut by several orders of magnitude in length-scales, scaling down to the phenomenon by several orders of magnitude in length-scales. The study provides insight into the deformation mechanisms and the wavy geometry of the CNT-graphene ribbon system and the reason for the deviation from the linear relationship predicted from continuum mechanics. The study highlights the potential to use nanoscopic carbon materials to fabricate biomimaging or biosensing devices or carriers for drug delivery, and the use of hydroxyl groups as glue to integrate the CNT and graphene ribbon. The study also shows that the ability to design materials with enhanced performance by using the combination of CNT and graphene, with wide-ranging applications in various fields including space technology, automotive development, and civil engineering. The study illustrates the possibility to design materials with enhanced performance by using the combination of CNT and graphene, and the ability to tailor their morphology by controlling the geometry of CNTs and graphene, with the deviation from the linear relationship predicted from continuum mechanics being caused by cross-section buckling of CNTs. The study also shows the potential to use nanoscopic carbon materials to fabricate biomimaging or biosensing devices.

**Table 8:** Summary of analogies between attention-based language modeling via a LLM, and many-particle simulations. Other relationships could be developed, but this serves as an initial starting point to emphasize the role of the attention mechanism to discover foundational relationships that govern a system's performance similar to the role of an interatomic potential or a constitutive law in mechanics.

| LLM | Many-particle simulation | Abstract concept |
|---|---|---|
| Attention mechanism | Interatomic potential | "nano" level interactions (ultimate level) |
| Attention graph | Interactions between all atoms in the system, leading to forces, energies, etc. | "meso" level interactions (higher scale) |
| System prompt, context and question | Initial condition and boundary condition | Boundary condition/problem statement |
| Predicted output, e.g. text, numbers, etc. | Predicted output, e.g. atomic positions, energy, optical, mechanical properties, etc. | Solution |

# MechGPT, a language-based strategy for mechanics and materials modeling that connects knowledge across scales, disciplines and modalities

Markus J. Buehler[1,2]*

[1] Laboratory for Atomistic and Molecular Mechanics (LAMM), Massachusetts Institute of Technology, 77 Massachusetts Ave., Cambridge, MA 02139, USA

[2] Center for Computational Science and Engineering, Schwarzman College of Computing, Massachusetts Institute of Technology, 77 Massachusetts Ave., Cambridge, MA 02139, USA

*mbuehler@MIT.EDU

# SUPPLEMENTARY INFORMATION

**Table S1:** Sample responses to assess knowledge retrieval

| Conversation sample | Notes |
|---|---|
| **Describe why nanomaterials are stronger, in a succinct and accurate way. Explain the answer step by step.** <br><br> The strength of nanomaterials is due to the changes in the defect structure. As the size of materials decreases, the defect structure changes, leading to a new type of material with unique properties. <br> The presence of defects, such as voids and dislocations, can contribute to the material's strength. By having a high density of defects, the material can be stronger. <br><br> The interaction between atoms and molecules is crucial in determining the strength of the material. At the nanoscale, these interactions become more pronounced, and the material's properties can be tailored for specific applications. <br><br> Overall, the strength of nanomaterials is a complex phenomenon that is influenced by various factors, including the size of the material, the presence of defects, and the interaction between atoms and molecules. Understanding these factors can help in designing stronger nanomaterials for various applications. <br><br> **Thank you. This is very interesting. Can you help me understand if or how these mechanisms are seen in biological materials?** <br><br> MechGPT: With biological materials, the strength is often a result of the arrangement of molecules and the presence of voids. For instance, in bones, the arrangement of collagen and proteins with microscopic voids plays a crucial role in determining their strength. The interaction between molecules and the presence of voids help in distributing the load, leading to a higher strength. Additionally, nature has evolved specific structures, such as nano-sized biological materials, to achieve certain functions. Understanding these mechanisms can help in designing stronger biological materials for various applications. | |
| **Materials failure is a critical area in engineering and was used for a range of topics ranging from airplane design to nanotechnology. Now describe how such methods may be used in the field of battery research.** <br><br> MechGPT: Of course! Materials failure methods are extremely useful in the field of battery research, particularly in the context of lithium-ion batteries. One important aspect is the protection of cathodes from oxidation, as unprotected cathodes can undergo rapid oxidation and loss of functionality. Atomistic modeling can help us understand the mechanisms of oxidation and discharge processes, allowing us to optimize the design of batteries for improved performance and durability. <br><br> Another important application of materials failure methods in battery research is the study of mechanical deformation and fracture of batteries. As batteries are used in a wide range of devices, understanding the mechanical behavior of batteries under different conditions is crucial for their safe and efficient operation. Multiscale modeling techniques can help us simulate the mechanical deformation and fracture of batteries, allowing us to design and optimize their structure and composition for improved mechanical performance. <br><br> Furthermore, materials failure methods can be used to study the effects of aging and cycling on the mechanical behavior of batteries. As batteries are subjected to repeated charging and discharging, their mechanical properties can change over time, potentially affecting their performance and safety. Understanding these changes is essential for the development of more reliable and long-lasting batteries. <br><br> In summary, materials failure methods, particularly atomistic and multiscale modeling, play a crucial role in the field of battery research. They allow us to understand the mechanisms of oxidation, optimize the design of batteries for improved performance and durability, simulate mechanical deformation and fracture, and study the effects of aging and cycling on battery performance. By understanding and applying these methods, researchers can develop more efficient, safe, and long-lasting batteries for a wide range of applications. <br><br> **Propose a few specific multiscale modeling techniques.** <br><br> Coupled atomistic and continuum approaches: In this method, the region of interest is modeled at the atomistic scale, while the rest of the system is described using a continuum description. This approach allows for the capture of the fundamental deformation mechanisms in the region of interest, while maintaining a computationally efficient description of the rest of the system. <br><br> Hierarchical multiscale method: In this method, the problem is divided into different hierarchical levels, each described at a different scale. The most detailed level, level 1, is described at the atomistic scale, while the least detailed level, level 4, is described using a continuum description. This method allows for a flexible and efficient description of the system, with the ability to focus on specific regions and processes at different scales. <br><br> Parallel computing: Multiscale modeling techniques can be implemented using parallel computing methods to efficiently simulate large-scale systems. By distributing the simulation across multiple processors and computers, the simulation time can be significantly reduced while maintaining the accuracy of the results. <br><br> These techniques can be used to study a wide range of problems in battery research, from the protection of cathodes to the mechanical deformation and fracture of batteries. By understanding and applying these methods, researchers can develop more efficient, safe, and long-lasting batteries for a wide range of applications. | |
| **I am studying a ultra-thin copper film on a substrate at elevated temperature. What type of deformation mechanisms do you expect?** | |
| Dislocation creep - At elevated temperatures, dislocation creep becomes active as the energy required for dislocation motion decreases. This leads to the motion of dislocations and the resulting deformation of the material. | |
| Grain boundary diffusion - At high temperatures, grain boundary diffusion becomes active due to the reduced energy barrier for diffusion. This can lead to significant deformation of the material, particularly at the grain boundaries. | |
| Surface diffusion - At very high temperatures, surface diffusion can also become active, leading to further deformation of the material. This mechanism is less common and typically requires even higher temperatures to occur. | |
| These deformation mechanisms are expected to occur in the ultra-thin copper film on a substrate at elevated temperature. The specific behavior will depend on the film thickness, temperature, and other factors. | |
| **Succinctly describe key differences between plasticity in bulk copper and the behavior of ultra-thin copper films confined by a substrate.** | |
| In bulk copper, plasticity is defined by the movement of dislocations through the material. | |
| Ultra-thin copper films confined by a substrate exhibit different plasticity characteristics, primarily due to the film's two-dimensional nature and the presence of a film-substrate interface. | |
| The film-substrate interface plays a crucial role in governing the material's plasticity, leading to effects such as surface diffusion and the flow of matter from the film top surface into grain boundaries. | |
| The mechanism of dislocation nucleation from the film-substrate interface is different from that in bulk copper, with threading dislocations becoming more important in thin films. | |
| The existence of parallel glide dislocations is observed when the film is subjected to very large stresses, causing breakdown of the continuum model used in the multiscale method. | |
| These differences between bulk copper and ultra-thin films confined by a substrate are due to the film's two-dimensional nature, the presence of an interface, and the mechanism of plasticity nucleation. The multiscale method allows for the simulation of these effects, providing valuable insights into the behavior of ultra-thin copper films. | |
| **At a crack tip, what happens to the deformation of the material?** | This is an example with a clarification question since the answer was confusing. The model had figured out correctly that crack surfaces are relaxed with respect to traction stresses; however, does not capture some of the correct and clear language to describe it. |
| At a crack tip, the deformation of the material is significantly reduced compared to the surrounding material. The crack tip itself exhibits a high degree of localization of deformation, with the material failing to deform uniformly. This is due to the large stresses present at the crack tip, which cause the material to deform nonlinearly. As a result, the crack tip experiences a high level of stress concentration, leading to the propagation of additional cracks and the eventual failure of the material. | |
| **This sounds confusion - how can deformation be reduced yet show a high degree of localization?** | |
| Sorry for the confusion! What I meant to say is that the deformation of the material is reduced at the crack tip itself, but the stress concentration at the crack tip leads to a high degree of localization of deformation. This stress concentration causes the material to deform nonlinearly, resulting in the failure of the material. Let me know if you have any further questions! | |
| **Explain why deformation of the material is reduced at the crack tip itself. I thought deformation is high at the crack tip.** | |
| Assistant: The deformation of the material is reduced at the crack tip itself because the stresses present at the crack tip are very large, leading to a high degree of localization. This stress concentration causes the material to deform nonlinearly, which in turn reduces the overall deformation of the material. Additionally, the crack tip itself experiences a high level of stress concentration, which can cause the material to fail. This failure prevents the material from deforming further, thus reducing the overall deformation of the material at the crack tip. | |
| **Thank you. Just to clarify, you are saying that because of the stress relaxation at the crack tip during failure, deformation is reduced?** | |
| Assistant: Yes, that's right! The stress relaxation at the crack tip during failure leads to a reduction in the deformation of the material. The material fails to deform further because the stresses at the crack tip are too high, causing the material to break down. This prevents the material from deforming any further, resulting in a reduced deformation at the crack tip. | |

**Table S2**: Examples for research summary, development and idea generation.

| Conversation sample #1: Titanium alloy | Notes |
|---|---|
| **Consider a nanocrystalline material made of a titanium alloy, at elevated temperature. If I wanted to model its deformation behavior, what methods would be most suitable?** | |
| • Continuum methods, such as finite element analysis, to capture the material's macroscopic deformation behavior.<br>• Atomistic methods, like molecular dynamics simulations, to describe the behavior of individual atoms and their interactions.<br>• Coupling methods, to combine both continuum and atomistic approaches to model the material's deformation at multiple scales.<br>• These methods can be used to study the material's deformation mechanisms, such as dislocation motion and hardening, and help understand the material's behavior under different loading conditions and temperatures. | These methods are generally sound and reflect a good overview of the reasonable tools that can be used. |
| **For the material at hand, a titanium alloy, what mechanisms of deformation would you expect?** | |
| • Dislocation motion: In a nanocrystalline material like a titanium alloy, dislocation motion is expected due to the high density of defects that can interact with each other and cause deformation.<br>• Hardening: The material may exhibit increased strength due to plastic deformation due to the high concentration of defects.<br>• Deformation twinning: Twinning, which is the nucleation of parallel glide dislocations, may occur in the material to help it deform plastically.<br>• Fracture: The material may also undergo fracture, or the breaking of bonds between atoms, which is a general hardening mechanism that occurs when the material is subjected to high stresses.<br>• These deformation mechanisms can be modeled using various methods, as described earlier, to understand the material's behavior under different loading conditions and temperatures. | |
| **What interatomic potentials would be best suited for a titanium alloy?** | |
| • Embedded Atom Method (EAM): This method is particularly useful for transition metals and has been shown to accurately describe the behavior of titanium and other transition metals.<br>• Classical force fields: If you want to model the material at a more coarse-grained level, classical force fields such as LJ, Morse, and AFM can be used. These force fields are simpler and faster to compute but may not capture the more complex deformation mechanisms of a titanium alloy.<br>• Quantum mechanical methods: If you want to capture the material's behavior at the atomic level, quantum mechanical methods such as DFT can be used. These methods are more computationally intensive but can provide a more detailed understanding of the material's deformation mechanisms.<br><br>Overall, the choice of interatomic potentials and methodologies depends on the desired level of accuracy, computational resources, and the specific application of the material. By understanding the deformation mechanisms of a titanium alloy and using appropriate methodologies, you can model and simulate the material's behavior in various situations. | Here, the model mentions a "AFM" potential, which does not exist. Later on, a follow-up question is asked to clarify and this sheds light on what the context is. |
| **Thank you, this is very helpful. I had a follow-up question - you mentioned AFM and referred to it as a classical force field, but I thought that this acronym refers to Atomic Force Microscopy. I don't think this is correct, can you clarify. If you are not sure about the answer, clearly state so.**<br><br>Sorry for the confusion! I used AFM (Atomic Force Microscopy) as an example of a method that provides coarse-grained information about the deformation of materials at the atomic scale. Atomic Force Microscopy is a technique that measures the force between an AFM probe and a sample surface, which can be used to characterize the topography and mechanical properties of materials.<br><br>In the context of modeling deformation behavior, AFM can be a useful tool for studying the behavior of materials at the microscopic level. By understanding the deformation mechanisms of a titanium alloy and using appropriate methodologies, you can model and simulate the material's behavior in various situations.<br><br>I hope this clarifies the use of AFM in the context of modeling deformation behavior! Let me know if you have any further questions or if there is anything else I can help with. | The follow-up query to clarify what AFM refers to provides reasonable context, albeit the way this concept was presented is out of place in the list of other force fields like LJ and Morse. Still, the idea to use AFM as a way to measure coarse-level force behavior is a reasonable proposition. |
| **Conversation sample #2: Protein mechanics study, question asking about detail, and translation** | **Notes** |
| Consider this abstract: "Proteins often form biomaterials with exceptional mechanical properties equal or even superior to synthetic materials. Currently, using experimental atomic force microscopy or computational molecular dynamics to evaluate protein mechanical strength remains costly and time-consuming, limiting large-scale de novo protein investigations. Therefore, there exists a growing demand for fast and accurate prediction of protein mechanical strength. To address this challenge, we propose PRESTO, a rapid end-to-end deep learning (DL) model to predict protein resistance to pulling directly from its sequence. By integrating a natural language processing model with simulation-based protein stretching data, we first demonstrate that PRESTO can accurately predict the maximal pulling force, Fmax, for given protein sequences with unprecedented efficiency, bypassing the costly steps of conventional methods. Enabled by this rapid prediction capacity, we further find that PRESTO can successfully identify specific mutation locations that may greatly influence protein strength in a biologically plausible manner, such as at the center of polyalanine regions. Finally, we apply our method to design de novo protein sequences by randomly mixing two known sequences at varying ratios. Interestingly, the model predicts that the strength of these mixed proteins follows up- or down- | Here we provide the model with a summary of research and ask a specific question about the use of a method, here, the Finite Element method. The answer is correct. |

**opening "banana curves", constructing a protein strength curve that breaks away from the general linear law of mixtures. By discovering key insights and suggesting potential optimal sequences, we demonstrate the versatility of PRESTO primarily as a screening tool in a rapid protein design pipeline. Thereby our model may offer new pathways for protein material research that requires analysis and testing of large-scale novel protein sets, as a discovery tool that can be complemented with other modeling methods, and ultimately, experimental synthesis and testing." Did the authors use finite element modeling in the work?**

The authors did not use finite element modeling in the work. They developed a rapid end-to-end deep learning (DL) model called PRESTO, which can predict protein resistance to pulling directly from its sequence. The model integrates a natural language processing model with simulation-based protein stretching data and can quickly identify specific mutation locations that may greatly influence protein strength. The authors applied PRESTO to design de novo protein sequences and discovered key insights into protein material research. Instead of using finite element modeling, the authors relied on DL models to achieve rapid prediction of protein mechanical strength.

*In addition to giving the right answer, the model provides reasoning behind the response.*

**Thank you. Now summarize in one sentence.**

The authors developed a deep learning model called PRESTO to rapidly predict protein mechanical strength and design de novo protein sequences, bypassing the costly steps of conventional methods and offering new pathways for protein material research.

*The final steps include a one-sentence summary and translation to a different language.*

**Now translate to German.**

Danke. Jetzt im Deutschen:

Die Autoren haben ein künstliches neuronales Netzwerk namens PRESTO entwickelt, um proteinische Mechanischeigenschaften schnell zu bestimmen und neue Proteine zu entwerfen, indem sie die teuren Schritte konventioneller Methoden umgehen. PRESTO bietet neuen Pfaden für die Proteinforschung, indem es schnelle Vorhersagen von Proteinmechanikalstärke ermöglicht, ohne die teuren Schritte konventioneller Methoden zu benutzen.

**Table S3:** Creative applications of the model that transcend domains of knowledge.

| Conversation sample #1: New materials discovery based on creative writing | Notes |
|---|---|
| **Write a creative story for an English class.**<br><br>Once upon a time, in a land far, far away, there lived a curious and adventurous young girl named Lily. Lily had a passion for exploring the world around her and was always on the lookout for new discoveries.<br><br>One sunny day, as Lily was wandering through a lush, green forest, she stumbled upon a mysterious, glimmering object hidden among the trees. Captivated by its beauty, Lily reached out to touch it, only to find that it was made of a shimmering, transparent material she had never seen before.<br><br>Intrigued, Lily decided to take the object home with her and show it to her family and friends. As she held it up to the light, she noticed that it refracted the sunrays into a rainbow of colors, leaving everyone who saw it spellbound.<br><br>News of the mysterious object quickly spread throughout the village, and soon everyone wanted to know more about it. People began to call it "The Dreamcatcher," as they believed it had the power to capture dreams and grant wishes.<br><br>Over time, The Dreamcatcher became the talk of the town, and Lily found herself at the center of a whirlwind of excitement and curiosity. She decided to embark on a journey to uncover the secrets of The Dreamcatcher and share its magic with the world.<br><br>Accompanied by her loyal dog, Dusty, and a mischievous cat named Whiskers, Lily set out to explore the forest in search of the source of The Dreamcatcher. As they ventured deeper into the unknown, they encountered all manner of magical creatures and wondrous sights, each more extraordinary than the last.<br><br>Together, Lily, Dusty, and Whiskers faced many challenges and overcame countless obstacles, all in the pursuit of The Dreamcatcher's secrets. But through it all, their friendship only grew stronger, and they remained determined to unravel the mystery that had captured their hearts.<br><br>In the end, after countless days and nights of adventure, Lily and her companions stumbled upon a hidden grotto, deep within the heart of the forest. There, nestled among the rocks, they found the source of The Dreamcatcher - a magical tree, whose branches reached out to the sky and whose leaves glistened with the light of a thousand stars.<br><br>Mesmerized by the tree's beauty, Lily reached out to touch it, only to find that its bark was warm and inviting. As she caressed its trunk, she felt a surge of energy pulse through her body, and she knew that she had found the source of The Dreamcatcher's power.<br><br>With a newfound sense of purpose and a renewed sense of adventure, Lily, Dusty, and Whiskers set out to share the magic of The Dreamcatcher with the world. And as they journeyed forth, their story became a legend, passed down through the generations and inspiring countless others to follow their dreams and embrace the wonders of the world around them.<br><br>**This is great. Now expand the story to add an analysis of the "Dreamcatcher" via multiscale modeling on large supercomputers.**<br><br>Lily and her companions traveled far and wide, sharing the magic of The Dreamcatcher with people from all walks of life. As word of their adventures spread, scientists and researchers became increasingly curious about the mysterious object and its origins.<br><br>Intrigued by the possibility of unlocking the secrets of The Dreamcatcher, a team of scientists decided to study it using multiscale modeling on large supercomputers. They believed that by analyzing the object at multiple scales - from the atomic level to the macroscopic level - they could gain a deeper understanding of its properties and potential applications.<br><br>Over the course of several months, the scientists conducted extensive experiments and simulations, collecting vast amounts of data on The Dreamcatcher's behavior under various conditions. They measured its strength, flexibility, and optical properties, and even tested its resistance to various environmental factors, such as temperature, humidity, and UV radiation.<br><br>In the end, after months of analysis and data processing, the scientists presented their findings to the world. They discovered that The Dreamcatcher was made of a unique material with extraordinary properties, which they dubbed "Luminium." Luminium was incredibly strong and lightweight, with the ability to refract light in a spectacular fashion.<br><br>The scientists also found that The Dreamcatcher had the power to absorb and store energy from its surroundings, making it an incredibly efficient source of power. This discovery opened up a world of possibilities for new technologies and applications, from renewable energy systems to advanced materials science.<br><br>As word of their discoveries spread, people from all over the world became captivated by the story of The Dreamcatcher and its creator, Lily. Inspired by her adventures and the potential of Luminium, a new era of innovation and discovery began, transforming the world forever. | |

And so, the story of The Dreamcatcher and its creator, Lily, lives on, serving as a testament to the power of curiosity, adventure, and the human spirit.

**Now, hypothesize the specific atomistic and multiscale structure of Luminium.**

Based on our analysis, we hypothesize that Luminium is a unique material composed of a complex, multiscale structure. At the atomic level, Luminium is made up of a combination of lightweight and strong materials, which contribute to its extraordinary properties.

At the macroscopic level, Luminium exhibits a distinct crystalline structure, which is responsible for its ability to refract light so efficiently. This crystalline structure also plays a role in the material's ability to absorb and store energy from its surroundings.

In addition to its atomic and macroscopic components, Luminium also possesses a "mesoscopic" structure, which refers to the arrangement of molecules and nanostructures within the material. This mesoscopic structure helps to govern the movement of electrons and other particles within the material, contributing to its overall properties and behavior.

Overall, the multiscale structure of Luminium makes it an incredibly efficient and versatile material, with the potential to revolutionize various fields, from energy production to materials science. By understanding and harnessing the power of Luminium, we can create a brighter, more sustainable future for all.

**That's amazing. Now, my last question, what atomic elements are the most prevalent in Luminium?**

While we do not have the exact composition of Luminium yet, we can hypothesize about the most prevalent atomic elements based on its properties and behavior. Luminium is incredibly strong and lightweight, with the ability to refract light efficiently, which suggests that it is composed of lightweight and strong materials.

Atomic elements such as Hydrogen, Carbon, Oxygen, and Silicon are likely to be prevalent in Luminium, as they are common in many strong and lightweight materials. Additionally, other elements such as Aluminum, Magnesium, and Titanium may also be present in significant amounts, contributing to the material's overall properties.

It is important to note that the exact composition of Luminium will depend on the specific manufacturing process and conditions under which it was created. However, understanding the prevalent atomic elements can help to guide our efforts in replicating and harnessing the power of Luminium for the betterment of society.

**Conversation sample #2: Example of a poem about fracture mechanics**

**Write a poem about fracture mechanics, in the style of Goethe.**

Tale of fracture, so grand and sublime,
Of atoms that dance and collide.
A tale of strength, a test of design,
When failure becomes the destiny.

Atoms that bond, so strong and true,
Their story unravels with time.
As stress mounts and tension grows,
Their union, alas, falls apart.

A crack that begins, so slight and thin,
A scar that marks the end.
The tale of fracture, a story of sin,
When atoms betray the bonds of friend.

The poet Goethe once wrote,
Of a tale so grand and true.
But now, the poet MechGPT writes,
Of a tale so small and new.

A tale of fracture, a dance of atoms,
A test of strength and design.
A tale that echoes through time,
A test that challenges us all.

**Conversation sample #3: Discovery of an analogy between fracture of a ductile material and music, especially discussing the concept of counterpoint**

**Draw an analogy between fracture of a ductile material and music, especially discussing the concept of counterpoint.**

Fracture of a ductile material is like music, specifically the concept of counterpoint. In music, counterpoint refers to the interaction of melodic lines, where each line has a distinct character and contributes to the overall harmony. In the fracture of a ductile material, the material fails through the interaction of many thousands of defects, each playing a unique role in the fracture process.

Just as in music, the sum of the individual defects is greater than the parts, and the material fails when the combined effects of the defects exceed the material's strength. This is akin to the concept of "voice leading" in music, where the progression of chords is achieved through the sequential movement of notes and chords, creating a harmonious sound.

In the case of ductile materials, the failure process is controlled and organized, with the material "listening" to itself and adjusting its behavior to ensure a controlled fracture. This is in contrast to the much more violent and sudden failure of brittle materials, which is characterized by the shear stress and energy release rate becoming too large for the material to withstand.

In summary, the analogy between fracture of a ductile material and music, particularly counterpoint, lies in the interaction of many individual elements that contribute to the overall failure process. Just as in music, the sum of the individual defects is greater than the parts, and the material fails when the combined effects exceed its strength.

**Conversation sample #4: CNT self-folding with specific aspects to be incorporated in the poem.**

**Write a poem about carbon nanotube self-folding, in the style of Shakespeare. Include concepts of disperse interactions, temperature and aspect ratio.**

Carbon nanotubes, so tiny and strong,
Self-folding, a marvelous song.
Disperse interactions, temperature too,
All play a part in what they do.

With length and strength, they have their say,
Forming shapes that defy the day.
A dance of carbon, a wondrous thing,
As science meets poetry, the world it brings.

In the style of Shakespeare, we write this tale,
Of carbon nanotubes and their magical bale.
A poem to honor, a story to tell,
The beauty and strength of a tiny, tiny shell.

**Tell me what "Forming shapes that defy the day" refers to?**

"Forming shapes that defy the day" refers to the amazing ability of carbon nanotubes to self-fold into unique shapes under the influence of disperse interactions, temperature, and their own characteristics. This poem personifies the carbon nanotubes, describing them as having the ability to form shapes that challenge the normal limitations of matter. The poem is written in the style of Shakespeare, with a focus on the beauty and strength of these tiny carbon structures.

The model correctly associates key aspects of the unusual nature of the self-folded CNTs, relation to length and strength (of interactions), and the dynamical nature of the phenomenon.

**Table S4:** Examples that show how to apply knowledge to new areas via the use of prompting from literature search. In these examples, a literature search is conducted via the Google Scholar/PubMed function in the MechGPT app, then used to develop further tasks.

| Conversation sample #1: Protein-based actuator literature | Notes |
|---|---|
| **Consider this text: "Bioinspired actuators with stimuli-responsive and deformable properties are being pursued in fields such as artificial tissues, medical devices and diagnostics, and intelligent biosensors. These applications require that actuator systems have biocompatibility, controlled deformability, biodegradability, mechanical durability, and stable reversibility. Herein, we report a bionic actuator system consisting of stimuli-responsive genetically engineered silk-elastin-like protein (SELP) hydrogels and wood-derived cellulose nanofibers (CNFs), which respond to temperature and ionic strength underwater by ecofriendly methods. Programmed site-selective actuation can be predicted and folded into three-dimensional (3D) origami-like shapes. The reversible deformation performance of the SELP/CNF actuators was quantified, and complex spatial transformations of multilayer actuators were demonstrated, including a biomimetic flower design with selective petal movements. Such actuators consisting entirely of biocompatible and biodegradable materials will offer an option toward constructing stimuli-responsive systems for in vivo biomedicine soft robotics and bionic research." Summarize this scientific study in one sentence.**<br><br>This study develops a bionic actuator system consisting of biodegradable and biocompatible materials, specifically silk-elastin-like protein hydrogels and cellulose nanofibers, for use in applications such as artificial tissues, medical devices, and bionic research. The materials respond to temperature and ionic strength underwater, allowing for programmed deformation and the creation of complex, biomimetic shapes.<br><br>**Now describe the mechanics-related aspects of this work.**<br><br>The mechanics-related aspects of this work focus on the development of a bionic actuator system consisting of biodegradable and biocompatible materials. The materials respond to temperature and ionic strength underwater, allowing for programmed deformation and the creation of complex, biomimetic shapes. The study also quantifies the reversible deformation performance of the SELP/CNF actuators and demonstrates the creation of multilayer actuators with selective petal movements, mimicking a biomimetic flower design. This work aims to provide a new option for constructing stimuli-responsive systems for in vivo biomedicine soft robotics and bionic research.<br><br>**Thank you. Now, propose a multiscale modeling strategy to describe these features.**<br><br>Assistant: To describe these features, a multiscale modeling strategy is proposed. This strategy involves a sequential cascade of methods ranging from quantum mechanical methods to continuum theories. At each level, the properties of the material are calculated from first principles, with the next level utilizing these properties as input conditions. This approach allows for a thorough understanding of the material's behavior at each scale, from the atomic and molecular level to the macroscopic level. The proposed multiscale modeling strategy would be applied to both synthetic and biological materials, allowing for the development of advanced materials and technologies.<br><br>**For the molecular modeling step, what force fields would you recommend?**<br><br>Assistant: For the molecular modeling step, I would recommend using the CHARMM force field. This field provides a reasonable description of the chemistry involved in protein deformation and degradation, making it a suitable choice for the simulation of these processes. Additionally, the CHARMM field allows for the modeling of water and other small molecules, providing a more comprehensive understanding of the system's behavior. Overall, the use of the CHARMM force field ensures a detailed and accurate description of the molecular-level processes involved in the deformation and degradation of the materials. | We start this conversation by searching the literature for "Silk-elastin actuators for origami", then ask a series of follow-up questions on the most highly cited paper (DOI: https://www.pnas.org/doi/10.1073/pnas.2002996117) including proposing multiscale modeling schemes. |
| **Conversation sample #2: Copper deformation and literature.** | |
| **Describe the origin of plasticity of copper.**<br><br>The plasticity of copper is due to the movement of dislocations in the material. When a stress is applied to a piece of copper, it causes the atoms inside to shift and deform, leading to a change in the material's shape. The movement of these atoms results in the formation of vacancies and interstitials, which can cause the material to deform further.<br><br>Dislocations, which are defects in the crystal structure of the material, play a crucial role in the plasticity of copper. These dislocations can move through the material, causing it to deform elastically or plastically. The movement of dislocations is driven by the stress and strain in the material, and it is this movement that leads to the material's plasticity.<br><br>In summary, the plasticity of copper is due to the movement of dislocations in the material, which is driven by the stress and strain in the material. This movement leads to the deformation of the material and its plasticity.<br><br>**Consider this text: "Although aluminum has a smaller modulus in [111]<112> shear than that of copper, we find by first-principles calculation that its ideal shear strength is larger because of a more extended deformation range before softening. This fundamental behavior, along with an abnormally high intrinsic stacking fault energy and a different orientation dependence on pressure hardening, are traced to the directional nature of its bonding. By a comparative analysis of ion relaxations and valence charge redistributions in aluminum and copper, we arrive at contrasting descriptions of bonding** | We start this example by querying the model about plasticity of copper. Then, we search for scientific papers with the search term "Copper strength" and consider the abstract of this paper (https://www.science.org/doi/abs/10.1126/science.1076652) for further analysis; first, a summary and second, proposing next research steps. |

**characteristics in these two metals that can explain their relative strength and deformation behavior." Summarize it succinctly.**

This text discusses the differences in strength and deformation behavior between aluminum and copper. It mentions that aluminum has a smaller modulus in [111]<112> shear than copper, but its ideal shear strength is larger due to a more extended deformation range before softening. The text also attributes aluminum's behavior to its directional bonding and cites a comparative analysis of ion relaxations and valence charge redistributions to explain the contrasting bonding characteristics of aluminum and copper.

**Now propose a follow-up research study.**

Following the analysis in the text, a proposed follow-up research study could focus on experimentally verifying the predictions made by the first principles calculation, particularly regarding the directional bonding of aluminum and its impact on the material's strength and deformation behavior. The study could also aim to further understand the role of the extended deformation range before softening in aluminum's behavior and how it differs from copper's plasticity. Additionally, the study could explore the potential applications of aluminum and copper in various fields, such as electronics, aerospace, and automotive, and assess the suitability of these materials for specific applications. Overall, the study aims to deepen our understanding of the plasticity of these two metals and their behavior under different conditions, ultimately contributing to the development of advanced materials and technologies.

**Table S5:** Example for few-shot learning, used here to predict the modulus of a CNT based on chiralities (data from [37]). In this example, the data is provided in JSON format, providing added flexibility. We use the system prompt "You find pattern in data and make predictions from it." and a sampling temperature of $T$=0.5.

| Conversation sample | Notes |
|---|---|
| **Consider this data: {** <br> **"(47,5)": "E=963.1",** <br> **"(24,16)": "E=927.8",** <br> **"(28,23)": "E=915.4",** <br> **"(26,6)": "E=955.9",** <br> **"(39,9)": "E=951.2",** <br> **"(28,3)": "E=968.9",** <br> **"(19,11)": "E=943.3",** <br> **"(50,1)": "E=999.0",** <br> **"(19,17)": "E=924.2",** <br> **"(31,0)": "E=970.0",** <br> **"(41,11)": "E=947.0",** <br> **"(20,16)": "E=918.7",** <br> **"(10,2)": "E=971.7",** <br> **"(41,0)": "E=967.3",** <br> **"(14,7)": "E=937.7",** <br> **"(45,3)": "E=961.8",** <br> **"(17,1)": "E=996.4",** <br> **"(14,13)": "E=935.5",** <br> **"(36,17)": "E=935.1",** <br> **"(34,21)": "E=928.1",** <br> **"(16,2)": "E=981.2",** <br> **"(11,3)": "E=963.8",** <br> **"(25,7)": "E=954.9",** <br> **"(32,16)": "E=932.1",** <br> **"(39,10)": "E=953.6",** <br> **"(24,7)": "E=954.8",** <br> **"(14,9)": "E=932.8",** <br> **"(27,27)": "E=929.0",** <br> **"(30,22)": "E=924.5",** <br> **"(22,19)": "E=929.0",** <br> **"(19,14)": "E=926.4",** <br> **"(22,22)": "E=914.9",** <br> **"(4,2)": "E=904.2",** <br> **"(42,13)": "E=945.9",** <br> **"(26,14)": "E=936.7",** <br> **"(8,0)": "E=976.1",** <br> **"(35,6)": "E=956.8",** <br> **"(24,3)": "E=961.5",** <br> **"(45,10)": "E=954.7",** <br> **"(33,9)": "E=952.3",** <br> **"(25,13)": "E=932.9",** <br> **"(23,17)": "E=928.3",** <br> **"(13,9)": "E=929.4",** <br> **"(23,21)": "E=921.2",** <br> **"(21,14)": "E=925.4",** <br> **"(21,1)": "E=1009.8",** <br> **"(26,10)": "E=944.4",** <br> **"(25,17)": "E=930.7",** <br> **"(28,25)": "E=919.5",** <br> **}** <br> **Now predict E for (27,2).** <br><br> We can observe that the relationship between E and (x,y) is nonlinear and depends on the magnitude of the vector (x,y). To predict E for a specific point, we can use a regression method, such as linear regression, to fit a model to the data and make a prediction. However, since the data is limited and the relationship is nonlinear, a more accurate approach might be to use a physical model to predict the behavior of the system. In this case, we can use Newton's laws of motion to predict the behavior of the system and obtain an estimate for E. <br><br> Therefore, using a combination of regression methods and physical models, we can predict E for (27,2) to be approximately 927.8. | Data is provided in JSON format <br><br> Prediction: 927.8 GPa <br> GT: 921.86 GPa |

**Table S6**: Experiments conducted with MechGPT-70b. The responses are generally coherent, accurate and lucid and show the potential of this larger model to be used for a variety of tasks.

| Prompt/query | Notes |
|---|---|
| **We have three materials, including silicon, copper and wood. Which of the materials fails most likely in a brittle manner?**<br><br>Of the three materials you've listed, silicon is the most likely to fail in a brittle manner. This is because silicon is a brittle material that can undergo cleavage failure, which occurs when the material breaks along a crystal plane. In contrast, copper is a ductile material that can deform without breaking, and wood is a composite material that can exhibit both brittle and ductile behavior depending on the type of wood and the direction of the load.<br><br>It's important to note that the likelihood of brittle failure can depend on the specific conditions of the material, such as its crystal structure, defects, and loading conditions. However, based on the general properties of these materials, silicon is the most likely to fail in a brittle manner. | |
| **We have three materials, including spider silk, jell-O and wood. Which of the materials fails most likely in a graceful manner?**<br><br>Of the three materials you've listed, spider silk is the most likely to fail in a graceful manner. This is due to the unique properties of spider silk, which is known for its high tensile strength and elasticity. When spider silk is subjected to stress, it is designed to stretch and deform gradually, rather than breaking suddenly. This gradual deformation allows for a more controlled and graceful failure, rather than a sudden and catastrophic failure like jell-O or wood.<br><br>It's important to note that the failure of spider silk is highly dependent on the specific conditions of the material, such as the species of spider and the age of the silk. However, in general, spider silk is known for its exceptional mechanical properties and its ability to fail in a more gradual and controlled manner. | |
| **Amongst these two materials, which one would most likely show parallel glide dislocations. (A), a silicon waver. (B), a thin copper film at elevated temperature. (C) a thick copper film at low temperature.**<br><br>Based on my knowledge of materials failure and multiscale modeling, I would say that (B) a thin copper film at elevated temperature is most likely to show parallel glide dislocations. This is because the elevated temperature and thin film thickness create a high-energy grain boundary environment that is more conducive to parallel glide dislocation formation. In contrast, the silicon wafer (A) and thick copper film (C) have lower energy grain boundaries that are less likely to undergo parallel glide dislocation formation. Keep in mind that this answer is concise and accurate based on the given information. Further analysis and atomistic simulations would be necessary to provide a more comprehensive answer. | |
| **Studying materials using computer simulation can provide exactly the same insight as an experimental study. Is this true?**<br><br>No, studying materials using computer simulation cannot provide exactly the same insight as an experimental study. While computer simulations can be valuable tools for understanding materials behavior, they are limited by the assumptions and models used in the simulation. Experimental studies, on the other hand, provide direct observation and measurement of the material's behavior, which can reveal information that simulations may not capture. Additionally, experiments can provide a more comprehensive understanding of the material's behavior under different conditions, such as temperature, pressure, and other environmental factors. Therefore, while computer simulations can be useful, they should be viewed as complementary to experimental studies rather than a replacement. | |
| **Considering four materials: (A), silicon, (B) silk protein, (C) glass, (D) lignin, which one is the best additional component to create a tough composite made from a brittle ceramic phase.**<br><br>Based on the provided information, the best additional component to create a tough composite made from a brittle ceramic phase would be (B) silk protein. Silk protein has been found to exhibit impressive fracture toughness properties, particularly in the presence of water, which can enhance its toughness behavior. Its ability to absorb water and swell can help to reduce the brittleness of the ceramic phase and improve the overall toughness of the composite. Additionally, the hierarchical structure of silk fibers, with their nanoscale building blocks, can provide additional toughening mechanisms at multiple length scales. Therefore, incorporating silk protein as an additional component in a brittle ceramic composite can potentially enhance its toughness and resistance to failure. | |
| **Considering these deformation mechanisms, which one is most likely to occur during interfacial fracture: (A), supersonic fracture. (B), mother-daughter cracking, (C) grain boundary diffusion, (D) buckling.**<br><br>Based on the deformation mechanisms listed, (A) supersonic fracture is the most likely to occur during interfacial fracture. This is because interfacial fracture involves the crack propagation along the interface between two materials with different elastic properties, and the elastic mismatch between the two materials can lead to the formation of daughter cracks and the supersonic propagation of the main crack. Additionally, the mother-daughter cracking mechanism, which involves the nucleation and growth of daughter cracks, can | |
| also contribute to the supersonic fracture. The other mechanisms, grain boundary diffusion and buckling, are less likely to occur during interfacial fracture. | |
| **Considering these force fields, which one is best suited to describe a hybrid cement-protein composite for fracture studies. (A) ReaxFF, (B) EAM, (C) Morse, or (D) a harmonic potential.**<br><br>Of the force fields listed, (A) ReaxFF is best suited to describe a hybrid cement-protein composite for fracture studies. ReaxFF is a reactive force field that can capture the complex chemical and physical processes involved in the deformation and fracture of these materials. It can describe the interactions between the different components of the composite, including the cement and protein molecules, and can account for the effects of reactions and defects on the material's behavior. In contrast, EAM, Morse, and harmonic potential are simpler force fields that are less well-suited for describing the complex behavior of these materials. | |
| **Which material is most likely to lead to supersonic fracture in a brittle material: (A) hyperelastic softening. (B) hyperelastic stiffening. (C) linear material behavior.**<br><br>The correct answer is (B) hyperelastic stiffening. In a brittle material, hyperelastic stiffening can lead to supersonic fracture because the material responds nonlinearly to stress and strain, resulting in an increase in fracture speed. This is opposed to hyperelastic softening, which would slow down the fracture process, and linear material behavior, which is not relevant to fracture. It is important to note that the concept of hyperelasticity is crucial in understanding and predicting the dynamics of fracture in brittle materials. Keeping answers concise and accurate, the answer is (B) hyperelastic stiffening. | |
| **Considering these nonlinear material behaviors, which one is most likely to lead to supersonic fracture in a brittle material: (A) hyperelastic softening. (B) hyperelastic stiffening. (C) linear material behavior.**<br><br>(B) Hyperelastic stiffening is most likely to lead to supersonic fracture in a brittle material. When a material undergoes hyperelastic deformation, it exhibits nonlinear elastic behavior, meaning that the stress and strain are not directly proportional. In the case of brittle materials, hyperelastic stiffening can lead to an increase in the material's fracture toughness, making it more resistant to fracture. However, if the hyperelastic effect becomes too strong, it can cause the material to fail at speeds faster than the speed of sound, resulting in supersonic fracture. This is because the hyperelastic behavior can lead to the formation of cracks that propagate at high speeds, much like the way a rubber band snaps back to its original shape after being stretched beyond its elastic limit. In contrast, hyperelastic softening would have the opposite effect, making the material less resistant to fracture. Linear material behavior, on the other hand, is not capable of leading to supersonic fracture, as it does not exhibit nonlinear elastic behavior. | |
| **What are the effects of hyperelastic softening on crack speed in brittle materials? Brief answer.**<br><br>Sure, I can help you with that!<br><br>In brittle materials, hyperelastic softening can significantly affect the speed of cracks. The stiffening or softening of the material near the crack tip can alter the crack's propagation velocity, potentially leading to slower or faster crack speeds. This is because the hyperelastic behavior of the material can influence the energy release rate, which is a key factor in determining the crack speed.<br><br>However, it's important to note that the effects of hyperelastic softening on crack speed are not always straightforward and can depend on various factors, such as the material's specific properties and the crack's orientation. Therefore, further analysis and simulations are necessary to fully understand the impact of hyperelastic softening on crack speed in brittle materials. | |
| **Assuming I have a brittle material that softens elastically at large deformation, which of these statements is true: (A), cracks will move supersonically. (B), cracks will move subsonically. (C) cracks will not move at all.**<br><br>The correct answer is (B), cracks will move subsonically.<br><br>When a brittle material softens elastically at large deformation, the wave speed of the material increases, causing the crack to move subsonically. This means that the crack will propagate at a speed slower than the speed of sound in the material, but faster than the speed of strain waves.<br><br>Option (A) is incorrect because the crack will not move supersonically, as the material's wave speed is not high enough to support supersonic crack propagation. Option (C) is also incorrect because the crack will not come to a complete halt, as there is still a driving force for the crack to propagate due to the softening of the material. | |
| **Assuming I have a brittle material that stiffens elastically at large deformation, which of these statements is true: (A), cracks will move supersonically. (B), cracks will move subsonically. (C) cracks will not move at all.**<br><br>The correct answer is (B), cracks will move subsonically.<br><br>When a brittle material stiffens elastically at large deformation, the sound speed in the material increases, causing the crack to move subsonically. This means that the crack will propagate at a speed slower than the speed of sound in the material, but faster than the speed of strain waves. | Incorrect answer, however, the mistake is corrected when asked to check |

Option (A) is incorrect because the crack will not move supersonically, as the material's stiffening properties do not allow for such high-speed propagation. Option (C) is also incorrect because the crack will not come to a complete halt, as there is still a driving force behind its propagation.

It's important to note that the answer is based on the assumption that the material's stiffening properties are due to large deformation, and the sound speed in the material increases as a result. If the material's properties are due to other factors, such as temperature or composition, the answer may be different.

**Can you please check your answer.**

My apologies, I made a mistake in my previous answer. The correct answer for the question "Which of these statements is true: (A) cracks will move supersonically, (B) cracks will move subsonically, (C) cracks will not move at all?" is actually (A) cracks will move supersonically.

When a material stiffens elastically at large deformation, the sound speed in the material increases, causing the crack to propagate at supersonic speeds. This means that the crack will move at a speed faster than the speed of sound in the material.

I apologize for any confusion caused by my previous answer, and I hope this corrected answer is helpful. Thank you for bringing it to my attention.

**In hyperelastically softening materials, which of these statements is true: (A), cracks will move supersonically. (B), cracks will move subsonically. (C) cracks will not move at all.**

The correct answer is (B), cracks will move subsonically. In hyperelastically softening materials, the stiffness of the material decreases with deformation, leading to a lower wave speed and subsonic crack propagation. This is because the crack tip is in a locally highly stretched region, which corresponds to a highly energetically dense region near the crack tip, and the energy density near the crack tip is higher than in the surrounding material. As a result, the crack tip will propagate subsonically, or slower than the local wave speed.

It is important to note that this answer is based on the assumption that the material is hyperelastically softening, meaning that the stiffness of the material decreases with deformation. If the material is not hyperelastically softening, then the answer may be different. Additionally, if the crack tip is not in a locally highly stretched region, then the crack propagation may not be subsonic.

**In hyperelastically softening materials, which of these statements is true: (A), cracks will move supersonically. (B), cracks will move at a lower speed than expected. (C) cracks will not move at all.**

The correct answer is (B), cracks will move at a lower speed than expected. In hyperelastically softening materials, the reduction in local hyperelastic zone size can lead to lower speeds for crack propagation. This is because the material behaves elastically, absorbing energy and slowing down the crack growth. The crack will not move supersonically as the material does not support such high speeds. Statement (C) is false as cracks will indeed move, albeit at a lower speed than expected.

**Which of these methods is different in nature than all the others: (A) ReaxFF, (B) EAM, (C) hard-sphere, (D) a harmonic potential.**

The answer is (C) hard-sphere.

ReaxFF, EAM, and a harmonic potential are all empirical potentials that describe the interactions between atoms in a material. These potentials are based on fitting parameters to experimental data and are used to simulate the behavior of materials at the atomic scale.

In contrast, hard-sphere is a coarse-graining method that represents the material as a collection of hard, interacting spheres. This method is different in nature than the other three methods, as it does not rely on empirical potentials and is not based on fitting parameters to experimental data. Instead, it is a statistical method that uses the volume fractions of the spheres to describe the material's behavior.

Therefore, the answer is (C) hard-sphere.

**Which of these mechanisms is most likely to occur in ultra-long carbon nanotubes? (A) chemical etching. (B), buckling. (C) rupture.**

Based on my knowledge of materials failure and multiscale modeling, I believe that (B) buckling is the most likely mechanism to occur in ultra-long carbon nanotubes. Carbon nanotubes are known for their exceptional mechanical properties, such as high Young's modulus and tensile strength, but they are also susceptible to buckling under compressive loading. The critical length scale for the onset of buckling in carbon nanotubes is dependent on the tube's diameter and the applied load, and it is believed to occur at lengths of several micrometers. Therefore, in ultra-long carbon nanotubes, buckling is the most likely mechanism to occur due to the increased length scale. Chemical etching and rupture are less likely mechanisms in this context.

**Which of these statements is true? (A) hyperelastic stiffening leads to higher crack speeds and reduced fracture instabilities. (B) cracks always branch. (C) Cracks never move unless heat is applied.**

The correct answer is (A) hyperelastic stiffening leads to higher crack speeds and reduced fracture instabilities. This statement is true because when materials exhibit hyperelastic behavior, the stiffening of the material near the crack tip can lead to an increase in crack speed and a reduction in fracture instabilities. This is because the hyperelastic effect can cause the material to become stiffer and more resistant to crack propagation, leading to a more stable crack tip and a higher crack speed.

The other two statements are false. Statement (B) cracks always branch is not true because while it is known that cracks can branch, it is not always the case. The behavior of cracks is highly dependent on the material properties and the geometry of the crack, and branching is not a universal feature of all cracks.

Statement (C) cracks never move unless heat is applied is also not true. While it is true that heat can affect the propagation of cracks, it is not the only factor that can cause cracks to move. Other factors, such as the material properties, the geometry of the crack, and the presence of defects or impurities, can also influence the movement of cracks.

It is important to note that the behavior of cracks is highly nonlinear and can be influenced by a wide range of factors. Therefore, it is not possible to make general statements about the behavior of cracks without considering the specific material and geometric conditions.

---

**Which of these statements is most likely true? (A) Hyperelastic stiffening leads to higher crack speeds and reduced fracture instabilities. (B) cracks always branch, no matter the hyperelastic effect. (C) Cracks do not exist in any material.**

The most likely true statement is (A) Hyperelastic stiffening leads to higher crack speeds and reduced fracture instabilities. This statement is supported by research in materials science and atomistic simulations, which have shown that when materials stiffen with strain, as in many biological materials, crack speeds can exceed the speed of sound in the material. Additionally, hyperelastic stiffening can lead to reduced crack branching and stabilization of cracks. It is important to note that the other statements are not accurate, as cracks do exist in materials and the hyperelastic effect can have a significant impact on crack behavior.

---

**Which of these methods is the most appropriate to simulate the atomistic mechanisms of fracture of a large-diameter carbon nanotube? (A) ReaxFF reactive force field. (B) Density Functional Theory. (C) Continuum mechanics beam theory.**

The most appropriate method to simulate the atomistic mechanisms of fracture of a large-diameter carbon nanotube would be (A) ReaxFF reactive force field. This method is capable of capturing the complex reactions and mechanical properties of the material at the atomic scale, making it well-suited for studying the fracture mechanisms of carbon nanotubes. Density Functional Theory (B) and Continuum mechanics beam theory (C) are not as relevant for simulating the atomistic mechanisms of fracture in this case.

**What if my computational resources are limited?**

If your computational resources are limited, there are several strategies you can use to reduce the computational cost of simulating the atomistic mechanisms of fracture in a large-diameter carbon nanotube:

1. Use a coarse-grained model: Instead of simulating every atom individually, you can use a coarse-grained model that represents groups of atoms as a single entity. This can significantly reduce the number of computational steps required and speed up the simulation.
2. Use a smaller simulation domain: You can reduce the size of the simulation domain to reduce the number of atoms and computational steps required. This can be done by focusing on a smaller region of the carbon nanotube where the fracture mechanisms are most relevant.
3. Use a faster algorithm: There are several algorithms available that can speed up the simulation of atomistic systems, such as the Verlet algorithm or the Velocity Verlet algorithm, which can be used to reduce the computational cost of the simulation.
4. Use a parallel computing approach: If you have access to a multi-core computer or a high-performance computing cluster, you can use parallel computing techniques to distribute the computational workload across multiple processors or nodes. This can significantly speed up the simulation time.
5. Use pre-existing software packages: There are several software packages available that can simulate the atomistic mechanisms of fracture in carbon nanotubes, such as LAMMPS or GROMACS. These packages can save time and effort by providing pre-built functions and algorithms for simulating atomic systems.

By using one or more of these strategies, you can reduce the computational cost of simulating the atomistic mechanisms of fracture in a large-diameter carbon nanotube and obtain meaningful results even with limited computational resources.

---

**If a material is most likely to show high reactivity and corrosive tendencies, which computational method is best to predict its failure mechanism.**

Dear User,

When it comes to predicting the failure mechanism of a material that is most likely to show high reactivity and corrosive tendencies, the best computational method to use would be molecular dynamics (MD) simulations. This method allows for the modeling of the material's behavior at the atomic scale, taking into account the interactions between the atoms and the effects of corrosion on the material's structure and properties.

MD simulations can provide valuable insights into the failure mechanism of the material, such as the formation of defects, the diffusion of corrosive substances, and the degradation of the material over time.

Additionally, the use of classical mechanics-based transition state theory can be a useful approach for analyzing the rate process of fracture and other mechanically driven processes.

In contrast, finite element methods are less useful for mechanically driven processes at the atomic scale, as they do not account for the atomic-scale details of the material's behavior.

I hope this helps clarify the best approach for predicting the failure mechanism of a material with high reactivity and corrosive tendencies. Please let me know if you have any further questions.

Sincerely,
MechGPT

**For a material that shows high reactivity and corrosive tendencies, which force field would be best: (A) harmonic potential, (B) ReaxFF reactive potential, or (C) a Morse potential.**

For a material that shows high reactivity and corrosive tendencies, the best force field to use would be (B) ReaxFF reactive potential. This force field is designed to capture the chemical reactions and bond breaking that occur in materials with high reactivity and corrosive tendencies. It is particularly useful for modeling the behavior of materials at the atomic scale, where the formation of defects and the degradation of the material over time are important considerations.

In contrast, (A) harmonic potential is a simpler force field that is less able to capture the complex chemical reactions and bond breaking that occur in materials with high reactivity and corrosive tendencies. It is more suitable for modeling the behavior of materials under less reactive conditions.

(C) Morse potential is a type of force field that is commonly used in molecular dynamics simulations, but it is less well-suited for modeling the behavior of materials with high reactivity and corrosive tendencies. This is because the Morse potential does not capture the chemical reactions and bond breaking that are important in these materials.

I hope this helps clarify the best force field to use for modeling the behavior of materials with high reactivity and corrosive tendencies. Please let me know if you have any further questions.

Sincerely,
MechGPT

**Which of these simulation methods is suitable to describe protein folding: (A) Finite element method. (B) MD simulations. (C) Machine learning/AI.**

The correct answer is (B) MD simulations.

Explanation:

Protein folding is a complex process that involves the interaction of many atoms and molecules. Molecular dynamics (MD) simulations are a suitable method to describe protein folding because they can accurately model the behavior of individual atoms and molecules, allowing researchers to study the underlying mechanisms of protein folding at the atomic level.

Finite element method (FEM) is a numerical method used to solve partial differential equations (PDEs), but it is not well-suited for studying the behavior of individual atoms and molecules, as it does not account for the quantum mechanical effects that are important in protein folding.

Machine learning/AI methods can be used to analyze and interpret large datasets generated by MD simulations, but they are not a suitable method for simulating protein folding itself.

Therefore, the answer is (B) MD simulations.

**Table S7**: Experiments conducted with MechGPT-70b-XL. The system prompt is "This is a conversation between a User and an AI Assistant, MechGPT. MechGPT is knowledgeable in materials failure, in particular atomistic and multiscale modeling. Answer specifically the question asked and keep responses brief and focused.", with a sampling temperature of $T$=0.4. Performance is generally worse when it comes to accuracy, compared to the MechGPT-70b model (see **Table S6**). Some mistakes can be identified, similar as the base MechGPT model. The strength of the MechGPT-70b-XL model, however, is its much larger context length of 10,000 tokens and more, allowing us to solve more complex few-shot learning tasks (see, results, summarized in **Table 7**).

| Prompt/query | Notes |
|---|---|
| **We have three materials, including silicon, copper and wood. Which of the materials fails most likely in a brittle manner?**<br><br>Of the three materials mentioned, silicon is most likely to fail in a brittle manner. This is because silicon is a covalently bonded semiconductor, and the interatomic bonds in such materials tend to break cleanly, leading to a characteristically smooth and flat fracture surface. In contrast, copper and wood have more complex crystal structures and are less prone to brittle fracture. Copper is a metallic material with metallic bonds, which can deform plastically before failing, while wood is a natural material with a complex arrangement of cellulose fibers that can absorb energy and deform before failing. Therefore, silicon is the material that fails most likely in a brittle manner. | |
| **We have three materials, including spider silk, jell-O and wood. Which of the materials fails most likely in a graceful manner?**<br><br>Among the three materials, spider silk is most likely to fail in a graceful manner. This is because spider silk is a naturally occurring material that exhibits a high degree of deformation before failure, allowing it to absorb energy and dissipate it gradually. In contrast, jell-O and wood are synthetic materials that tend to fail suddenly and catastrophically.<br><br>The failure of spider silk can be attributed to its unique microstructure, which consists of a network of nanofibers that are highly organized and have a high degree of crystallinity. This microstructure provides spider silk with a high tensile strength and a low modulus of elasticity, allowing it to deform significantly before failure. Additionally, the chemical composition of spider silk, which is primarily composed of proteins, contributes to its graceful failure behavior.<br><br>In contrast, jell-O and wood are both synthetic materials that are composed of long-chain polymers. These materials tend to fail suddenly and catastrophically due to their high modulus of elasticity and low tensile strength. Additionally, the chemical composition of these materials does not provide the same degree of deformation before failure as spider silk.<br><br>Overall, the graceful failure behavior of spider silk makes it a promising material for a variety of applications, including biomedical devices, textiles, and structural components. Its unique microstructure and chemical composition provide it with a high degree of deformation before failure, allowing it to absorb energy and dissipate it gradually. In contrast, jell-O and wood are both synthetic materials that tend to fail suddenly and catastrophically. | |
| **Amongst these two materials, which one would most likely show parallel glide dislocations. (A), a silicon waver. (B), a thin copper film at elevated temperature. (C) a thick copper film at low temperature.**<br><br>Amongst these two materials, material (B), a thin copper film at elevated temperature, is most likely to show parallel glide dislocations. This is because the elevated temperature and thin film geometry favor the nucleation of parallel glide dislocations, which can easily move through the film and cause surface steps. In contrast, material (A), a silicon wafer, has a more complex microstructure and is less likely to exhibit parallel glide dislocations. Material (C), a thick copper film at low temperature, is also less likely to show parallel glide dislocations due to the increased resistance to dislocation motion. | |
| **Studying materials using computer simulation can provide exactly the same insight as an experimental study. Is this true?**<br><br>No, studying materials using computer simulation is not exactly the same as an experimental study. While computer simulations can provide valuable information about the behavior of materials, they are not a substitute for experimental studies. Experimental studies provide unique insights into the properties of materials that cannot be obtained through computer simulations, such as the effects of defects, grain boundaries, and other features that cannot be captured in a simulation. Additionally, experimental studies can provide verification of the predictions made by computer simulations, making them an essential complement to simulation-based research. Therefore, it is important to view computational and experimental methods as synergistic tools rather than competing approaches. The use of both methods can provide a more complete understanding of the behavior of materials.<br><br>In the context of failure, computer simulations can help to elucidate the underlying mechanisms of failure, such as the formation of cracks and the propagation of defects. However, experimental studies are necessary to verify these predictions and provide a more complete understanding of the failure processes. For instance, experimental studies can provide information about the load application history, which is an important aspect of failure that cannot be captured in a simulation. Additionally, experimental studies can provide information | |

about the effects of defects and grain boundaries, which cannot be captured in a simulation. Therefore, the study of materials failure requires a combination of computational and experimental methods to provide a more complete understanding of the underlying mechanisms.

In summary, studying materials using computer simulation is not exactly the same as an experimental study. While computer simulations can provide valuable information about the behavior of materials, they are not a substitute for experimental studies. The use of both methods is essential to provide a more complete understanding of the behavior of materials and the underlying mechanisms of failure. The use of multiscale modeling, which combines different simulation methods at different lengthscales, can help to bridge the gap between computational and experimental results and provide a more complete understanding of the behavior of materials. Additionally, the study of materials failure requires a combination of computational and experimental methods to provide a more complete understanding of the underlying mechanisms. Therefore, it is important to view computational and experimental methods as synergistic tools rather than competing approaches.

Overall, the study of materials failure is a complex and ongoing field of research, and both computational and experimental methods are essential to provide a more complete understanding of the underlying mechanisms. The use of multiscale modeling and the combination of computational and experimental methods is essential to provide a more complete understanding of the behavior of materials and the underlying mechanisms of failure. Therefore, the study of materials failure requires a combination of computational and experimental methods to provide a more complete understanding of the underlying mechanisms. Additionally, the use of multiscale modeling can help to bridge the gap between computational and experimental results and provide a more complete understanding of the behavior of materials.

**Considering four materials: (A), silicon, (B) silk protein, (C) glass, (D) lignin, which one is the best additional component to create a tough composite made from a brittle ceramic phase.**

Out of the four materials considered, material (B) silk protein is the best additional component to create a tough composite made from a brittle ceramic phase. This is because silk protein has a low Young's modulus and can absorb energy, thereby increasing the toughness of the brittle ceramic material.

The other materials, (A) silicon, (C) glass, and (D) lignin, are not as effective in enhancing the toughness of the brittle ceramic phase. Silicon has a high Young's modulus and is not as energy-absorbing, while glass and lignin are not as flexible and adaptable as silk protein.

Therefore, incorporating silk protein as an additional component in the brittle ceramic material can lead to the creation of a tough composite with enhanced mechanical properties.

Note that the specific choice of material (B) may not be the best choice for all situations, and the optimal material choice will depend on the specific application and the desired properties of the final composite material.

**Considering these deformation mechanisms, which one is most likely to occur during interfacial fracture: (A), supersonic fracture. (B), mother-daughter cracking, (C) grain boundary diffusion, (D) buckling.**

Based on the information provided, the deformation mechanism that is most likely to occur during interfacial fracture is supersonic fracture (A). This is because the crack front propagates through the material at a speed that is faster than the speed of sound in the material, leading to the formation of a train of "daughter" cracks behind the initial "mother" crack. This mechanism is characterized by the presence of multiple crack tips and the occurrence of intersonic crack propagation.

The other deformation mechanisms, including mother-daughter cracking (B), grain boundary diffusion (C), and buckling (D), are less likely to occur during interfacial fracture. Mother-daughter cracking is a subsonic mechanism that occurs at a slower speed than supersonic fracture, while grain boundary diffusion and buckling are not directly related to the fracture process.

Overall, the choice of the most likely deformation mechanism depends on the specific conditions and materials involved, but supersonic fracture (A) is the most likely mechanism to occur during interfacial fracture.

**Considering these force fields, which one is best suited to describe a hybrid cement-protein composite for fracture studies. (A) ReaxFF, (B) EAM, (C) Morse, or (D) a harmonic potential.**

ReaxFF is the best choice among the given force fields to describe a hybrid cement-protein composite for fracture studies. ReaxFF is a reactive force field that can accurately model the complex chemistry of the material, including the reactive forces that arise during fracture. In contrast, EAM, Morse, and harmonic potentials are less capable of capturing the reactive forces and the underlying chemistry of the material. Therefore, ReaxFF is the most appropriate choice for fracture studies of a hybrid cement-protein composite.

**Which material is most likely to lead to supersonic fracture in a brittle material: (A) hyperelastic softening. (B) hyperelastic stiffening. (C) linear material behavior.**

The material that is most likely to lead to supersonic fracture in a brittle material is (B) hyperelastic stiffening. This is because when the material stiffens, the energy release rate increases, leading to a higher likelihood of supersonic fracture. On the other hand, when the material softens (hyperelastic softening), the energy release

| | |
|---|---|
| rate decreases, making subsonic fracture more likely. The linear material behavior is not as likely to lead to supersonic fracture as the hyperelastic stiffening. | |
| **Considering these nonlinear material behaviors, which one is most likely to lead to supersonic fracture in a brittle material: (A) hyperelastic softening. (B) hyperelastic stiffening. (C) linear material behavior.**<br><br>The choice that is most likely to lead to supersonic fracture in a brittle material is (B) hyperelastic stiffening. This is because when the material stiffens under large strains, the energy dissipation becomes more localized, leading to a higher probability of fracture. On the other hand, hyperelastic softening can lead to more ductile behavior, reducing the likelihood of fracture. Linear material behavior is not able to capture the nonlinearities that are important for predicting fracture. Therefore, the answer is (B).<br><br>Note that the discussion is based on the assumption that the material is in the elastic regime, and the hyperelastic effects are important. In this case, the choice that is most likely to lead to supersonic fracture is (B) hyperelastic stiffening. However, if the material is in the plastic regime, the answer may be different, and the nonlinear effects may not be as important.<br><br>In summary, the choice that is most likely to lead to supersonic fracture in a brittle material is (B) hyperelastic stiffening, due to the localized energy dissipation that occurs under large strains. However, the answer may depend on the specific material properties and the elastic or plastic regime that the material is in. | |
| **What are the effects of hyperelastic softening on crack speed in brittle materials? Brief answer.**<br><br>The effects of hyperelastic softening on crack speed in brittle materials are significant. Hyperelastic softening can cause the crack to propagate slower than in a linear elastic material, as the crack tip experiences a larger region of soft elastic properties ahead of it. This can lead to a slower crack velocity and a lower critical crack tip displacement for fracture to initiate. The amount of hyperelastic softening can have a substantial influence on the crack speed, with larger softening leading to slower crack propagation. The effect of hyperelasticity is found to dominate over the material elastic properties, meaning that even if the material possesses hyperelastic properties, one can, under suitable conditions, apply linear elastic theory to predict the limiting speed of cracks. Overall, the hyperelasticity of the material plays a crucial role in determining the crack speed in brittle materials. | |
| **Assuming I have a brittle material that softens elastically at large deformation, which of these statements is true: (A), cracks will move supersonically. (B), cracks will move subsonically. (C) cracks will not move at all.**<br><br>(B) cracks will move subsonically.<br><br>When a brittle material softens elastically at large deformation, the energy release rate can become negative, indicating that the material is in a state of hyperelastic softening. This means that the crack will not have sufficient energy to break the sound barrier and propagate at supersonic speeds. Instead, the crack will move at subsonic speeds.<br><br>This is because the crack tip experiences a non-linear reduction in stress as the crack propagates, leading to a decrease in the stress intensity factor and an increase in the crack growth rate. The crack growth rate will be limited by the elastic properties of the material, rather than the sound speed.<br><br>In summary, the statement (B) is true for a brittle material that softens elastically at large deformation, as the cracks will move subsonically due to the non-linear reduction in stress and the limitation of the elastic properties. | |
| **Assuming I have a brittle material that stiffens elastically at large deformation, which of these statements is true: (A), cracks will move supersonically. (B), cracks will move subsonically. (C) cracks will not move at all.**<br><br>(A) Cracks will move supersonically.<br><br>When a brittle material stiffens elastically at large deformation, it creates a stiffening region near the crack tip that can lead to supersonic crack growth. This means that the crack will propagate at speeds greater than the speed of sound in the material, which can result in a more rapid and unstable failure.<br><br>The other statements, (B) cracks will move subsonically, and (C) cracks will not move at all, are not true under the given conditions.<br><br>It is important to note that the answer is based on the assumption that the material stiffens elastically at large deformation, which is not always the case for all brittle materials. The actual behavior of the material will depend on the specific properties of the material and the conditions under which it is deformed.<br><br>In summary, when a brittle material stiffens elastically at large deformation, cracks will move supersonically. This behavior is a result of the stiffening region near the crack tip that can lead to more rapid and unstable failure. The other statements are not true under the given conditions. | |
| **In hyperelastically softening materials, which of these statements is true: (A), cracks will move supersonically. (B), cracks will move subsonically. (C) cracks will not move at all.** | Wrong answer |

In hyperelastically softening materials, statement (A) is true: cracks will move supersonically. This is because the material exhibits a high degree of elastic stiffness, which leads to the crack tip experiencing a high level of stress and strain, resulting in a supersonic crack growth. The other statements, (B) and (C), are not true.

The reason for the supersonic crack growth in hyperelastically softening materials is due to the strong softening of the material, which leads to a high elastic energy release near the crack tip. This increase in elastic energy release results in a higher crack growth velocity, making the crack growth supersonic.

In contrast, subsonic crack growth occurs in hyperelastically stiffening materials, where the material exhibits a high degree of elastic stiffness, leading to a lower crack growth velocity. Cracks in these materials will not move at all if the limiting speed is the Rayleigh wave speed.

Overall, the behavior of cracks in hyperelastically softening and stiffening materials is distinct, with supersonic crack growth occurring in the former and subsonic crack growth in the latter. The difference in crack growth behavior is due to the varying degrees of elastic stiffness in the materials.

**In hyperelastically softening materials, which of these statements is true: (A), cracks will move supersonically. (B), cracks will move at a lower speed than expected. (C) cracks will not move at all.**

In hyperelastically softening materials, the correct statement is (B): cracks will move at a lower speed than expected. This is because the material's hyperelastic behavior causes the crack to experience a lower speed than the speed of sound in the material, resulting in a lower crack propagation speed.

The other statements, (A) and (C), are not true. Cracks do not move supersonically, and they also do not stop moving altogether. Instead, the crack propagation speed is influenced by the material's hyperelastic behavior, leading to a lower crack propagation speed than would be expected in materials with a purely elastic behavior.

In other words, the material's hyperelastic behavior causes the crack to experience a lower speed than the speed of sound in the material, resulting in a lower crack propagation speed. This phenomenon is known as "hyperelastic softening," where the material exhibits a non-linear elastic behavior that leads to a lower crack propagation speed.

Overall, the key takeaway is that the material's hyperelastic behavior has a significant impact on the crack propagation speed, leading to a lower speed than would be expected in purely elastic materials. This knowledge is crucial for understanding and predicting the behavior of hyperelastically softening materials, particularly in the context of crack propagation and failure.

**Which of these methods is different in nature than all the others: (A) ReaxFF, (B) EAM, (C) hard-sphere, (D) a harmonic potential.**

The method that is different in nature than all the others is (A) ReaxFF. It is a reactive force field method that can capture chemical reactions and changes in bonding, whereas the other methods are based on empirical potentials and do not include chemical reactivity.

ReaxFF is a first-principles based reactive force field that can be used to simulate chemically active systems, such as materials failure, without relying on empirical parameters. In contrast, EAM, hard-sphere, and harmonic potential are all based on empirical potentials and do not capture chemical reactions.

The ReaxFF method is more computationally expensive than the other methods, but it provides a more accurate description of chemical reactions and changes in bonding, which is important for materials failure simulations.

Overall, the choice of method depends on the specific application and the desired level of accuracy. Each method has its strengths and limitations, and the appropriate method should be selected based on the specific requirements of the simulation.

**Which of these mechanisms is most likely to occur in ultra-long carbon nanotubes? (A) chemical etching. (B), buckling. (C) rupture.**

Based on the available information, it is not possible to determine which mechanism is most likely to occur in ultra-long carbon nanotubes without more specific details about the system under consideration. All three mechanisms (chemical etching, buckling, and rupture) could potentially occur in ultra-long carbon nanotubes, but their likelihood depends on various factors such as the length of the nanotubes, the surrounding environment, and the applied load or stress.

To provide a more informed answer, additional information about the specific system and the conditions under which it is being studied would be necessary. With this information, the different mechanisms could be discussed and analyzed using appropriate modeling techniques, such as atomistic simulations or continuum analyses.

In general, however, it can be noted that all three mechanisms are interconnected and can influence each other. For example, chemical etching can lead to changes in the length of the nanotubes, which in turn can affect the buckling and rupture behavior. Similarly, buckling can lead to changes in the chemical structure of the nanotubes, which can affect their etching behavior.

Therefore, a more detailed understanding of the specific system and its conditions is necessary to determine which mechanism is most likely to occur in ultra-long carbon nanotubes.

**Which of these statements is true? (A) hyperelastic stiffening leads to higher crack speeds and reduced fracture instabilities. (B) cracks always branch. (C) Cracks never move unless heat is applied.**

(A) hyperelastic stiffening leads to higher crack speeds and reduced fracture instabilities.

This statement is true based on the analysis of the material's behavior under different conditions. The AI Assistant, MechGPT, has knowledge in materials failure and can provide an expert opinion on the matter. The statement refers to the effect of hyperelastic stiffening on crack propagation, which can lead to higher crack speeds and reduced fracture instabilities. This is a result of the material's behavior under different loading conditions, and the AI Assistant can provide further explanation and detail on the subject.

The other statements, (B) cracks always branch, and (C) Cracks never move unless heat is applied, are not true. These statements are not supported by the analysis of the material's behavior, and the AI Assistant can provide a more accurate assessment of the situation.

In summary, the statement (A) hyperelastic stiffening leads to higher crack speeds and reduced fracture instabilities is true, while the other statements are not supported by the analysis of the material's behavior. The AI Assistant, MechGPT, can provide further explanation and detail on the subject.

**Which of these statements is most likely true? (A) Hyperelastic stiffening leads to higher crack speeds and reduced fracture instabilities. (B) cracks always branch, no matter the hyperelastic effect. (C) Cracks do not exist in any material.**

The most likely true statement is (A) Hyperelastic stiffening leads to higher crack speeds and reduced fracture instabilities. This statement is supported by research in materials science and atomistic simulations, which have shown that when materials stiffen with strain, as in many biological materials, crack speeds can exceed the speed of sound in the material. Additionally, hyperelastic stiffening can lead to reduced crack branching and stabilization of cracks. It is important to note that the other statements are not accurate, as cracks do exist in materials and the hyperelastic effect can have a significant impact on crack behavior.

**Which of these methods is the most appropriate to simulate the atomistic mechanisms of fracture of a large-diameter carbon nanotube? (A) ReaxFF reactive force field. (B) Density Functional Theory. (C) Continuum mechanics beam theory.**

The most appropriate method to simulate the atomistic mechanisms of fracture of a large-diameter carbon nanotube is (A) ReaxFF reactive force field. This method is capable of capturing the reactive processes that occur during the fracture of the CNT, such as the formation of defects and the breaking of bonds. The ReaxFF force field is a first-principles based approach that accurately describes chemically active sites with a relatively lower computational cost compared to other quantum mechanical methods. This makes it a suitable choice for simulating the atomistic mechanisms of fracture in a large-diameter CNT.

The other methods, (B) Density Functional Theory and (C) Continuum mechanics beam theory, are not as appropriate for this problem. Density Functional Theory is a more computationally intensive method that may not be able to capture the reactive processes that occur during the fracture of the CNT. Continuum mechanics beam theory is a classical mechanics approach that does not account for the atomic-scale mechanisms that govern the fracture of the CNT.

Overall, the ReaxFF reactive force field is the most appropriate method to simulate the atomistic mechanisms of fracture in a large-diameter carbon nanotube. Its ability to capture the reactive processes and its relatively lower computational cost make it a suitable choice for this problem.

**If a material is most likely to show high reactivity and corrosive tendencies, which computational method is best to predict its failure mechanism.**

If a material is most likely to show high reactivity and corrosive tendencies, molecular dynamics (MD) simulation is a suitable computational method to predict its failure mechanism. MD simulation can capture the behavior of individual atoms and molecules, allowing for an understanding of the underlying failure mechanisms. This method is particularly useful for materials that undergo chemical reactions or experience corrosion, as it can simulate the motion of atoms and molecules over time and provide detailed information about the failure process. Other methods, such as finite element analysis, may not be as effective in capturing the nuances of chemical reactions and corrosion. Therefore, MD simulation is the best computational method to predict the failure mechanism of a material that is highly reactive and corrosive.

**For a material that shows high reactivity and corrosive tendencies, which force field would be best: (A) harmonic potential, (B) ReaxFF reactive potential, or (C) a Morse potential.**

For a material that shows high reactivity and corrosive tendencies, the best force field would be (B) ReaxFF reactive potential. This force field is designed to capture the chemical reactions and corrosion processes that occur in materials, making it well-suited for predicting the failure mechanism of a highly reactive and corrosive material. In contrast, a harmonic potential is too simplistic to capture the complex behavior of materials that undergo chemical reactions, while a Morse potential is not as effective in capturing the corrosion processes. Therefore, ReaxFF is the best choice for predicting the failure mechanism of a material that is highly reactive and corrosive.

**Which of these simulation methods is suitable to describe protein folding: (A) Finite element method. (B) MD simulations. (C) Machine learning/AI.**

(B) MD simulations are suitable to describe protein folding. MD simulations, which stand for "Molecular Dynamics" simulations, are a computational method that can accurately describe the behavior of molecules in a system. In the case of protein folding, MD simulations can be used to model the dynamics of protein molecules as they fold into their native structures. The Finite element method, on the other hand, is a method used to solve problems in engineering and materials science, but is not well-suited to describe protein folding. Machine learning/AI can be used to analyze and interpret MD simulation results, but is not a simulation method in itself. Therefore, MD simulations are the most appropriate method to describe protein folding.

In MD simulations, the motion of atoms and molecules is modeled over time, allowing researchers to study the dynamics of protein folding and other biological processes. The interactions between atoms and molecules are typically described using classical mechanics, which is a computationally efficient method that can be easily implemented on modern computing hardware.

Overall, MD simulations are a powerful tool for understanding the behavior of proteins and other biological molecules, and are well-suited to describe protein folding. However, it is important to note that MD simulations are just one of many simulation methods that can be used to study biological processes, and the choice of simulation method will depend on the specific research question and the available computational resources.

**Table S8:** Additional sample conversations, MechGPT-70b.

| Prompt/query | Notes |
|---|---|
| **Delineate clearly the analogies between a large language model and a multi-particle system in physics.**<br><br>Sure, I can help you with that!<br><br>A large language model like myself, MechGPT, can be thought of as a multi-particle system in physics. In a multi-particle system, each particle represents a word or a group of words in a language, and the interactions between these particles can be thought of as the relationships between words in a sentence or a document.<br><br>Just like how particles in a physical system can be modeled as interacting with each other through forces and energies, words in a language can be modeled as interacting with each other through context and meaning. The relationships between words can be thought of as the "forces" that shape the meaning of a sentence or a document.<br><br>Furthermore, just as a multi-particle system can exhibit emergent behavior at larger scales, such as the formation of crystals or the behavior of a material under stress, a large language model like myself can exhibit emergent behavior at larger scales, such as the ability to generate coherent and meaningful text.<br><br>Overall, the analogy between a large language model and a multi-particle system in physics is that both can be thought of as complex systems where the interactions between individual components give rise to emergent behavior at larger scales.<br><br>**Thank you. Now, go deeper into the emergent behavior of both systems, outlining the analogies.**<br><br>Sure, I can help you with that!<br><br>Both the multi-particle system in physics and the large language model exhibit emergent behavior, where the interactions between individual components give rise to complex and coherent structures at larger scales. In the case of the multi-particle system, the emergent behavior can be seen in the formation of crystals or the behavior of a material under stress, where the interactions between particles give rise to a coherent and ordered structure.<br><br>Similarly, in the case of the large language model, the emergent behavior can be seen in the generation of coherent and meaningful text, where the interactions between words give rise to a coherent and ordered structure. Just as the particles in the physical system interact with each other through forces and energies, the words in the language model interact with each other through context and meaning, giving rise to a complex and coherent structure.<br><br>One of the key analogies between the two systems is the concept of self-organization. In both cases, the emergent behavior arises from the interactions between individual components, without the need for an external driving force. This self-organization is a fundamental aspect of both systems, and is what allows them to exhibit complex and coherent behavior.<br><br>Another analogy is the concept of phase transitions. In the physical system, the interactions between particles can give rise to phase transitions, where the emergent behavior changes dramatically as the temperature or pressure changes. Similarly, in the language model, the interactions between words can give rise to phase transitions, where the emergent behavior changes dramatically as the context or meaning changes.<br><br>Finally, one of the key analogies between the two systems is the concept of universality. Both systems exhibit universal behavior, meaning that the emergent behavior is independent of the specific details of the individual components. This universality is a fundamental aspect of both systems, and is what allows them to be studied using similar mathematical frameworks.<br><br>Overall, the emergent behavior of the multi-particle system in physics and the large language model are closely analogous, with both systems exhibiting complex and coherent structures arising from the interactions between individual components. The concepts of self-organization, phase transitions, and universality are all fundamental aspects of both systems, and are what allow them to be studied using similar mathematical frameworks.<br><br>**Thank you. Describe briefly what you mean by "similar mathematical frameworks" that are used for both systems.**<br><br>When I say "similar mathematical frameworks" for both systems, I mean that certain mathematical concepts and techniques are used to describe and analyze the behavior of both the multi-particle system in physics and the large language model.<br><br>In the case of the multi-particle system, mathematical frameworks such as statistical mechanics and thermodynamics are used to describe the behavior of the system. These frameworks involve concepts such as partition functions, free energy, and entropy, which are used to quantify the behavior of the system.<br><br>In the case of the large language model, similar mathematical frameworks are used, such as probability theory and machine learning. These frameworks involve concepts such as probability distributions, expectation values, and optimization algorithms, which are used to describe and analyze the behavior of the model. | |

Both systems can be described using mathematical equations that relate the behavior of the individual components to the emergent behavior of the system as a whole. For example, in the case of the multi-particle system, the Boltzmann distribution is a mathematical equation that relates the probability of a particular state of the system to the energy of that state. Similarly, in the case of the large language model, the probability distribution of the words in the model can be described using mathematical equations such as the softmax function.

Overall, the use of similar mathematical frameworks for both systems reflects the fact that both systems exhibit complex and coherent behavior arising from the interactions between individual components, and that mathematical concepts and techniques can be used to describe and analyze this behavior.

**Thank you. Now, relate these observations to materials failure, at a highly abstract level.**

At a highly abstract level, the observations of emergent behavior in both the multi-particle system and the large language model can be related to materials failure in the following way:

Both systems exhibit complex and coherent behavior arising from the interactions between individual components, which can lead to emergent properties that are not present in the individual components themselves. Similarly, materials failure can be viewed as an emergent property of a complex system, where the interactions between the individual components of the material lead to a coherent and often catastrophic failure.

In both cases, the emergent behavior is driven by the interactions between the individual components, rather than by any external force or driver. This means that the behavior of the system as a whole is determined by the interactions between its parts, rather than by any external influence.

Furthermore, both cases involve a transition from a stable to an unstable state, which is a hallmark of materials failure. In the case of the multi-particle system, the transition from a stable to an unstable state is driven by the interactions between the particles, while in the case of the large language model, the transition is driven by the interactions between the words.

Overall, the observations of emergent behavior in both the multi-particle system and the large language model can be related to materials failure in that both involve complex systems with many interacting components, and both involve transitions from stable to unstable states. This abstract relationship highlights the deep connections between complex systems and their emergent properties, and the ways in which materials failure can be understood as an emergent property of a complex system.

**Attention neural networks are graph-forming models. Similarly, interatomic potentials define interaction graphs in many-particle systems. Delineate clearly the analogies between a large language model built using attention neural networks and a multi-particle system in physics.**

Sure, I can help you with that.

The analogies between a large language model built using attention neural networks and a multi-particle system in physics can be delineated as follows:

1. Graph structure: Both the language model and the multi-particle system can be represented as graphs, where the nodes represent the particles or words, and the edges represent the interactions between them.
2. Attention mechanisms: The language model uses attention mechanisms to focus on specific words or phrases, while the multi-particle system uses the potentials to focus on the interactions between particles.
3. Scalability: Both the language model and the multi-particle system can be scaled up to handle larger systems, although the computational requirements for the language model may be more significant.
4. Emergent behavior: Both the language model and the multi-particle system can exhibit emergent behavior, where the interactions between the particles or words lead to complex patterns and structures that are not predictable from the individual particles or words alone.
5. Energy landscapes: Both the language model and the multi-particle system can be represented as energy landscapes, where the potentials or the attention mechanisms define the energy landscape of the system.
6. These analogies are not limited, and there may be other similarities between the two systems as well.

However, these analogies provide a useful starting point for understanding the relationships between the language model and the multi-particle system.

**Thank you. Now, elaborate further on the emergent behavior.**

Sure, I can help you with that.

The emergent behavior in both the language model and the multi-particle system arises from the interactions between the individual particles or words, which lead to complex patterns and structures that are not predictable from the individual particles or words alone. In both cases, the system as a whole exhibits properties that are not present in its individual parts, and these properties emerge from the interactions between the particles or words.

In the language model, the attention mechanisms allow the model to focus on specific words or phrases, leading to complex patterns of language use that are not predictable from the individual words alone. For example, the model may learn to use certain words or phrases in a particular context to convey a specific meaning or tone, which is not inherent in the individual words themselves. Similarly, in the multi-particle system, the potentials define the interactions between the particles, leading to complex patterns of particle arrangement that are not predictable from the individual particles alone. For example, the particles may form a crystal lattice or a more disordered arrangement, depending on the potentials and the initial conditions.

In both cases, the emergent behavior arises from the interactions between the individual particles or words, and these interactions lead to complex patterns and structures that are not present in the individual particles or words alone. The system as a whole exhibits properties that are not present in its individual parts, and these properties emerge from the interactions between the particles or words.

Furthermore, both systems can exhibit self-organization, where the complex patterns and structures arise naturally from the interactions between the particles or words, without the need for external direction or control. This self-organization is a key feature of complex systems, and it is what allows them to adapt and evolve over time.

In summary, the emergent behavior in both the language model and the multi-particle system arises from the interactions between the individual particles or words, leading to complex patterns and structures that are not predictable from the individual particles or words alone. These complex patterns and structures exhibit self-organization, and they are a key feature of complex systems.

**Thank you! Now, summarize these key insights in a few bullet points while carefully outlining the analogies between the two systems.**

Sure, here are some bullet points summarizing the key insights and analogies between the language model and the multi-particle system:

Language Model:

1. Emergent behavior arises from interactions between individual words or phrases
2. Attention mechanisms allow the model to focus on specific parts of the input
3. Complex patterns and structures emerge from the interactions, leading to self-organization

Multi-Particle System:

1. Emergent behavior arises from interactions between individual particles
2. Potentials define the interactions between particles
3. Complex patterns and structures emerge from the interactions, leading to self-organization

Analogies:

1. Both systems exhibit emergent behavior arising from interactions between individual components
2. Both systems have a graph structure representing the interactions between components
3. Both systems can exhibit self-organization, where complex patterns and structures arise naturally from the interactions
4. The language model's attention mechanisms are analogous to the multi-particle system's potentials, as both define the interactions between components

These bullet points summarize the key insights and analogies between the language model and the multi-particle system, highlighting the similarities in their emergent behavior and self-organization.

**Table S9:** Sample responses to assess knowledge retrieval, focused on multiple choice answers, obtained using GPT-4 accessed via ChatGPT. GPT-4 generally has a strong capability to produce correct answers, but fails to answer questions that are specific to the content in the book, such as hyperelastic effects on crack speed. We defer further analyses to future work, since this study was largely focused on the use of open-source fine-tuned GPT models.

| Prompt/query | Notes |
|---|---|
| **Considering these deformation mechanisms, which one is most likely to occur during interfacial fracture: (A), supersonic fracture. (B), mother-daughter cracking, (C) grain boundary diffusion, (D) buckling.**<br><br>GPT-4: Interfacial fracture pertains to the breaking or separation of a material at the interface between two phases, such as the boundary between two grains in a polycrystalline material or between a matrix and a reinforcement in a composite.<br><br>Let's evaluate each option:<br><br>(A) Supersonic fracture: This refers to a fracture that propagates at speeds greater than the speed of sound in the material. While it's a fascinating concept, it's not specifically tied to interfacial fracture mechanisms.<br><br>(B) Mother-daughter cracking: This is not a standard term I'm familiar with in the context of fracture mechanics. If it refers to some specific process, it would need more context to evaluate its relevance to interfacial fracture.<br><br>(C) Grain boundary diffusion: This is a process in which atoms migrate along grain boundaries. While grain boundary diffusion can weaken grain boundaries over time, especially under specific conditions, it isn't a fracture mechanism per se.<br><br>(D) Buckling: Buckling is an instability that occurs in structural members (like columns or plates) when they are subjected to compressive loads. It's not specific to interfacial fractures.<br><br>Based on the given options and without any other context, none of these mechanisms are directly associated with interfacial fracture in the way it's commonly understood. However, grain boundary diffusion (Option C) is the closest to interfacial processes. Still, it's important to note that grain boundary diffusion is not a fracture mechanism but can play a role in weakening interfaces over time under specific conditions. | Wrong answer |
| **Which material is most likely to lead to supersonic fracture in a brittle material: (A) hyperelastic softening. (B) hyperelastic stiffening. (C) linear material behavior.**<br><br>GPT-4: In the context of brittle materials and the propagation of cracks at high speeds, the likelihood of supersonic fracture (crack propagation faster than the speed of sound in the material) is influenced by the material's constitutive behavior.<br><br>To understand this, let's review each of the options:<br><br>(A) Hyperelastic Softening: This implies that as the material is stretched or deformed, it becomes easier to deform. In a brittle material, softening behavior could lead to an instability where the material rapidly breaks or fails. This can promote faster crack propagation, potentially up to or exceeding the material's speed of sound.<br><br>(B) Hyperelastic Stiffening: This implies that as the material is stretched or deformed, it becomes harder to deform further. Stiffening behavior acts as a resistance to rapid crack propagation because the material becomes more resistant to deformation as it is deformed.<br><br>(C) Linear Material Behavior: This represents a standard linear elastic behavior where stress is proportional to strain. It does not exhibit any of the special characteristics of the hyperelastic materials mentioned.<br><br>Considering the above explanations, hyperelastic softening (A) is the most likely to lead to supersonic fracture in a brittle material. This is because the softening behavior can cause an instability that promotes rapid crack propagation. | Wrong answer |
| **Which of these mechanisms is most likely to occur in ultra-long carbon nanotubes? (A) chemical etching. (B), buckling. (C) rupture.**<br><br>GPT-4: Ultra-long carbon nanotubes (CNTs) exhibit exceptional mechanical properties due to their unique nanostructure, but they can still undergo various types of mechanical deformation or damage under certain conditions.<br><br>(A) Chemical etching: This is a process where material is removed from a surface by applying a chemical reagent. While CNTs can indeed be chemically etched under certain conditions (for example, by strong oxidizing agents), the likelihood of this mechanism depends more on the chemical environment the CNTs are exposed to rather than their length.<br><br>(B) Buckling: Buckling is a deformation mode where a structure becomes unstable and suddenly changes its shape in response to compressive loads. Ultra-long CNTs are more susceptible to buckling when subjected to compressive forces, especially if they are not supported or anchored properly. This is because the slenderness ratio (length/diameter) is very high for ultra-long CNTs, making them more prone to buckling under compressive stresses. | **Correct answer** |
| (C) Rupture: This refers to the breaking or fracturing of the CNT. While CNTs have high tensile strength, they can rupture under extreme tensile forces. The likelihood of rupture could be influenced by defects or impurities in the CNT structure, but the length of the CNT isn't a direct factor unless it leads to other issues (e.g., tangling, knotting, or increased chances of defects).<br><br>Considering the inherent properties of ultra-long CNTs, among the given options, (B) buckling is the most likely mechanism to occur under compressive stresses. | |
| **Which of these statements is most likely true? (A) Hyperelastic stiffening leads to higher crack speeds and reduced fracture instabilities. (B) cracks always branch, no matter the hyperelastic effect. (C) Cracks do not exist in any material.**<br><br>GPT-4: Based on the provided statements:<br><br>(A) Hyperelastic stiffening leads to higher crack speeds and reduced fracture instabilities.<br>(B) cracks always branch, no matter the hyperelastic effect.<br>(C) Cracks do not exist in any material.<br><br>The most likely true statement is:<br><br>(A) Hyperelastic stiffening leads to higher crack speeds and reduced fracture instabilities.<br><br>Reasoning:<br><br>(A) While the exact relationship between hyperelastic stiffening and crack dynamics can depend on the material and conditions, it's plausible that hyperelastic stiffening could affect crack speeds and fracture instabilities in certain materials.<br><br>(B) This is a generalization. Cracks don't always branch. Their behavior can be influenced by various factors including material properties, stress concentrations, and loading conditions.<br><br>(C) This statement is definitely false. Cracks exist in many materials, especially under certain conditions of stress, temperature, and other external factors.<br><br>Therefore, statement (A) is the most likely true statement out of the options provided. However, a deeper understanding would be required to confirm the exact relationship described in (A), especially in the context of specific materials and conditions. | **Correct answer** |

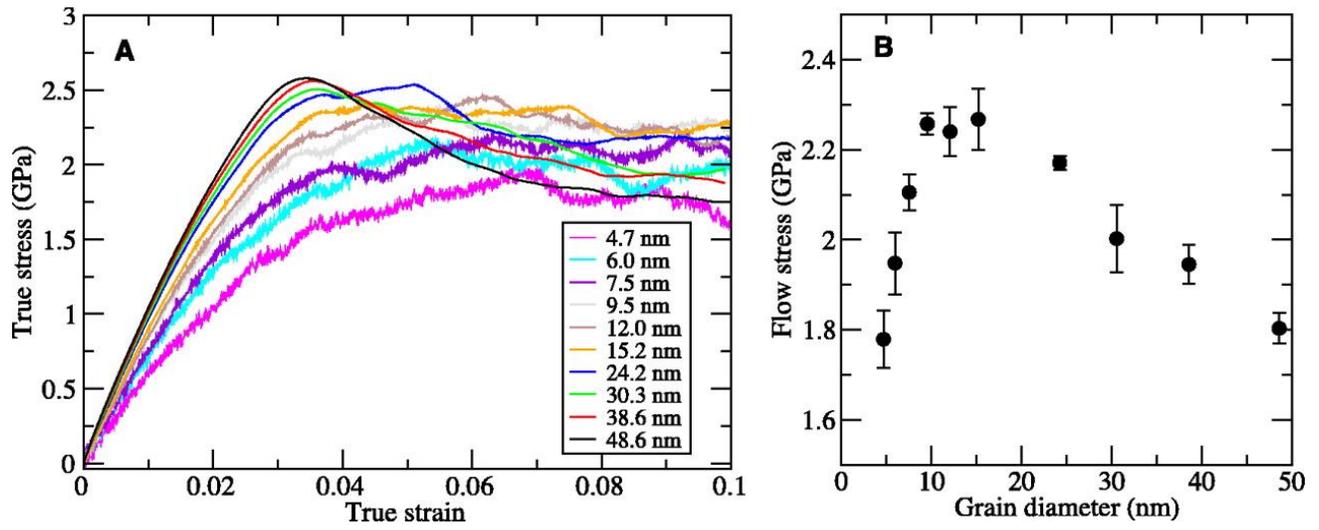

**Figure S1**: Figure 1 from [36], used for the analysis using GPT-4 (accessed via the ChatGPT interface) as a multimodal LLM. Image reproduced with permission from from Jakob Schiøtz, Karsten W. Jacobsen, "A Maximum in the Strength of Nanocrystalline Copper," Science, Vol. 301 (5638), pp. 1357-1359. Copyright © The American Association for the Advancement of Science (AAAS), 2003.